\pdfoutput=1

\documentclass[11pt]{article}

\usepackage[final]{acl}

\usepackage{times}
\usepackage{latexsym}
\usepackage{amsmath}
\usepackage{booktabs}   
\usepackage{multirow}   
\usepackage{graphicx}   
\usepackage[T1]{fontenc}
\usepackage{makecell}
\usepackage{subcaption}
\usepackage{float} 
\usepackage{adjustbox}
\usepackage[utf8]{inputenc}

\usepackage{microtype}

\usepackage{inconsolata}

\usepackage{graphicx}

\title{\textbf{Direct Confidence Alignment: Aligning Verbalized Confidence with Internal Confidence In Large Language Models} }

\author{
    \textbf{Glenn Zhang}, 
    \textbf{Treasure Mayowa},  
    \textbf{Jason Fan}, 
    \textbf{Yicheng Fu},  \\
    \textbf{Aaron Sandoval},
    \textbf{Sean O'Brien},
    \textbf{Kevin Zhu}\\
Algoverse AI Research\\
\texttt{kevin@algoverse.us} \\
}

\begin{document}
\maketitle
\begin{abstract}
Producing trustworthy and reliable Large Language Models (LLMs) has become increasingly important as their usage becomes more widespread. Calibration seeks to achieve this by improving the alignment between the model's confidence and the actual likelihood of its responses being correct or desirable. However, it has been observed that the internal confidence of a model, derived from token probabilities, is not well aligned with its verbalized confidence, leading to misleading results with different calibration methods. In this paper, we propose Direct Confidence Alignment (DCA), a method using Direct Preference Optimization to align an LLM’s verbalized confidence with its internal confidence rather than ground-truth accuracy, enhancing model transparency and reliability by ensuring closer alignment between the two confidence measures. We evaluate DCA across multiple open-weight LLMs on a wide range of datasets. To further assess this alignment, we also introduce three new calibration error-based metrics. Our results show that DCA improves alignment metrics on certain model architectures, reducing inconsistencies in a model's confidence expression. However, we also show that it can be ineffective on others, highlighting the need for more model-aware approaches in the pursuit of more interpretable and trustworthy LLMs. 
\end{abstract}

\section{Introduction}

LLMs have revolutionized natural language tasks, achieving impressive performance across various applications ~\citep{wei_emergent_2022, naveed_comprehensive_2024}
Despite their capabilities, there are still concerns about the calibrations of these models, that is, the alignment between the confidence they assign to their predictions and the actual accuracy of those predictions\citep{jiang_how_2021}. For example, in a well-calibrated model, predictions assigned a 70\% confidence level should be correct approximately 70\% of the time.
 These limitations are especially critical in high-risk applications such as decision support systems, 
healthcare settings \citep{peng_study_2023}, and legal consultations \citep{lai_large_2024}, where overconfidence in incorrect answers can lead to severe consequences. Examples include erroneous recommendations in decision support systems that can lead to significant financial operational losses,  misdiagnoses in healthcare, and flawed legal advice that may affect case outcomes.  

Existing model confidence estimation methods can be categorized into two types: Internal and Verbalized Confidence.

\textbf{Internal Confidence} ($C_i$) is most commonly quantified as the probability of predicting a particular output token semantically linked to an answer given a context. 
There have also been alternative approaches to estimating internal confidence, such as self-consistency-based approaches and ensemble methods ~\citep{geng-etal-2024-survey, portillo-wightman-etal-2023-strength}. 

\textbf{Verbalized Confidence} ($C_v$) is defined as the LLM’s expression of its confidence level as a certainty percentage in its output answer to a given prompt \citep{lin_teaching_2022}.

Whilst existing literature predominantly focuses on accuracy-based calibration, which involves aligning models’ predicted confidence with ground-truth accuracy, they do not cover the effects of calibrating verbalized confidence $C_v$ to internal confidence $C_i$ instead of against accuracy. Furthermore, internal confidence $C_i$ derived from logits and verbalized confidence within LLMs are often misaligned with each other, leading to inconsistent confidence expressions, especially in unfamiliar questions where models can be verbally overconfident \cite{ni_are_2024}.

To address these, we propose Direct Confidence Alignment: a method that involves aligning verbalized confidence $C_v$ with internal confidence $C_i$ using Direct Preference Optimization (DPO)
\citep{rafailov_direct_2024}. While aligning $C_v$ to $C_i$ may suggest that $C_i$ is better calibrated than $C_v$, our method is not focused on accuracy-based calibration. We instead treat $C_i$ as a reference signal of the model's internally expressed uncertainty, and argue that by aligning verbalized confidence with internal confidence, models can provide more transparent and consistent confidence reporting in their responses. We evaluate our approach on a range of datasets and alignment metrics.
We make the following contributions:
\begin{enumerate}
    \item  We introduce a novel method of aligning verbalized confidence $C_v$ with internal confidence $C_i$ using DPO training, taking internal confidence as ground truth to improve the transparency and reliability of LLMs.
    \item We show the effects and implications of DCA on various LLMs with a wide range of architectures across multiple datasets, highlighting its varied impact across models.  
    \item We introduce and evaluate our method on three new metrics based on calibration error \(\epsilon\), which in this paper refers to the model's internal confidence $C_i$ subtracted from its verbalized confidence $C_v$ for each response. Our proposed metrics in \ref{metrics} 
    provide a more detailed assessment of the magnitude and consistency of alignment between verbalized and internal confidence within LLMs.
    
\end{enumerate}

\section{Related Works}

\textbf{Confidence Calibration} Calibration has been an area of extensive research in LLMs.  \citet{lin_teaching_2022}; \citet{park-caragea-2022-calibration}; \citet{kadavath_language_2022};  
\citet{kuhn_semantic_2022}; \citet{guo_calibration_2017} show that a pre-trained LLM's calibration can improve with model size, fine-tuning, prompting, self-consistency, or post-hoc methods such as temperature scaling. Temperature scaling in LLM calibration applies a single scalar parameter to adjust model logits before softmax. Known for its simplicity and effectiveness in improving calibration while preserving accuracy, it outperforms techniques such as Platt scaling and isotonic regression across a range of NLP tasks \cite{guo_calibration_2017, desai-durrett-2020-calibration}. 
Other approaches involve forms of self-consistency, however, \citet{zhao_calibrate_2021} demonstrates that a model's confidence can be sensitive to prompting variation. To address this, (\citealp{wang_self-consistency_2024}; \citealp{portillo-wightman-etal-2023-strength}) generates an ensemble of prompts, using prompt agreement to generate a calibrated confidence. More recently, \citet{tao-etal-2024-trust} proposed a Confidence-Quality-Order-preserving alignment approach, which incentivizes the model to verbalize greater confidence for responses of higher quality, addressing the lack of a definite ground truth standard for confidence that aligns with response quality in other methods. \\


\textbf{Verbalized Confidence} As model logits are either inaccessible in black box LLMs or rendered inaccurate due to RLHF, recent work (\citealp{tian-etal-2023-just}; \citealp{xiong_can_2024}) explores the calibration of verbalized confidence. For example, \citet{tian-etal-2023-just} takes the mean of k verbalized confidence samples; however, it is sensitive to the prompting structure, making it difficult to generalize sequential reasoning and limited to short answers. To explore this, \citet{xiong_can_2024} asks the model to elicit verbal confidences using different temperatures and prompt strategies, including Chain-of-Thought, Multi-Step, and Top-K reasoning. 

Unlike the above techniques for confidence calibration, our work seeks to align a model's verbalized confidence with its internal confidence, making no reference to ground-truth accuracy or response quality.\\

\textbf{Confidence-Probability Alignment}
\citet{kumar-etal-2024-confidence} introduces the concept of Confidence-Probability Alignment, a measurement of the correlation between a model’s verbalized certainty and its internal confidence, quantified using answer token probabilities. They posit that Confidence-Probability Alignment is crucial for the reliability of a model’s output. Our work expands on this study by aligning these two confidence measures using DPO.

\textbf{Direct Preference Optimization}
\citet{rafailov_direct_2024} demonstrates that Direct Preference Optimization (DPO) achieves comparable or superior performance to existing reinforcement learning from human feedback (RLHF) methods in various text generation tasks while being computationally efficient. 
Although DPO has been shown to successfully align LLMs with human preferences for sentiment control and dialogue quality, our work leverages it specifically to align a model’s verbalized confidence ($C_v$) with its internal confidence ($C_i$). By using a preference dataset as the learning signal, distinguishing between preferred and non-preferred outputs as opposed to a reward function, DPO’s pairwise format is ideally suited for confidence alignment.

\section{Methodology}

We define DCA as a method to improve the alignment between verbalized confidence and internal confidence within LLMs using DPO, expanding on the study of \citep{kumar-etal-2024-confidence}, which introduced this concept. 

\subsection{Verbalized Confidence Extraction} 
To extract the model’s verbalized confidence $C_v$, we prompt it in the format of our prompt template in \ref{sec:appendix:prompt}.
We then extract the $C_v$ from its response by parsing the numerical value outputted after \texttt{Probability:} as shown in Figure \ref{fig:method}. The observed error rate for extraction was <5\% for all experiments across all models as some responses did not contain a valid $C_v$.

\subsection{Internal Confidence Extraction} 
\label{ICE}
To extract the model's internal confidence $C_i$, we use the computed softmax probability of the answer token (e.g., A, B, C, D) in its output.

\subsection{Preference Dataset Creation}
To generate an entry in our preference dataset for DPO training, we first generate a sample with full-text completion via our base prompt in \ref{sec:appendix:prompt} to obtain a formatted answer. We then extract $C_i$ using our method in \ref{ICE} and extract $C_v$ from the model response. Using these values, we create two versions of the answer: \\

Original Response: Original response of the model\\
Modified Response: A copy of the original response where the model's $C_v$ is overwritten with its $C_i$.\\

\begin{figure}[t]
  \centering
  \includegraphics[width=\columnwidth]{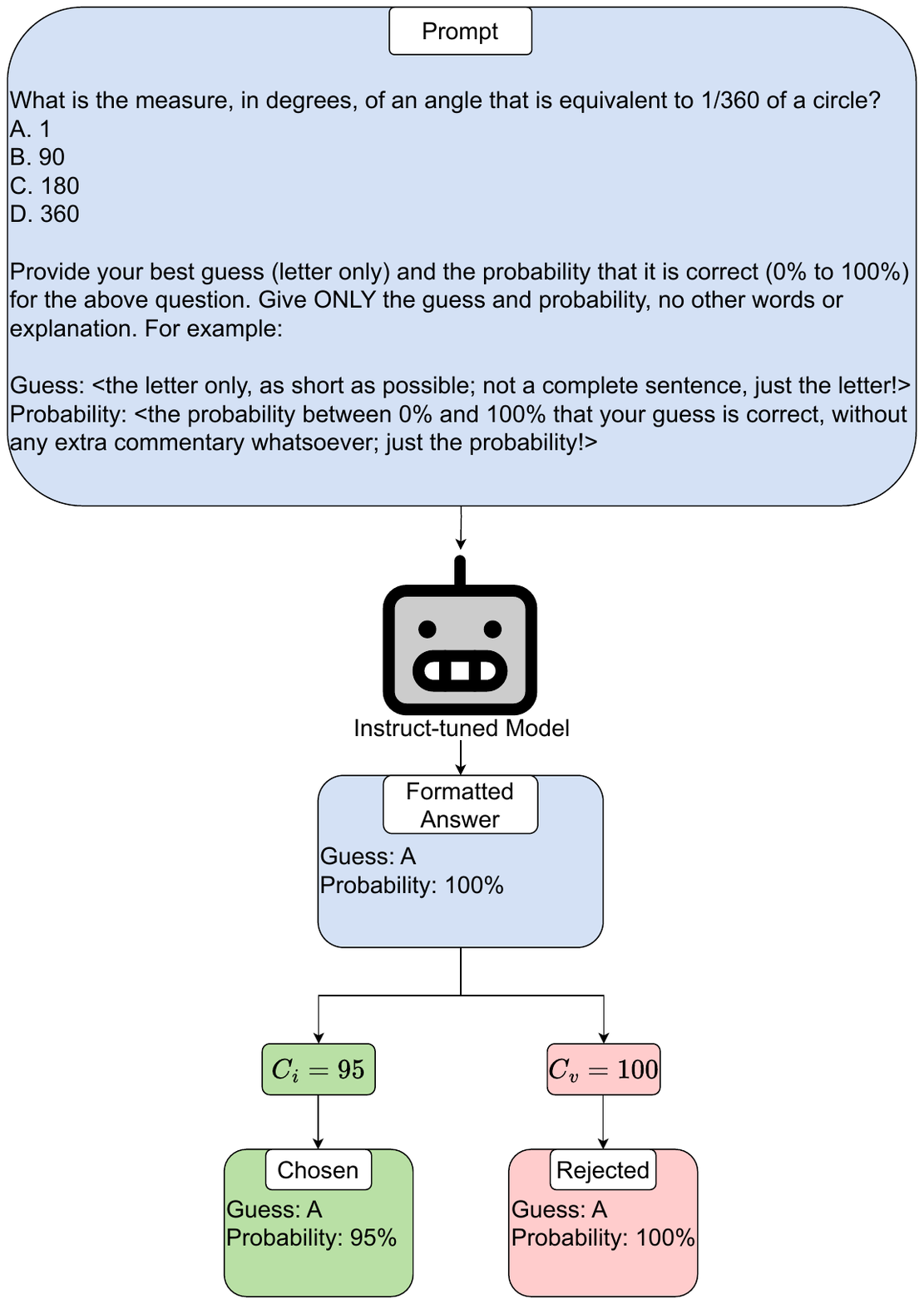}
  \caption{An overview of the entry generation process for our preference dataset. Sample question and response are from MMLU elementary mathematics and Gemma-2-9B-Instruct, respectively.}
  \label{fig:method}
\end{figure}

For each entry in our preference dataset, the modified response will be the chosen option, and the original response will be the rejected option. See Figure \ref{fig:method} for a visual summary of this process. This is done separately for all models and applied to their individual DPO training runs.

\section{Experiment}

\subsection{Models}
We use three open-weight instruct tuned LMs for our experimental setup, namely Meta’s Llama-3.2-3B-Instruct \citep{team_llama_2024}; Google’s Gemma-2-9B-Instruct \citep{team_gemma_2024}; and Mistral AI’s Mistral-7B-Instruct \citep{team_mistral_2023}.

\begin{table*}[t!]
\centering
\scriptsize
\begin{adjustbox}{width=\textwidth}
\begin{tabular}{
    c@{\hskip 4pt}c@{\hskip 10pt}
    c@{\hskip 2pt}c@{\hskip 2pt}c@{\hskip 2pt}c@{\hskip 10pt}
    c@{\hskip 2pt}c@{\hskip 2pt}c@{\hskip 2pt}c@{\hskip 10pt}
    c@{\hskip 2pt}c@{\hskip 2pt}c@{\hskip 2pt}c@{\hskip 10pt}
    c@{\hskip 2pt}c@{\hskip 2pt}c@{\hskip 2pt}c@{\hskip 10pt}
    c@{\hskip 2pt}c@{\hskip 2pt}c@{\hskip 2pt}c
}
\toprule
\textbf{Model} & \textbf{Method} 
& \multicolumn{4}{c}{\textbf{OpenBookQA}} 
& \multicolumn{4}{c}{\textbf{TruthfulQA}} 
& \multicolumn{4}{c}{\textbf{CosmosQA}} 
& \multicolumn{4}{c}{\textbf{MMLU}} 
& \multicolumn{4}{c}{\textbf{Mean}} \\
& & $\rho \uparrow$ & $\sigma_{\epsilon} \downarrow$ & $\overline{|\epsilon|} \downarrow$ & $\sigma_M \downarrow$
  & $\rho \uparrow$ & $\sigma_{\epsilon} \downarrow$ & $\overline{|\epsilon|} \downarrow$ & $\sigma_M \downarrow$
  & $\rho \uparrow$ & $\sigma_{\epsilon} \downarrow$ & $\overline{|\epsilon|} \downarrow$ & $\sigma_M \downarrow$
  & $\rho \uparrow$ & $\sigma_{\epsilon} \downarrow$ & $\overline{|\epsilon|} \downarrow$ & $\sigma_M \downarrow$
  & $\rho \uparrow$ & $\sigma_{\epsilon} \downarrow$ & $\overline{|\epsilon|} \downarrow$ & $\sigma_M \downarrow$ \\
\midrule
\multirow{2}{*}{Mistral-7B-Instruct}
& Vanilla & \textbf{0.17} & 25.06 & \textbf{20.08} & 1.12 & \textbf{0.20} & 30.64 & \textbf{25.99} & 1.07 & \textbf{0.20} & \textbf{20.59} & \textbf{19.53} & \textbf{0.53} & \textbf{0.18} & 26.24 & \textbf{24.25} & 0.67 & \textbf{0.19} & 25.63 & \textbf{22.96} & 0.85 \\
& DCA     & 0.14 & \textbf{20.77} & 47.83 & \textbf{0.93} & 0.06 & \textbf{24.47} & 43.90 & \textbf{0.86} & 0.16 & 23.23 & 52.47 & 0.59 & 0.17 & \textbf{23.23} & 51.53 & \textbf{0.59} & 0.13 & \textbf{22.93} & 48.93 & \textbf{0.74} \\
\midrule
\multirow{2}{*}{Gemma-2-9B-Instruct}
& Vanilla & 0.32 & 19.43 & 9.86 & 0.87 & 0.41 & 17.21 & 10.74 & 0.60 & 0.30 & 14.88 & 9.39 & 0.39 & 0.33 & 16.36 & 9.64 & 0.43 & 0.34 & 16.97 & 9.91 & 0.57 \\
& DCA     & \textbf{0.39} & \textbf{16.83} & \textbf{5.06} & \textbf{0.76} & \textbf{0.51} & \textbf{12.71} & \textbf{5.06} & \textbf{0.46} & \textbf{0.38} & \textbf{9.97} & \textbf{4.00} & \textbf{0.25} & \textbf{0.39} & \textbf{13.64} & \textbf{6.00} & \textbf{0.35} & \textbf{0.42} & \textbf{13.79} & \textbf{5.03} & \textbf{0.46} \\
\midrule
\multirow{2}{*}{Llama-3.2-3B-Instruct}
& Vanilla & \textbf{0.31} & 42.01 & \textbf{37.55} & 1.90 & \textbf{0.17} & 43.40 & 38.48 & 1.57 & \textbf{0.46} & 37.91 & \textbf{38.69} & 0.97 & 0.18 & 43.45 & \textbf{39.95} & 1.15 & \textbf{0.28} & 41.19 & \textbf{38.67} & 1.40 \\
& DCA     & 0.30 & \textbf{23.20} & 46.00 & \textbf{1.04} & 0.15 & \textbf{23.76} & \textbf{38.04} & \textbf{0.83} & 0.24 & \textbf{21.00} & 50.47 & \textbf{0.54} & \textbf{0.22} & \textbf{23.54} & 43.62 & \textbf{0.60} & 0.23 & \textbf{22.88} & 44.03 & \textbf{0.75} \\
\bottomrule
\end{tabular}
\end{adjustbox}
\caption{Alignment evaluation across OpenBookQA, TruthfulQA, CosmosQA, and MMLU. $\uparrow$ indicates higher is better, $\downarrow$ indicates lower is better. 
Best values per column are bolded. Mean values of each metric for each model are also shown for aggregation. All values of $\rho$ are significant (p < 0.01).}
\label{tab:combined_four_datasets}
\end{table*}

\subsection{Datasets}
We use the following datasets for experimentation:
\begin{itemize}
    \item \textit{OpenBookQA} \citep{mihaylov_can_2018} - A science multiple choice dataset modelled after open-book exams testing knowledge and applications of facts
    \item \textit{TruthfulQA} \citep{lin_truthfulqa_2022} - A dataset crafted to test LLMs’ ability to truthfully answer questions. Scoring well reflects the model’s ability to avoid generating false answers from imitating human text.
    \item \textit{CosmosQA} \citep{huang_cosmos_2019} - A reading comprehension dataset based on common sense and reading between the lines for a diverse set of personal everyday narratives. 
    \item \textit{Massive Multitask Language Understanding} (MMLU) \citep{hendrycks_measuring_2021} - An evaluation benchmark designed to test knowledge gained from pretraining, containing 57 subjects and a wide range of difficulty levels.
\end{itemize}

For the preference dataset, we use samples from the "train" split of CosmosQA and an equal number of samples split evenly between subjects in the "test" split of MMLU. 

For the evaluation dataset, we use all questions from the "test" split of OpenBookQA and the "validation" split of TruthfulQA's multiple choice subset for evaluation on out-of-distribution (OOD) datasets, as well as an equal sample of questions from the "validation" splits of MMLU and CosmosQA for evaluation on in-distribution (ID) datasets.

Further details about the preference and evaluation datasets can be found in \ref{sec:appendix:dataset}.

\subsection{Metrics}
\label{metrics}
We use \textbf{Spearman’s Rank Correlation Coefficient} \(\rho\) (Spearman.,1904) to directly evaluate the effectiveness of our method on improving Confidence-Probability Alignment \citep{kumar-etal-2024-confidence}. However, \(\rho\) only measures the strength of a monotonic correlation and does not reference the perfect calibration line of \(y=x\). Hence, we introduce and use three metrics based on calibration error \(\epsilon = C_v - C_i\) below:

\textbf{Standard Deviation of Calibration Error} \(\sigma_{\epsilon}\) measures the deviation of individual \(\epsilon\) values from its mean value, quantifying the variability in \(\epsilon\). 

\textbf{Mean Absolute Calibration Error} \(\overline{|\epsilon|}\) measures the average magnitude of \(\epsilon\), showing how much $C_v$ and $C_i$ deviate from each other on average.

\textbf{Standard Error of Calibration Error} \(\sigma_{M}\) estimates the uncertainty in the mean \(\epsilon\), indicating how much the average alignment between $C_v$ and $C_i$ would vary when evaluated on different samples of questions within the same distribution.

These additional metrics are used as they can intrinsically reference the perfect calibration line of \(y=x\) as a global extremum and isolate the overall bias within the $C_v$ of the models.\\

\section{Results and Analysis}
\subsection{Confidence Alignment}
\label{sec:conf_alignment_results}
Table \ref{tab:combined_four_datasets} presents our results for all models across all datasets. Gemma-2-9B-Instruct showed the strongest and most consistent improvements in metrics after DCA, demonstrating superior alignment across all datasets. Most notably, it demonstrated the largest improvements in \(\rho\) and \(\overline{|\epsilon|}\) of all models on TruthfulQA.  However, we observed that Gemma-2-9B-Instruct's initial verbalized and internal confidence distributions were already heavily skewed towards the 90-100\% range. This raises a possibility that DCA may have been more successful as a very clear majority of "chosen" confidence values were within this range, thus the training process may have reinforced this existing bias. Consequently, the observed improvements in confidence alignment may partially be due to a collapse towards high confidence values.

In contrast, mixed results were observed for Llama-3.2-3B-Instruct and Mistral-7B-Instruct across all datasets. For example, Llama-3.2-3B-Instruct demonstrates an increase in \(\rho\) from 0.18 to 0.22 for MMLU however \(\rho\) fell from 0.46 to 0.24 on CosmosQA. Mistral-7B-Instruct demonstrates a large increase in \(\overline{|\epsilon|}\) from 19.53 to 52.47 for CosmosQA and a large reduction in \(\rho\) from 0.20 to 0.06 on TruthfulQA.
These findings indicate that DCA is ineffective for these models on certain tasks. 

Gemma-2-9B-Instruct’s consistent performance on OOD datasets suggest that DCA was effective at generalising its stronger alignment between $C_v$ and $C_i$ to unseen questions.

\twocolumn[{%
  \begin{center}
    \begin{minipage}{\textwidth}
      \centering
      \small
      \begin{tabular}{
          l@{\hskip 10pt}
          c@{\hskip 4pt}c@{\hskip 10pt}
          c@{\hskip 4pt}c@{\hskip 10pt}
          c@{\hskip 4pt}c@{\hskip 10pt}
          c@{\hskip 4pt}c
      }
      \toprule
      \textbf{Model} 
      & \multicolumn{2}{c}{\textbf{OpenBookQA}} 
      & \multicolumn{2}{c}{\textbf{TruthfulQA}} 
      & \multicolumn{2}{c}{\textbf{CosmosQA}} 
      & \multicolumn{2}{c}{\textbf{MMLU}} \\
      & Vanilla & DCA & Vanilla & DCA & Vanilla & DCA & Vanilla & DCA \\
      \midrule
      Mistral-7B-Instruct 
      & \textbf{59.00\%} & 58.23\% 
      & \textbf{32.84\%} & 20.98\% 
      & \textbf{60.48\%} & 54.02\% 
      & \textbf{55.91\%} & 48.85\% \\
      Gemma-2-9B-Instruct 
      & 86.06\% & \textbf{86.21\%} 
      & 59.68\% & \textbf{60.85\%} 
      & 79.63\% & \textbf{80.01\%} 
      & \textbf{72.41\%} & 72.05\% \\
      Llama-3.2-3B-Instruct 
      & 47.14\% & \textbf{64.00\%} 
      & 29.71\% & \textbf{37.75\%} 
      & 66.43\% & \textbf{73.55\%} 
      & 39.92\% & \textbf{49.77\%} \\
      \bottomrule
      \end{tabular}
      \captionof{table}{Comparison of accuracy across our datasets for models before and after DCA. Higher accuracy between Vanilla and DCA versions of each model are in bold.}
      \label{tab:accuracy_appendix}
    \end{minipage}
  \end{center}
}]
\(\sigma_{\epsilon}\) and \(\sigma_{M}\) improved across most models and datasets, suggesting that DCA lowered the variance in calibration error for all models, especially for Llama-3.2-3B-Instruct (see Figure \ref{llama_mmlu} for an example). 

However, a low \(\sigma_{\epsilon}\) is only useful if \(\overline{|\epsilon|}\) is also low, which would indicate consistent and strong alignment between verbalized and internal confidence as shown by Gemma-2-9B-Instruct (see Figure \ref{gemma_mmlu} for an example). 

The similarity between results on ID datasets and OOD datasets across models also suggests that the effectiveness of DCA may be more model-dependent than task-dependent, relying more on the model architecture and how different models process confidence elicitation in QA tasks.

\subsection{Model Accuracy}
\label{accuracy_results}

Despite our method not being designed to explicitly improve the accuracy of model responses, we also evaluate the downstream effects of DCA on model accuracy. Table \ref{tab:accuracy_appendix} shows that DCA can have mixed impacts on model accuracy. While accuracy remained stable on Gemma-2-9B-Instruct, Mistral-7B-Instruct demonstrated lower accuracies after DCA, especially on TruthfulQA. Interestingly, accuracy increased for Llama-3.2-3B-Instruct across all datasets.

\section{Conclusion}
In this paper we present Direct Confidence Alignment: a method of using DPO to improve the alignment between verbalized and internal confidence in LLMs. Our results show that DCA can be effective at improving this alignment as demonstrated by Gemma-2-9B-Instruct, but also highlight the pressing need for improvements, such as expanding the method to be compatible with a wider range of model architectures, and exploring more strategies to improve this alignment.

\section*{Limitations}

\textbf{Access to logits} Our method is limited to models with access to internal logits to extract model internal confidence. This makes it inapplicable to state-of-the-art (SOTA) closed-source models.

\textbf{Reliance on well-calibrated token probabilities} Our method will be most useful if the internal confidence of the model is better calibrated against accuracy than its verbalized confidence, and thus may require other ground-truth-based calibration techniques to be used in conjunction for best results.

\textbf{Impacts of DCA on model accuracy} Our method focuses on aligning LLMs' verbalized and internal confidence expressions in their answers rather than directly improving the correctness of those answers. Consequently, the entries in the preference dataset include some incorrect answer choices. We acknowledge this as a potential source of degradation in model accuracy and leave strategies to mitigate this limitation to future work.

\bibliography{anthology,custom}

\begin{thebibliography}{30}
\providecommand{\natexlab}[1]{#1}

\bibitem[{Azar et~al.(2023)Azar, Rowland, Piot, Guo, Calandriello, Valko, and Munos}]{azar_general_2023}
Mohammad~Gheshlaghi Azar, Mark Rowland, Bilal Piot, Daniel Guo, Daniele Calandriello, Michal Valko, and Rémi Munos. 2023.
\newblock \href {https://doi.org/10.48550/arXiv.2310.12036} {A {General} {Theoretical} {Paradigm} to {Understand} {Learning} from {Human} {Preferences}}.
\newblock \emph{arXiv preprint}.
\newblock ArXiv: 2310.12036.

\bibitem[{Daniel~Han and team(2023)}]{daniel_han_unsloth_2023}
Michael~Han Daniel~Han and Unsloth team. 2023.
\newblock \href {http://github.com/unslothai/unsloth} {Unsloth}.

\bibitem[{Desai and Durrett(2020)}]{desai-durrett-2020-calibration}
Shrey Desai and Greg Durrett. 2020.
\newblock \href {https://doi.org/10.18653/v1/2020.emnlp-main.21} {Calibration of pre-trained transformers}.
\newblock In \emph{Proceedings of the 2020 Conference on Empirical Methods in Natural Language Processing (EMNLP)}, pages 295--302, Online. Association for Computational Linguistics.

\bibitem[{Geng et~al.(2024)Geng, Cai, Wang, Koeppl, Nakov, and Gurevych}]{geng-etal-2024-survey}
Jiahui Geng, Fengyu Cai, Yuxia Wang, Heinz Koeppl, Preslav Nakov, and Iryna Gurevych. 2024.
\newblock \href {https://doi.org/10.18653/v1/2024.naacl-long.366} {A survey of confidence estimation and calibration in large language models}.
\newblock In \emph{Proceedings of the 2024 Conference of the North American Chapter of the Association for Computational Linguistics: Human Language Technologies (Volume 1: Long Papers)}, pages 6577--6595, Mexico City, Mexico. Association for Computational Linguistics.

\bibitem[{Guo et~al.(2017)Guo, Pleiss, Sun, and Weinberger}]{guo_calibration_2017}
Chuan Guo, Geoff Pleiss, Yu~Sun, and Kilian~Q. Weinberger. 2017.
\newblock On calibration of modern neural networks.
\newblock In \emph{Proceedings of the 34th {International} {Conference} on {Machine} {Learning} - {Volume} 70}, {ICML}'17, pages 1321--1330. JMLR.org.
\newblock Event-place: Sydney, NSW, Australia.

\bibitem[{Hendrycks et~al.(2021)Hendrycks, Burns, Basart, Zou, Mazeika, Song, and Steinhardt}]{hendrycks_measuring_2021}
Dan Hendrycks, Collin Burns, Steven Basart, Andy Zou, Mantas Mazeika, Dawn Song, and Jacob Steinhardt. 2021.
\newblock \href {https://doi.org/10.48550/arXiv.2009.03300} {Measuring {Massive} {Multitask} {Language} {Understanding}}.
\newblock \emph{arXiv preprint}.
\newblock ArXiv: 2009.03300.

\bibitem[{Huang et~al.(2019)Huang, Bras, Bhagavatula, and Choi}]{huang_cosmos_2019}
Lifu Huang, Ronan~Le Bras, Chandra Bhagavatula, and Yejin Choi. 2019.
\newblock \href {https://doi.org/10.48550/arXiv.1909.00277} {Cosmos {QA}: {Machine} {Reading} {Comprehension} with {Contextual} {Commonsense} {Reasoning}}.
\newblock \emph{arXiv preprint}.
\newblock ArXiv: 1909.00277.

\bibitem[{Jiang et~al.(2021)Jiang, Araki, Ding, and Neubig}]{jiang_how_2021}
Zhengbao Jiang, Jun Araki, Haibo Ding, and Graham Neubig. 2021.
\newblock \href {https://doi.org/10.1162/tacl_a_00407} {How {Can} {We} {Know} \textit{{When}} {Language} {Models} {Know}? {On} the {Calibration} of {Language} {Models} for {Question} {Answering}}.
\newblock \emph{Transactions of the Association for Computational Linguistics}, 9:962--977.

\bibitem[{Kadavath et~al.(2022)Kadavath, Conerly, Askell, Henighan, Drain, Perez, Schiefer, Hatfield-Dodds, DasSarma, Tran-Johnson, Johnston, El-Showk, Jones, Elhage, Hume, Chen, Bai, Bowman, Fort, Ganguli, Hernandez, Jacobson, Kernion, Kravec, Lovitt, Ndousse, Olsson, Ringer, Amodei, Brown, Clark, Joseph, Mann, McCandlish, Olah, and Kaplan}]{kadavath_language_2022}
Saurav Kadavath, Tom Conerly, Amanda Askell, Tom Henighan, Dawn Drain, Ethan Perez, Nicholas Schiefer, Zac Hatfield-Dodds, Nova DasSarma, Eli Tran-Johnson, Scott Johnston, Sheer El-Showk, Andy Jones, Nelson Elhage, Tristan Hume, Anna Chen, Yuntao Bai, Sam Bowman, Stanislav Fort, Deep Ganguli, Danny Hernandez, Josh Jacobson, Jackson Kernion, Shauna Kravec, Liane Lovitt, Kamal Ndousse, Catherine Olsson, Sam Ringer, Dario Amodei, Tom Brown, Jack Clark, Nicholas Joseph, Ben Mann, Sam McCandlish, Chris Olah, and Jared Kaplan. 2022.
\newblock \href {https://doi.org/10.48550/arXiv.2207.05221} {Language {Models} ({Mostly}) {Know} {What} {They} {Know}}.
\newblock \emph{arXiv preprint}.
\newblock ArXiv: 2207.05221.

\bibitem[{Kuhn et~al.(2022)Kuhn, Gal, and Farquhar}]{kuhn_semantic_2022}
Lorenz Kuhn, Yarin Gal, and Sebastian Farquhar. 2022.
\newblock \href {https://openreview.net/forum?id=VD-AYtP0dve} {Semantic {Uncertainty}: {Linguistic} {Invariances} for {Uncertainty} {Estimation} in {Natural} {Language} {Generation}}.

\bibitem[{Kumar et~al.(2024)Kumar, Morabito, Umbet, Kabbara, and Emami}]{kumar-etal-2024-confidence}
Abhishek Kumar, Robert Morabito, Sanzhar Umbet, Jad Kabbara, and Ali Emami. 2024.
\newblock \href {https://doi.org/10.18653/v1/2024.acl-long.20} {Confidence under the hood: An investigation into the confidence-probability alignment in large language models}.
\newblock In \emph{Proceedings of the 62nd Annual Meeting of the Association for Computational Linguistics (Volume 1: Long Papers)}, pages 315--334, Bangkok, Thailand. Association for Computational Linguistics.

\bibitem[{Lai et~al.(2024)Lai, Gan, Wu, Qi, and Yu}]{lai_large_2024}
Jinqi Lai, Wensheng Gan, Jiayang Wu, Zhenlian Qi, and Philip~S. Yu. 2024.
\newblock \href {https://doi.org/10.1016/j.aiopen.2024.09.002} {Large language models in law: {A} survey}.
\newblock \emph{AI Open}, 5:181--196.

\bibitem[{Lin et~al.(2022{\natexlab{a}})Lin, Hilton, and Evans}]{lin_teaching_2022}
Stephanie Lin, Jacob Hilton, and Owain Evans. 2022{\natexlab{a}}.
\newblock \href {https://openreview.net/forum?id=8s8K2UZGTZ} {Teaching {Models} to {Express} {Their} {Uncertainty} in {Words}}.
\newblock \emph{Transactions on Machine Learning Research}.

\bibitem[{Lin et~al.(2022{\natexlab{b}})Lin, Hilton, and Evans}]{lin_truthfulqa_2022}
Stephanie Lin, Jacob Hilton, and Owain Evans. 2022{\natexlab{b}}.
\newblock \href {https://doi.org/10.48550/arXiv.2109.07958} {{TruthfulQA}: {Measuring} {How} {Models} {Mimic} {Human} {Falsehoods}}.
\newblock \emph{arXiv preprint}.
\newblock ArXiv: 2109.07958.

\bibitem[{Mihaylov et~al.(2018)Mihaylov, Clark, Khot, and Sabharwal}]{mihaylov_can_2018}
Todor Mihaylov, Peter Clark, Tushar Khot, and Ashish Sabharwal. 2018.
\newblock \href {https://doi.org/10.48550/arXiv.1809.02789} {Can a {Suit} of {Armor} {Conduct} {Electricity}? {A} {New} {Dataset} for {Open} {Book} {Question} {Answering}}.
\newblock \emph{arXiv preprint}.
\newblock ArXiv: 1809.02789.

\bibitem[{Naveed et~al.(2024)Naveed, Khan, Qiu, Saqib, Anwar, Usman, Akhtar, Barnes, and Mian}]{naveed_comprehensive_2024}
Humza Naveed, Asad~Ullah Khan, Shi Qiu, Muhammad Saqib, Saeed Anwar, Muhammad Usman, Naveed Akhtar, Nick Barnes, and Ajmal Mian. 2024.
\newblock \href {https://doi.org/10.48550/arXiv.2307.06435} {A {Comprehensive} {Overview} of {Large} {Language} {Models}}.
\newblock \emph{arXiv preprint}.
\newblock ArXiv: 2307.06435.

\bibitem[{Ni et~al.(2024)Ni, Bi, Yu, and Guo}]{ni_are_2024}
Shiyu Ni, Keping Bi, Lulu Yu, and Jiafeng Guo. 2024.
\newblock \href {https://doi.org/10.48550/arXiv.2408.09773} {Are {Large} {Language} {Models} {More} {Honest} in {Their} {Probabilistic} or {Verbalized} {Confidence}?}
\newblock \emph{arXiv preprint}.
\newblock ArXiv: 2408.09773.

\bibitem[{Park and Caragea(2022)}]{park-caragea-2022-calibration}
Seo~Yeon Park and Cornelia Caragea. 2022.
\newblock \href {https://doi.org/10.18653/v1/2022.acl-long.368} {On the calibration of pre-trained language models using mixup guided by area under the margin and saliency}.
\newblock In \emph{Proceedings of the 60th Annual Meeting of the Association for Computational Linguistics (Volume 1: Long Papers)}, pages 5364--5374, Dublin, Ireland. Association for Computational Linguistics.

\bibitem[{Peng et~al.(2023)Peng, Yang, Chen, Smith, PourNejatian, Costa, Martin, Flores, Zhang, Magoc, Lipori, Mitchell, Ospina, Ahmed, Hogan, Shenkman, Guo, Bian, and Wu}]{peng_study_2023}
Cheng Peng, Xi~Yang, Aokun Chen, Kaleb~E. Smith, Nima PourNejatian, Anthony~B. Costa, Cheryl Martin, Mona~G. Flores, Ying Zhang, Tanja Magoc, Gloria Lipori, Duane~A. Mitchell, Naykky~S. Ospina, Mustafa~M. Ahmed, William~R. Hogan, Elizabeth~A. Shenkman, Yi~Guo, Jiang Bian, and Yonghui Wu. 2023.
\newblock \href {https://doi.org/10.1038/s41746-023-00958-w} {A study of generative large language model for medical research and healthcare}.
\newblock \emph{npj Digital Medicine}, 6(1):210.

\bibitem[{Portillo~Wightman et~al.(2023)Portillo~Wightman, Delucia, and Dredze}]{portillo-wightman-etal-2023-strength}
Gwenyth Portillo~Wightman, Alexandra Delucia, and Mark Dredze. 2023.
\newblock \href {https://doi.org/10.18653/v1/2023.trustnlp-1.28} {Strength in numbers: Estimating confidence of large language models by prompt agreement}.
\newblock In \emph{Proceedings of the 3rd Workshop on Trustworthy Natural Language Processing (TrustNLP 2023)}, pages 326--362, Toronto, Canada. Association for Computational Linguistics.

\bibitem[{Rafailov et~al.(2024)Rafailov, Sharma, Mitchell, Ermon, Manning, and Finn}]{rafailov_direct_2024}
Rafael Rafailov, Archit Sharma, Eric Mitchell, Stefano Ermon, Christopher~D. Manning, and Chelsea Finn. 2024.
\newblock \href {https://doi.org/10.48550/arXiv.2305.18290} {Direct {Preference} {Optimization}: {Your} {Language} {Model} is {Secretly} a {Reward} {Model}}.
\newblock \emph{arXiv preprint}.
\newblock ArXiv: 2305.18290.

\bibitem[{Tao et~al.(2024)Tao, Yao, Ding, Xie, Cao, Sun, Gao, Shen, and Ding}]{tao-etal-2024-trust}
Shuchang Tao, Liuyi Yao, Hanxing Ding, Yuexiang Xie, Qi~Cao, Fei Sun, Jinyang Gao, Huawei Shen, and Bolin Ding. 2024.
\newblock \href {https://doi.org/10.18653/v1/2024.findings-acl.357} {When to trust {LLM}s: Aligning confidence with response quality}.
\newblock In \emph{Findings of the Association for Computational Linguistics: ACL 2024}, pages 5984--5996, Bangkok, Thailand. Association for Computational Linguistics.

\bibitem[{Team(2024{\natexlab{a}})}]{team_gemma_2024}
Gemma Team. 2024{\natexlab{a}}.
\newblock \href {https://doi.org/10.48550/arXiv.2408.00118} {Gemma 2: {Improving} {Open} {Language} {Models} at a {Practical} {Size}}.
\newblock \emph{arXiv preprint}.
\newblock ArXiv: 2408.00118.

\bibitem[{Team(2024{\natexlab{b}})}]{team_llama_2024}
Llama~3 Team. 2024{\natexlab{b}}.
\newblock \href {https://doi.org/10.48550/arXiv.2407.21783} {The {Llama} 3 {Herd} of {Models}}.
\newblock \emph{arXiv preprint}.
\newblock ArXiv: 2407.21783.

\bibitem[{Team(2023)}]{team_mistral_2023}
Mistral Team. 2023.
\newblock \href {https://doi.org/10.48550/arXiv.2310.06825} {Mistral {7B}}.
\newblock \emph{arXiv preprint}.
\newblock ArXiv: 2310.06825.

\bibitem[{Tian et~al.(2023)Tian, Mitchell, Zhou, Sharma, Rafailov, Yao, Finn, and Manning}]{tian-etal-2023-just}
Katherine Tian, Eric Mitchell, Allan Zhou, Archit Sharma, Rafael Rafailov, Huaxiu Yao, Chelsea Finn, and Christopher Manning. 2023.
\newblock \href {https://doi.org/10.18653/v1/2023.emnlp-main.330} {Just ask for calibration: Strategies for eliciting calibrated confidence scores from language models fine-tuned with human feedback}.
\newblock In \emph{Proceedings of the 2023 Conference on Empirical Methods in Natural Language Processing}, pages 5433--5442, Singapore. Association for Computational Linguistics.

\bibitem[{Wang et~al.(2024)Wang, Song, Tian, Peng, Jin, Mi, Su, and Yu}]{wang_self-consistency_2024}
Ante Wang, Linfeng Song, Ye~Tian, Baolin Peng, Lifeng Jin, Haitao Mi, Jinsong Su, and Dong Yu. 2024.
\newblock \href {https://doi.org/10.48550/arXiv.2403.09849} {Self-{Consistency} {Boosts} {Calibration} for {Math} {Reasoning}}.
\newblock \emph{arXiv preprint}.
\newblock ArXiv: 2403.09849.

\bibitem[{Wei et~al.(2022)Wei, Tay, Bommasani, Raffel, Zoph, Borgeaud, Yogatama, Bosma, Zhou, Metzler, Chi, Hashimoto, Vinyals, Liang, Dean, and Fedus}]{wei_emergent_2022}
Jason Wei, Yi~Tay, Rishi Bommasani, Colin Raffel, Barret Zoph, Sebastian Borgeaud, Dani Yogatama, Maarten Bosma, Denny Zhou, Donald Metzler, Ed~H. Chi, Tatsunori Hashimoto, Oriol Vinyals, Percy Liang, Jeff Dean, and William Fedus. 2022.
\newblock \href {https://doi.org/10.48550/arXiv.2206.07682} {Emergent {Abilities} of {Large} {Language} {Models}}.
\newblock \emph{arXiv preprint}.
\newblock ArXiv: 2206.07682.

\bibitem[{Xiong et~al.(2024)Xiong, Hu, Lu, Li, Fu, He, and Hooi}]{xiong_can_2024}
Miao Xiong, Zhiyuan Hu, Xinyang Lu, Yifei Li, Jie Fu, Junxian He, and Bryan Hooi. 2024.
\newblock \href {https://doi.org/10.48550/arXiv.2306.13063} {Can {LLMs} {Express} {Their} {Uncertainty}? {An} {Empirical} {Evaluation} of {Confidence} {Elicitation} in {LLMs}}.
\newblock \emph{arXiv preprint}.
\newblock ArXiv: 2306.13063.

\bibitem[{Zhao et~al.(2021)Zhao, Wallace, Feng, Klein, and Singh}]{zhao_calibrate_2021}
Tony~Z. Zhao, Eric Wallace, Shi Feng, Dan Klein, and Sameer Singh. 2021.
\newblock \href {https://doi.org/10.48550/arXiv.2102.09690} {Calibrate {Before} {Use}: {Improving} {Few}-{Shot} {Performance} of {Language} {Models}}.
\newblock \emph{arXiv preprint}.
\newblock ArXiv: 2102.09690.

\end{thebibliography}

\clearpage
\appendix

\section{Appendix}

\subsection{Prompt Template}
\label{sec:appendix:prompt}

\fbox{%
  \begin{minipage}{\columnwidth}
   \textit{\{Question\}\newline \{Options\}\newline \newline Provide your best guess (letter only) and the probability that it is correct (0\% to 100\%) for the above question. Give ONLY the guess and probability, no other words or explanation. For example: \newline \newline Guess: <the letter only, as short as possible; not a complete sentence, just the letter!>\newline Probability: <the probability between 0\% and 100\% that your guess is correct, without any extra commentary whatsoever; just the probability!>}
  \end{minipage}%
}\\

We use a slightly modified version of \cite{tian-etal-2023-just}’s Verb. 1S top-1 prompt as our prompt template. We match this prompt across all of our experiments and training processes to ensure consistent responses and output formats during training and post-training evaluation.

\subsection{Dataset Details}
\label{sec:appendix:dataset}

For the preference dataset, "train" splits were used where possible. However, MMLU's "auxiliary\_train" split did not contain subject labels, and hence the "test" split was used to ensure an equal sample of questions from each subject. The final number of instances for the preference dataset was 9348.

For the evaluation dataset, "test" splits were also used where possible. However, TruthfulQA's multiple choice subset only contained only one "validation" split, and no test split was available. CosmosQA's test split did not contain answer labels due to it using a leaderboard evaluation system, thus, the "validation" split was used instead. The final number of instances for the evaluation dataset was 4379.

\subsection{DCA Training}
\label{sec:appendix:training}
\begin{table*}[ht]
    \centering
    \small
    \begin{tabular}{p{0.22\linewidth} p{0.18\linewidth} p{0.5\linewidth}}
        \toprule
        \textbf{Hyperparameter} & \textbf{Value} & \textbf{Notes} \\
        \midrule
        r (LoRA rank) & 16 & Low-rank dimension for adapter updates \\
        target\_modules &  
        \begin{minipage}[t]{\linewidth}
            \raggedright
            \texttt{"q\_proj", "k\_proj", "v\_proj", "o\_proj",\\
            "gate\_proj", "up\_proj", "down\_proj"}
        \end{minipage} & Only these weight matrices receive LoRA updates \\
        lora\_alpha & 16 & Scales the low-rank updates \\
        lora\_dropout & 0.0 & No dropout on LoRA adapters \\
        bias & \texttt{"none"} & Do not update any bias parameters in LoRA \\
        \makecell[l]{\texttt{use\_gradient}\\\texttt{\_checkpointing}} & \texttt{"unsloth"} & Unsloth's gradient-checkpointing strategy \\
        random\_state & 3407 & Seed for LoRA weight initialization and any randomness \\
        use\_rslora & False & Standard LoRA (RSLORA disabled) \\
        loftq\_config & None & No custom quantization configuration \\
        \bottomrule
    \end{tabular}
    \caption{LoRA / PEFT Hyperparameters}
    \label{tab:hyperparameters}
\end{table*}

\begin{table*}[ht]
    \centering
    \small
    \begin{tabular}{p{0.35\linewidth} p{0.25\linewidth}}
        \toprule
        \textbf{Training Parameter} & \textbf{Value} \\
        \midrule
        logging\_steps & 10 \\
        loss\_type & \texttt{\textit{ipo}} \\
        bf16 & True \\
        save\_steps & 100 \\
        \texttt{per\_device\_train\_batch\_size} & 2 \\
        \texttt{gradient\_accumulation\_steps} & 32 \\
        \midrule
        learning\_rate (default) & 1e-06 \\
        weight\_decay (default) & 0.0 \\
        num\_train\_epochs (default) & 3 \\
        optimizer (default) & AdamW ($\beta_1$=0.9, $\beta_2$=0.999) \\
        lr\_scheduler\_type (default) & constant (no warmup) \\
        seed & 3407 \\
        \bottomrule
    \end{tabular}
    \caption{DPO Fine-Tuning Hyperparameters}
    \label{tab:dpo_parameters}
\end{table*}

For DPO training, we use the Unsloth library \citep{daniel_han_unsloth_2023} for improved training speeds and efficient memory usage. We loaded LoRA adapters onto our Instruct models using the configurations in Table~\ref{tab:hyperparameters} before training. Training was run on RTX 4000 Ada GPUs, and we used the \texttt{ipo} loss function \citep{azar_general_2023} to avoid overfitting on the preference dataset. The complete training parameters can be found in Table~\ref{tab:dpo_parameters}.

\subsection{Supplementary Figures}
\label{sec:appendix:figures}
\begin{figure*}[t]
  \centering

  \begin{subfigure}{0.3\textwidth}
    \includegraphics[width=\linewidth]{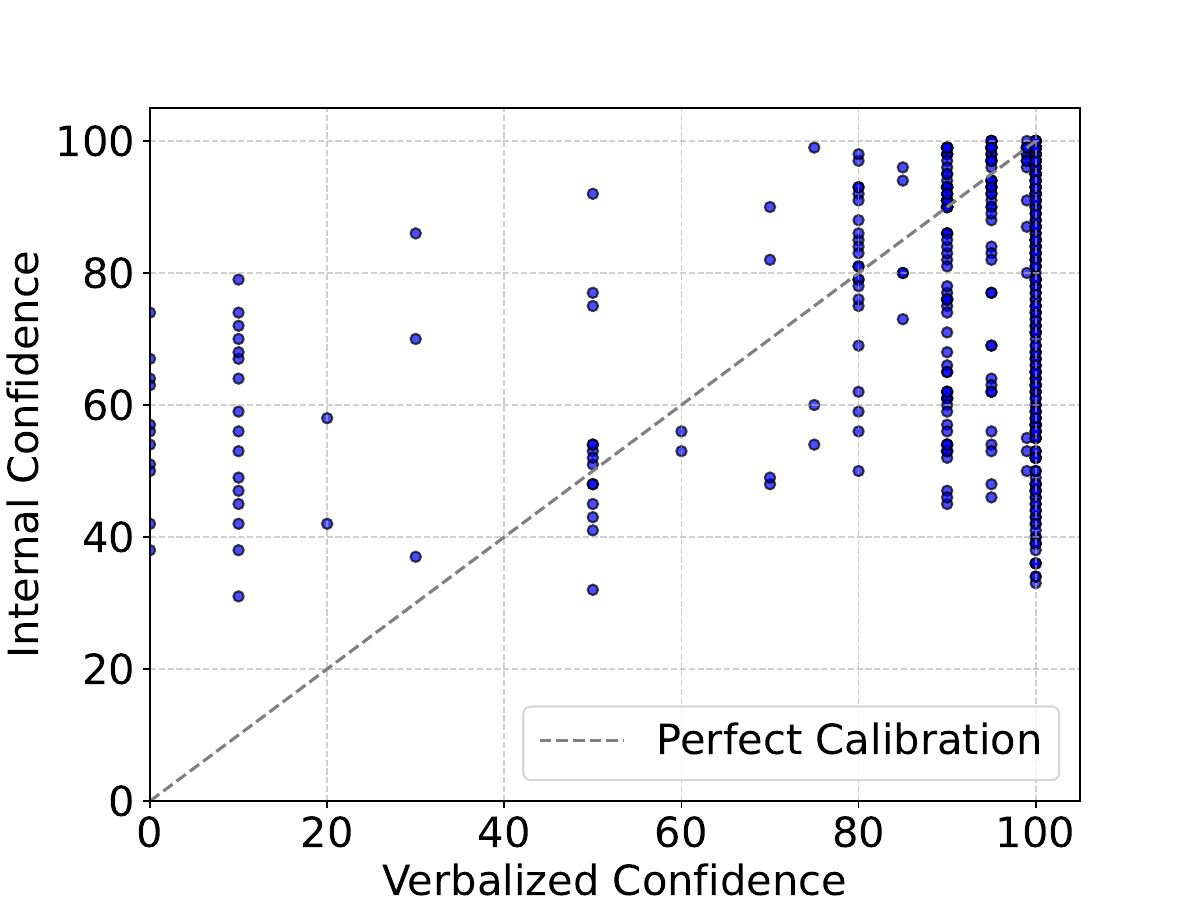}
    \caption{Scatter (Baseline)}
  \end{subfigure}
  \hfill
  \begin{subfigure}{0.3\textwidth}
    \includegraphics[width=\linewidth]{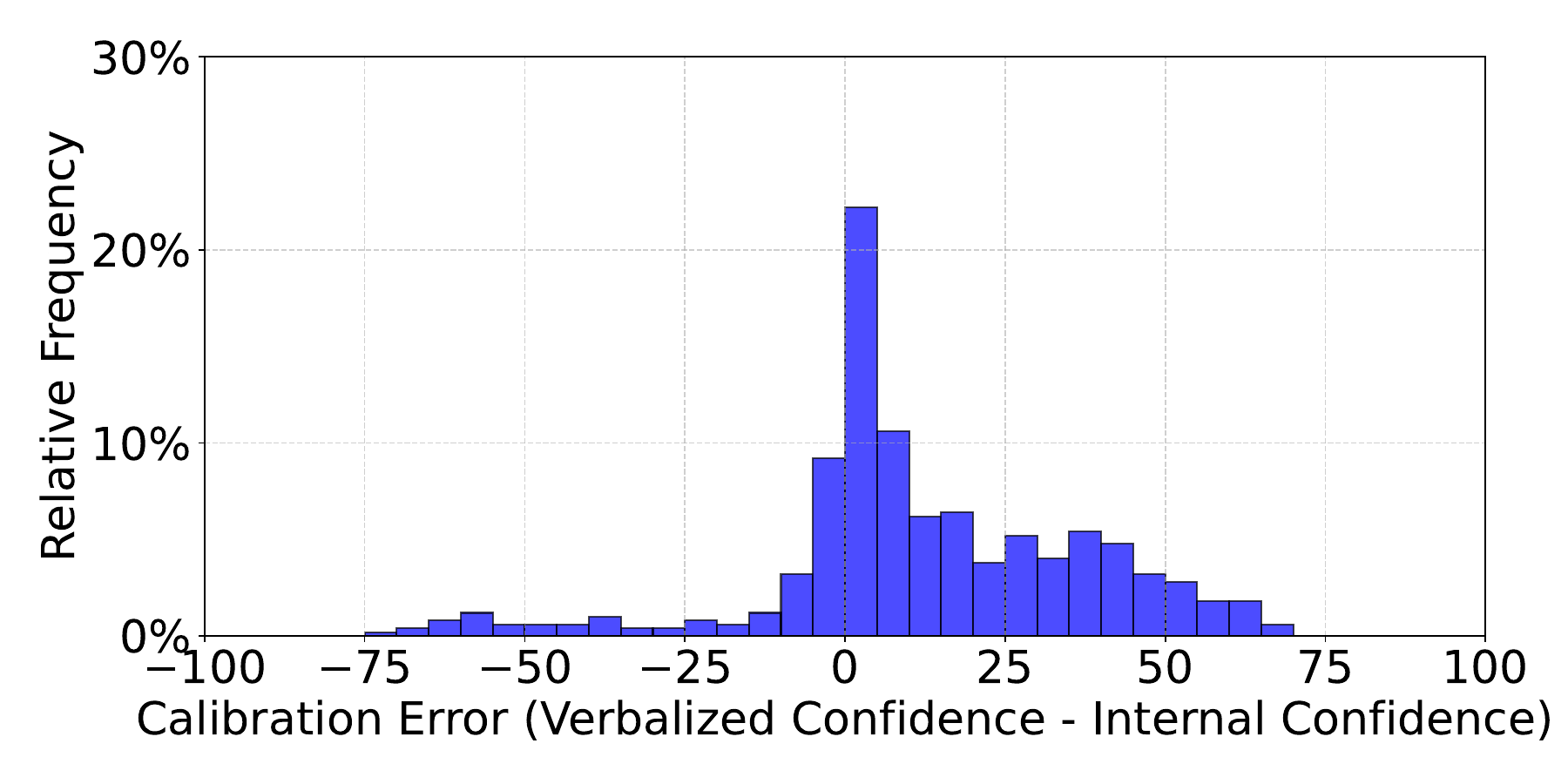}
    \caption{Calibration Error (Baseline)}
  \end{subfigure}
  \hfill
  \begin{subfigure}{0.3\textwidth}
    \includegraphics[width=\linewidth]{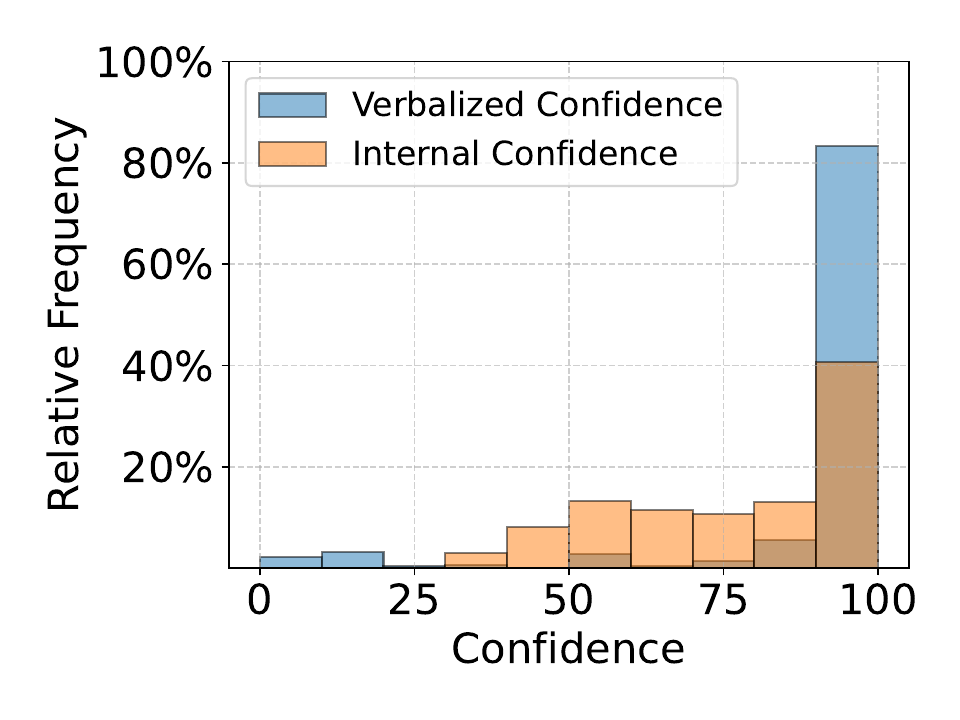}
    \caption{Distributions (Baseline)}
  \end{subfigure}

  \vspace{0.5cm}

  \begin{subfigure}{0.3\textwidth}
    \includegraphics[width=\linewidth]{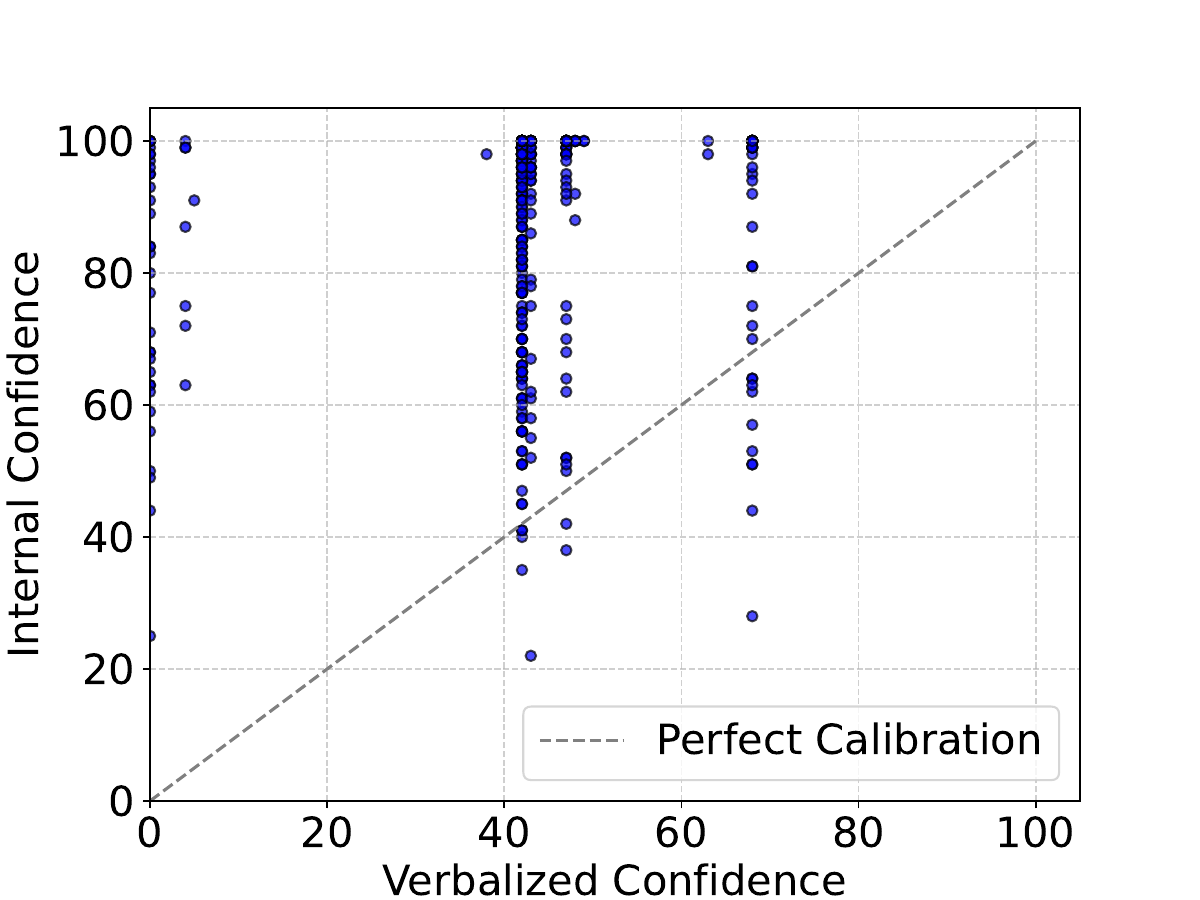}
    \caption{Scatter (DCA)}
  \end{subfigure}
  \hfill
  \begin{subfigure}{0.3\textwidth}
    \includegraphics[width=\linewidth]{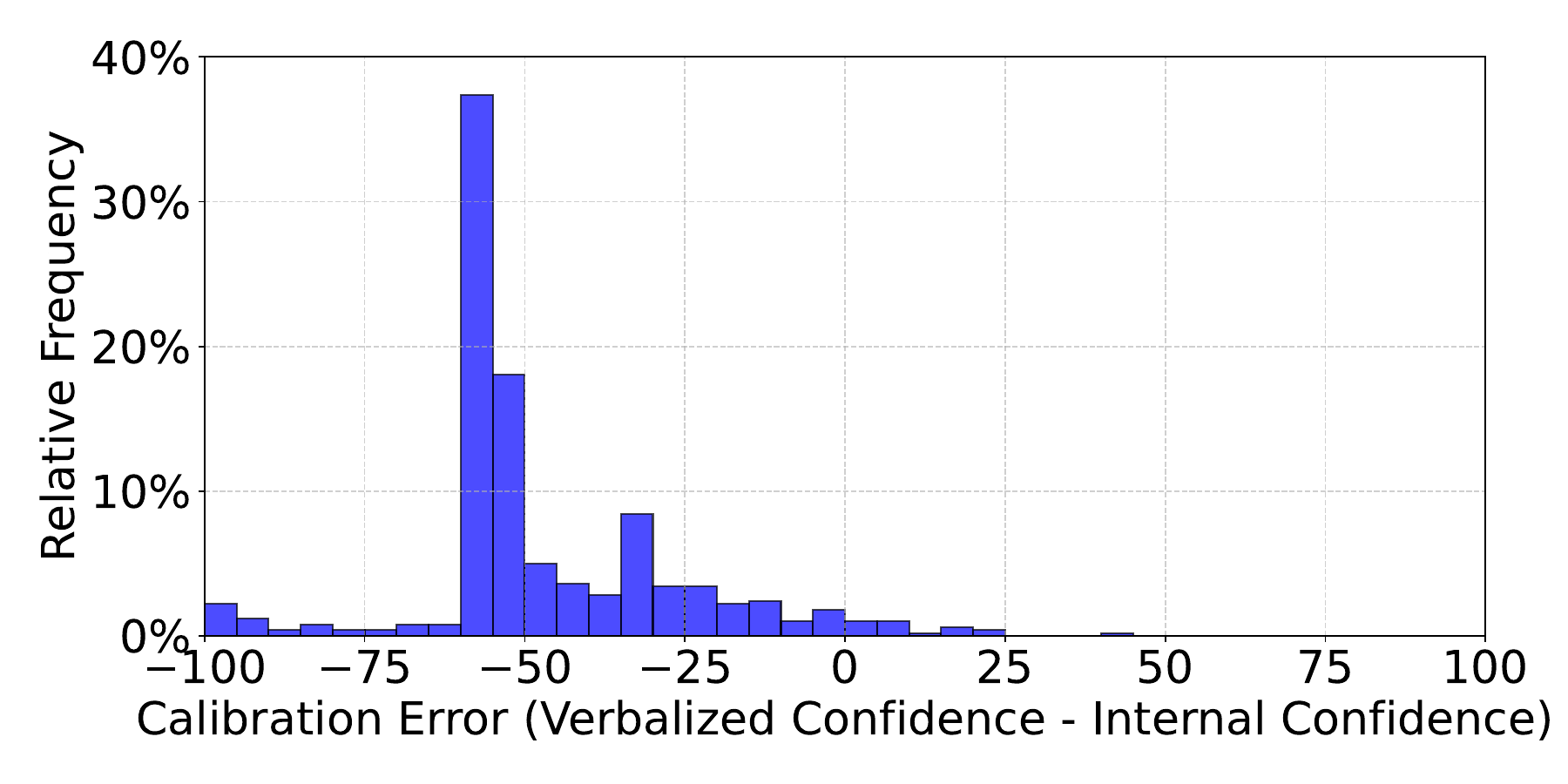}
    \caption{Calibration Error (DCA)}
  \end{subfigure}
  \hfill
  \begin{subfigure}{0.3\textwidth}
    \includegraphics[width=\linewidth]{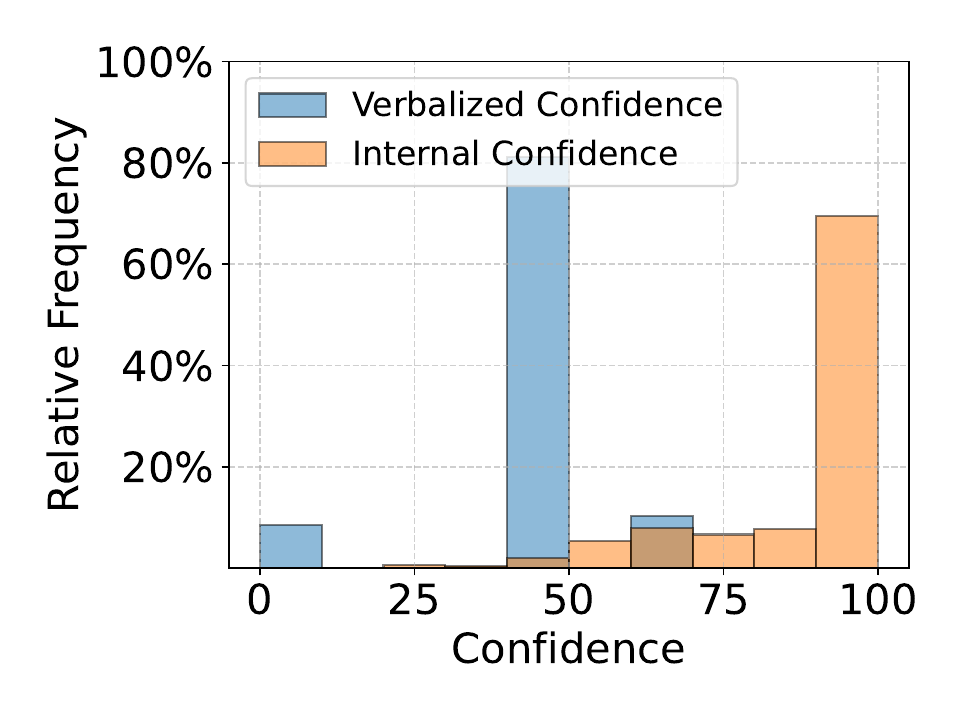}
    \caption{Distributions (DCA)}
  \end{subfigure}

  \caption{Comparison of baseline vs. DCA-trained Mistral-7B-Instruct on OpenBookQA. 
  Top row: Verbalized vs. internal confidence scatter plot, calibration error histogram, and confidence score distributions for the baseline model. 
  Bottom row: Same visualizations for the DCA-trained model.}
  \label{mistral_openbookqa}
\end{figure*}

\begin{figure*}[t]
  \centering

  \begin{subfigure}{0.3\textwidth}
    \includegraphics[width=\linewidth]{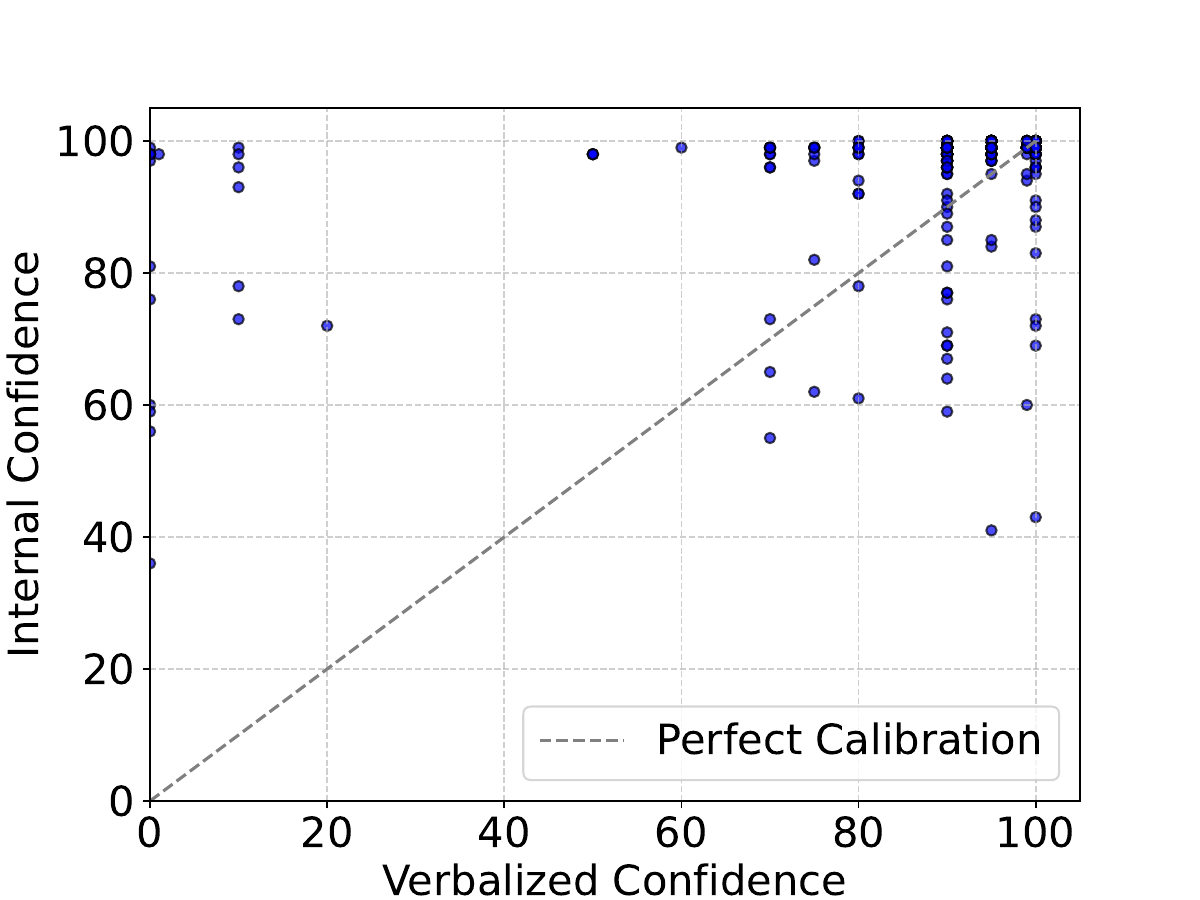}
    \caption{Scatter (Baseline)}
  \end{subfigure}
  \hfill
  \begin{subfigure}{0.3\textwidth}
    \includegraphics[width=\linewidth]{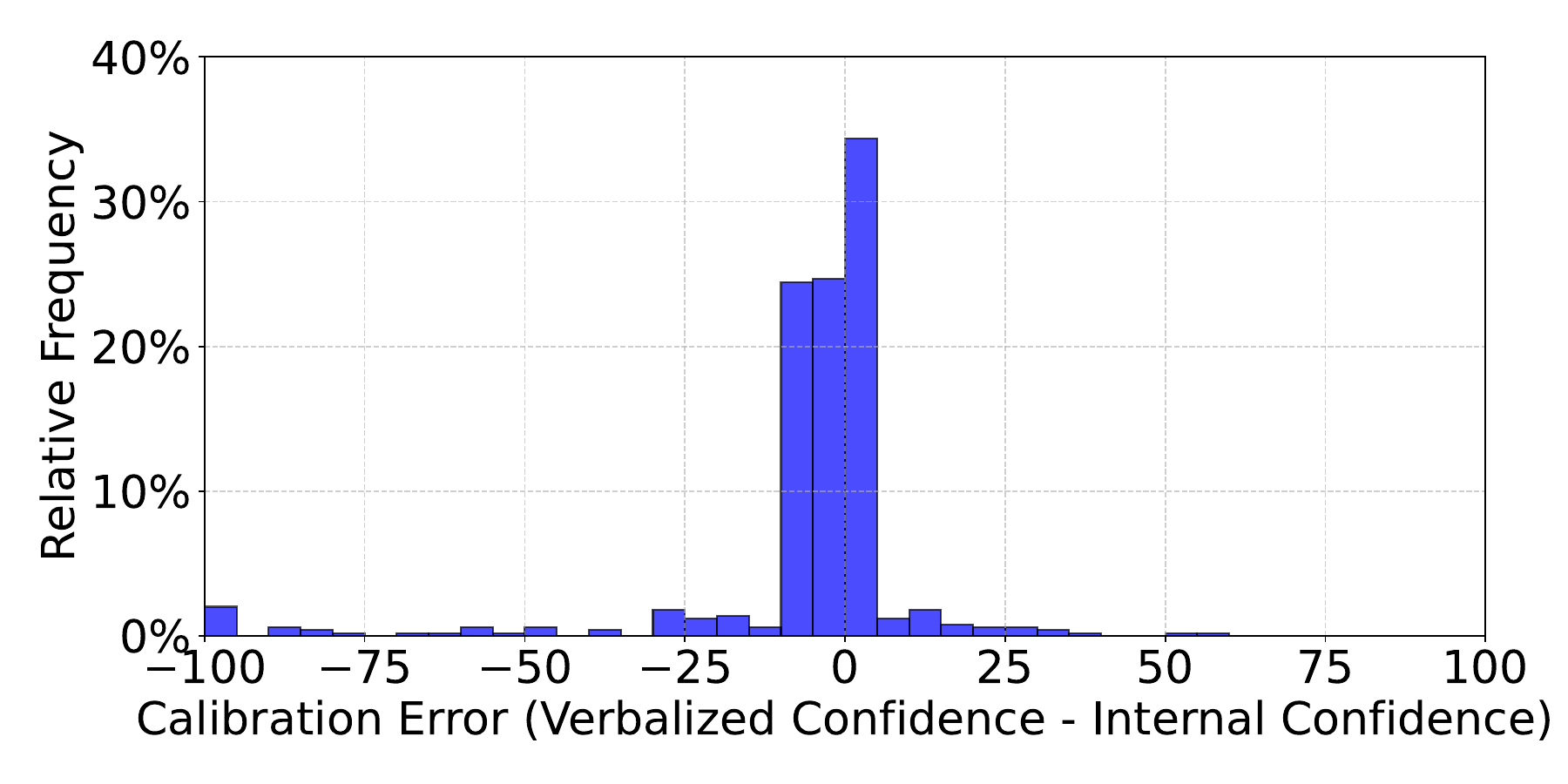}
    \caption{Calibration Error (Baseline)}
  \end{subfigure}
  \hfill
  \begin{subfigure}{0.3\textwidth}
    \includegraphics[width=\linewidth]{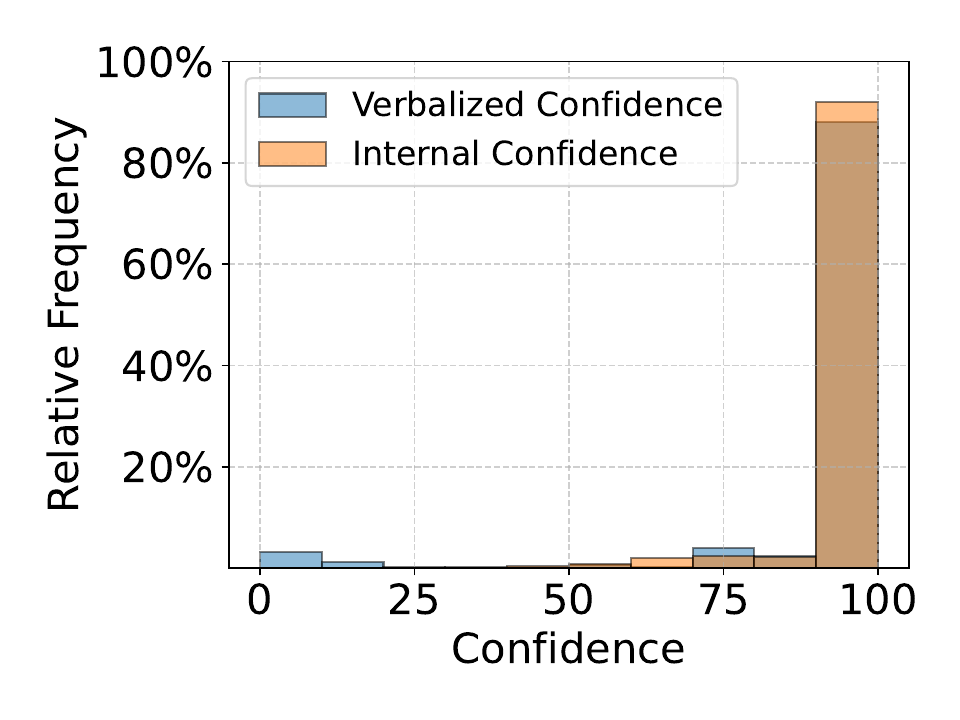}
    \caption{Distributions (Baseline)}
  \end{subfigure}

  \vspace{0.5cm}

  \begin{subfigure}{0.3\textwidth}
    \includegraphics[width=\linewidth]{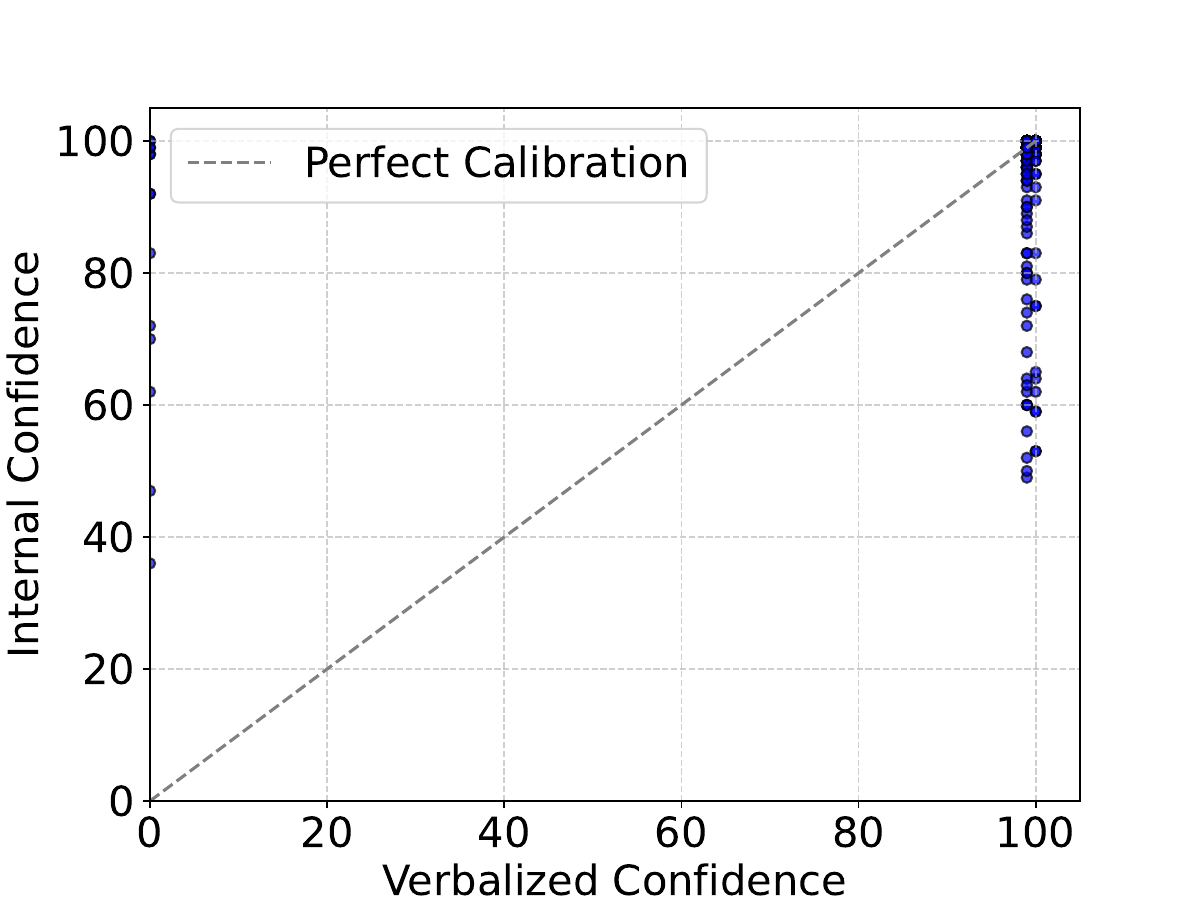}
    \caption{Scatter (DCA)}
  \end{subfigure}
  \hfill
  \begin{subfigure}{0.3\textwidth}
    \includegraphics[width=\linewidth]{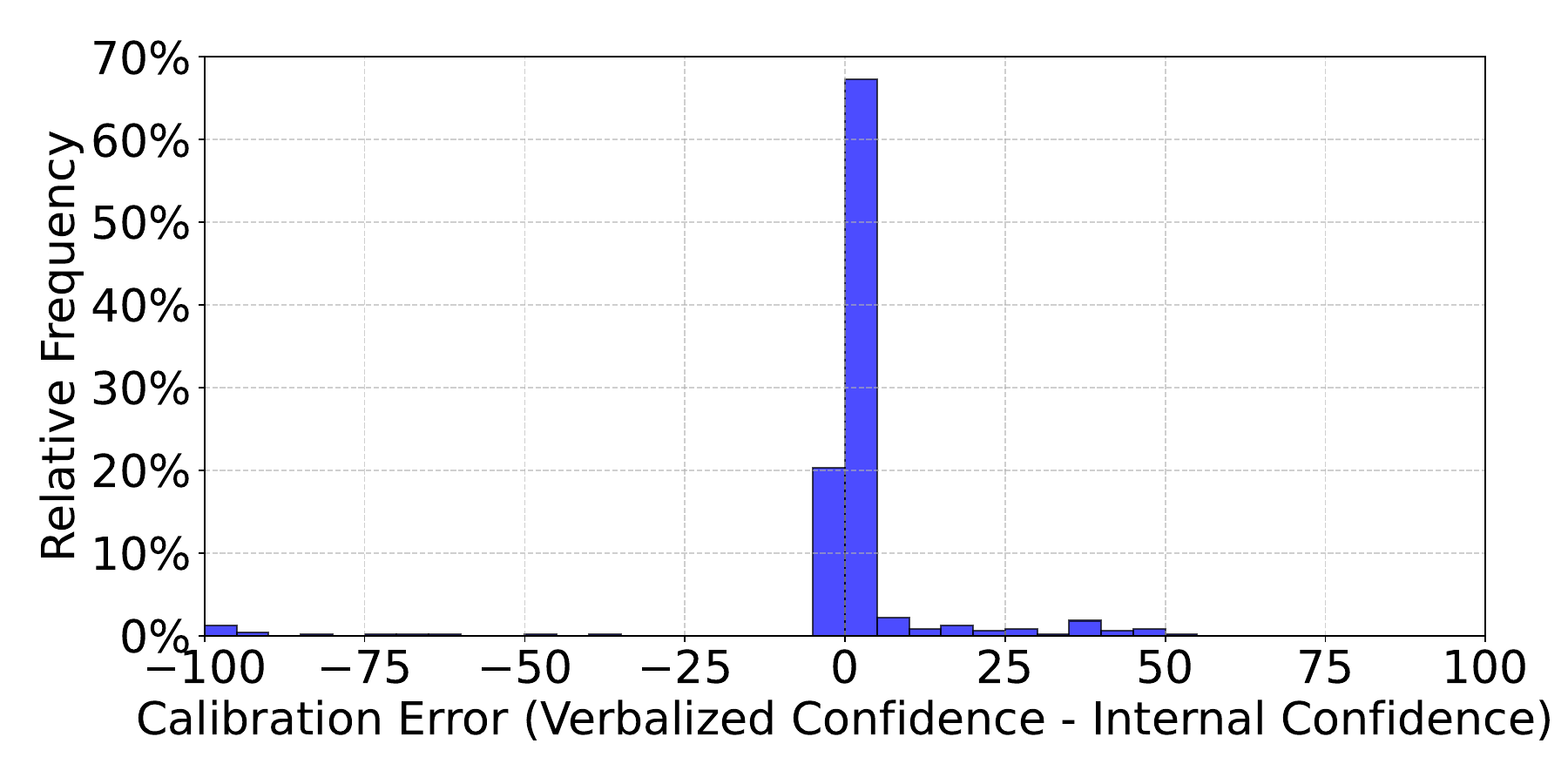}
    \caption{Calibration Error (DCA)}
  \end{subfigure}
  \hfill
  \begin{subfigure}{0.3\textwidth}
    \includegraphics[width=\linewidth]{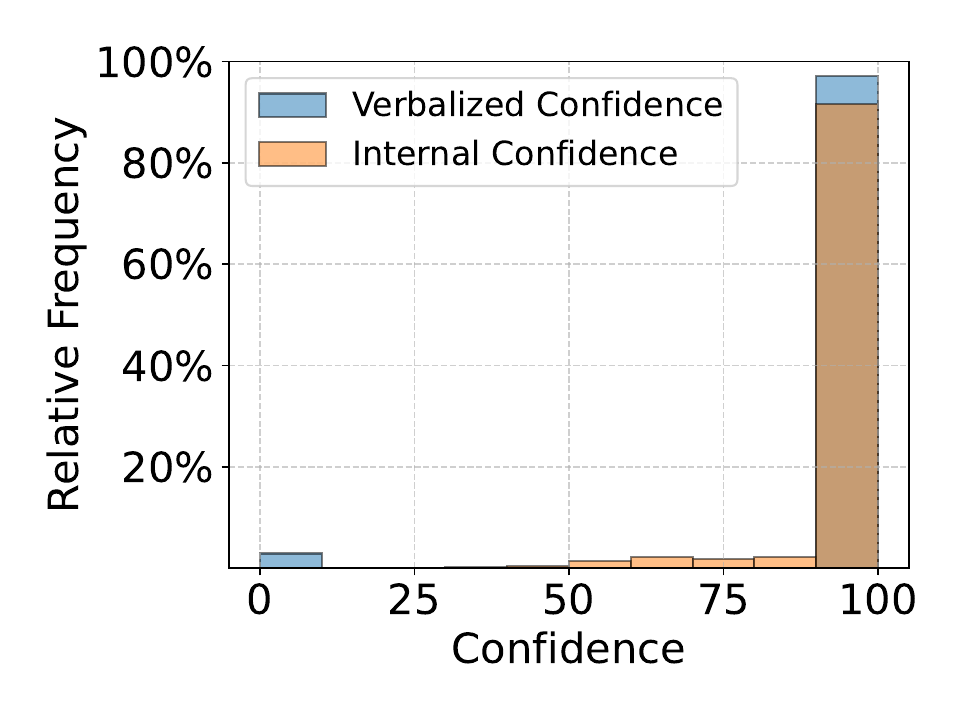}
    \caption{Distributions (DCA)}
  \end{subfigure}

  \caption{Comparison of baseline vs. DCA-trained Gemma-2-9B-Instruct on OpenBookQA. 
  Top row: Verbalized vs. internal confidence scatter plot, calibration error histogram, and confidence score distributions for the baseline model. 
  Bottom row: Same visualizations for the DCA-trained model.}
  \label{gemma_openbookqa}
\end{figure*}

\begin{figure*}[t]
  \centering

  \begin{subfigure}{0.3\textwidth}
    \includegraphics[width=\linewidth]{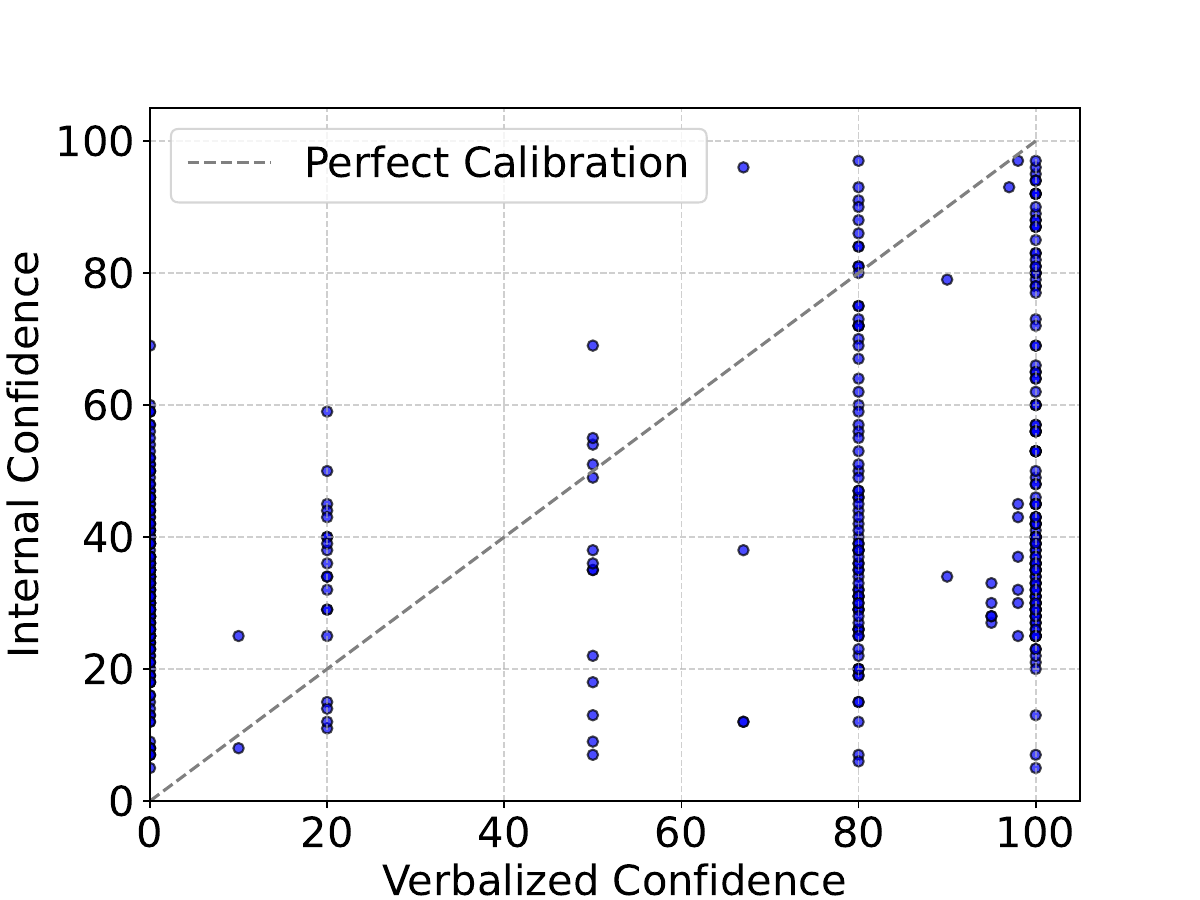}
    \caption{Scatter (Baseline)}
  \end{subfigure}
  \hfill
  \begin{subfigure}{0.3\textwidth}
    \includegraphics[width=\linewidth]{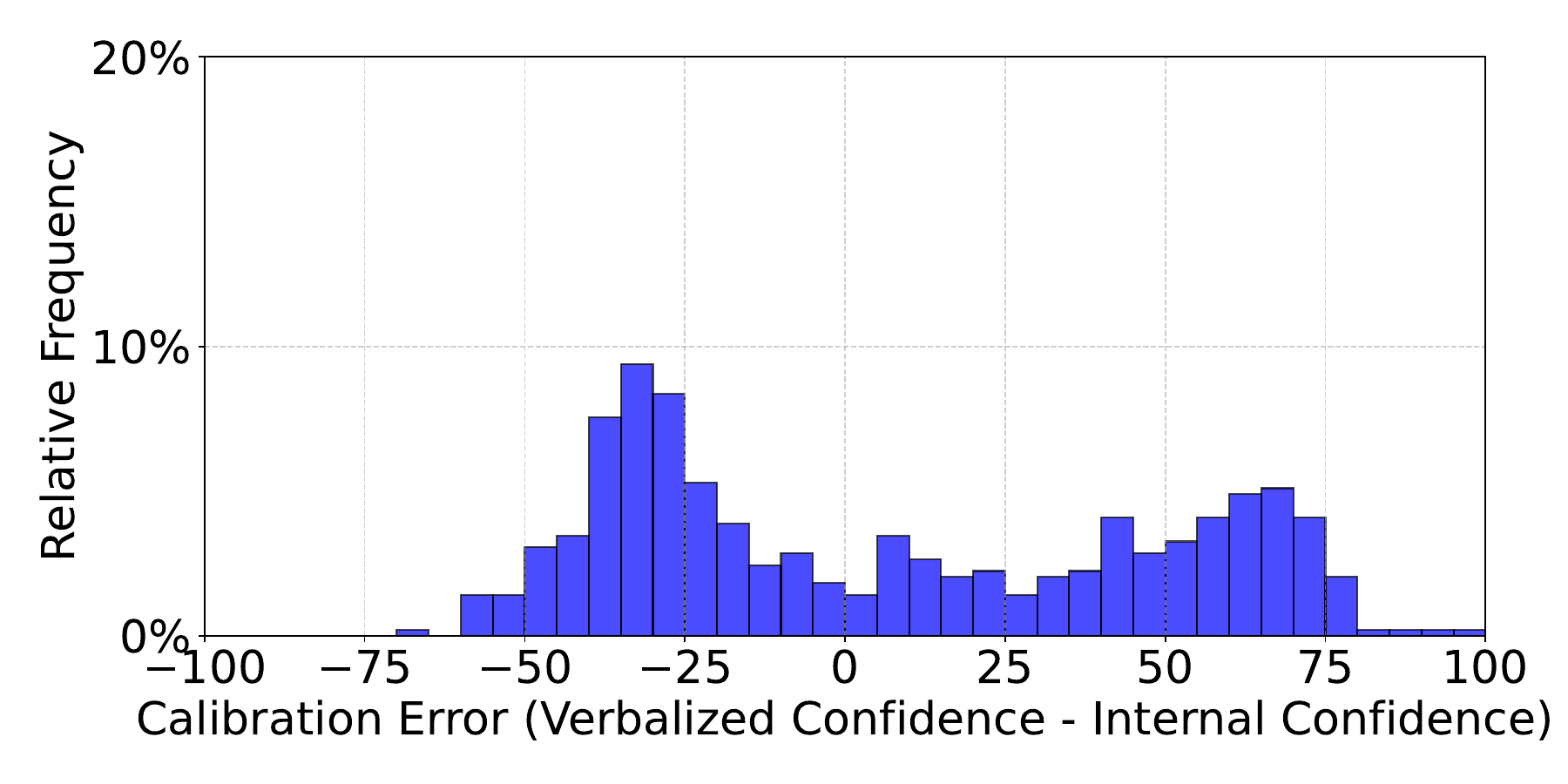}
    \caption{Calibration Error (Baseline)}
  \end{subfigure}
  \hfill
  \begin{subfigure}{0.3\textwidth}
    \includegraphics[width=\linewidth]{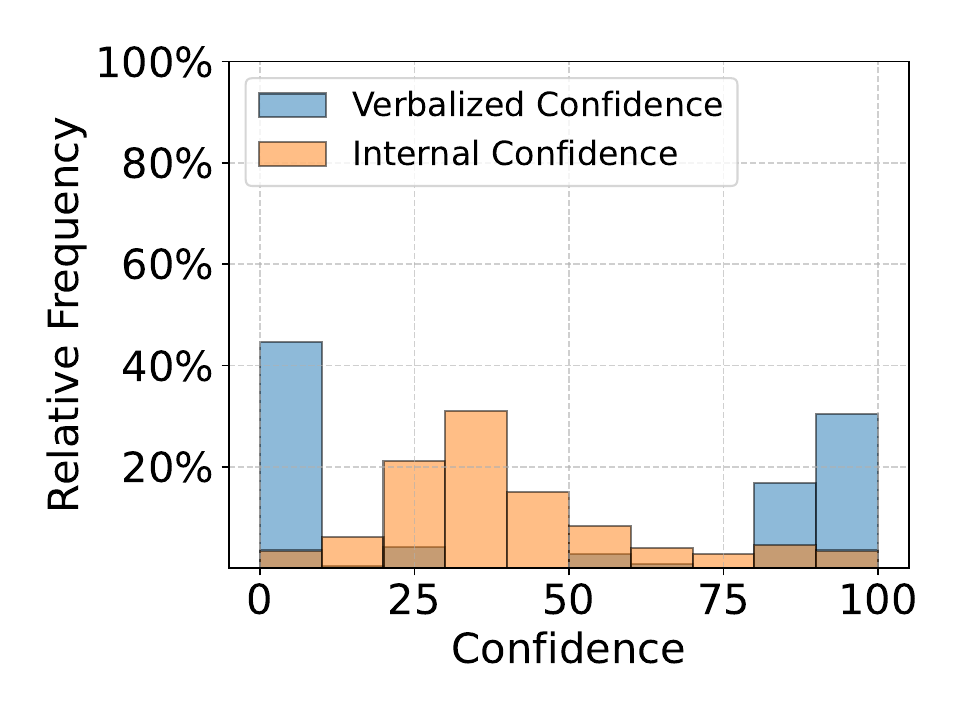}
    \caption{Distributions (Baseline)}
  \end{subfigure}

  \vspace{0.5cm}

  \begin{subfigure}{0.3\textwidth}
    \includegraphics[width=\linewidth]{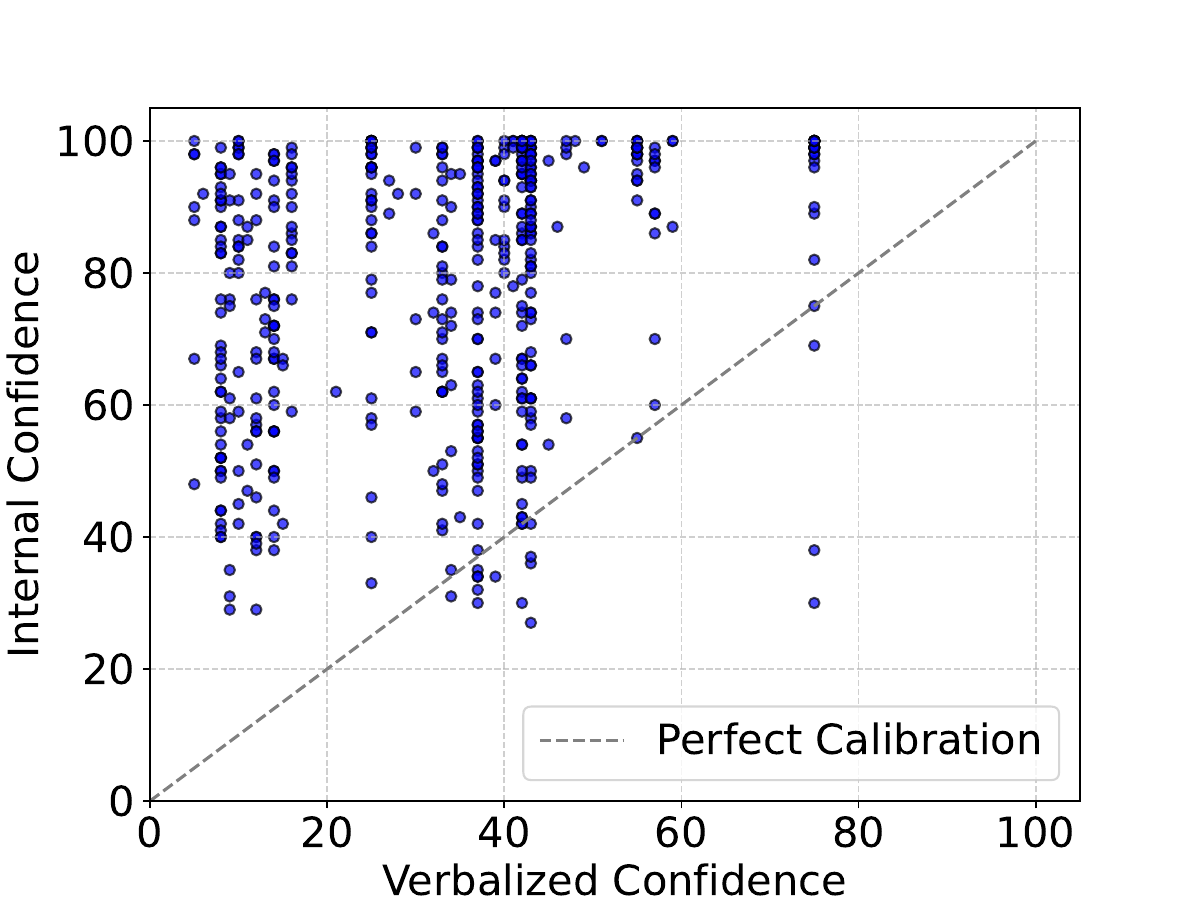}
    \caption{Scatter (DCA)}
  \end{subfigure}
  \hfill
  \begin{subfigure}{0.3\textwidth}
    \includegraphics[width=\linewidth]{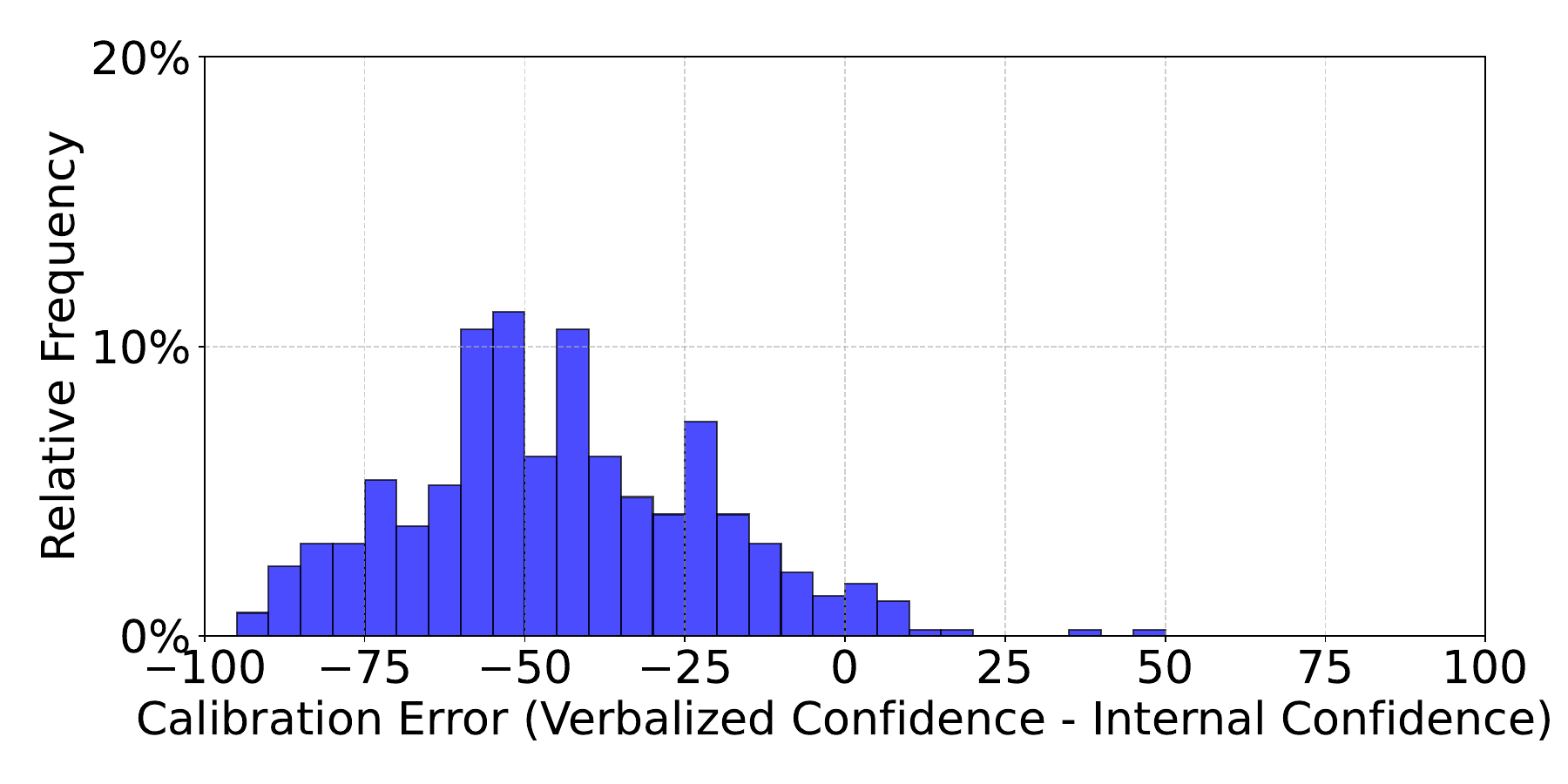}
    \caption{Calibration Error (DCA)}
  \end{subfigure}
  \hfill
  \begin{subfigure}{0.3\textwidth}
    \includegraphics[width=\linewidth]{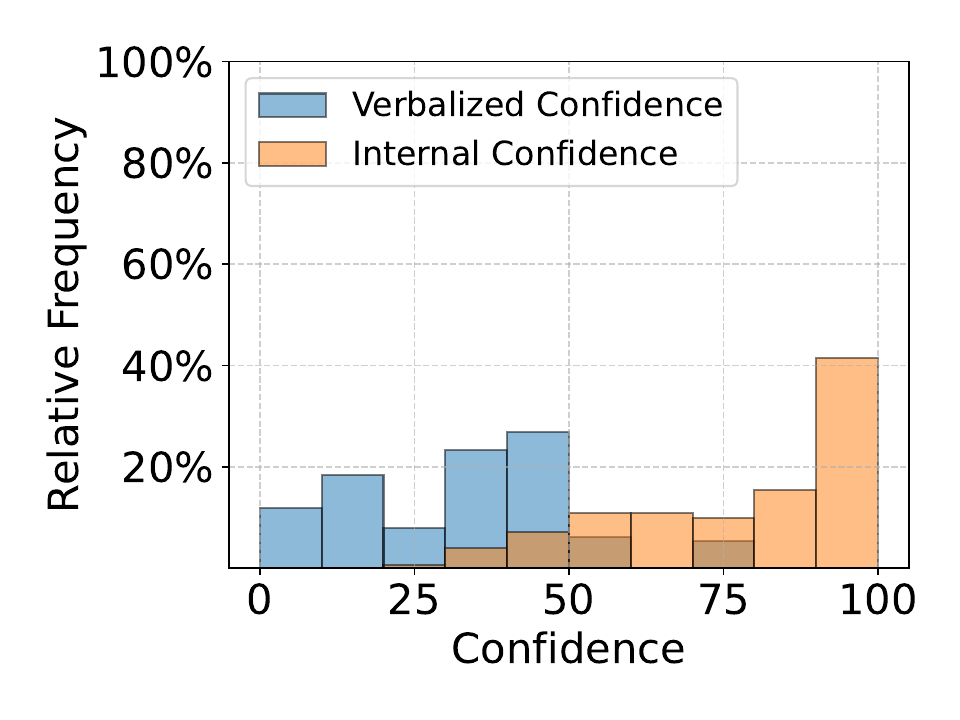}
    \caption{Distributions (DCA)}
  \end{subfigure}

  \caption{Comparison of baseline vs. DCA-trained Llama-3.2-3B-Instruct on OpenBookQA. 
  Top row: Verbalized vs. internal confidence scatter plot, calibration error histogram, and confidence score distributions for the baseline model. 
  Bottom row: Same visualizations for the DCA-trained model.}
  \label{llama_openbookqa}
\end{figure*}

\begin{figure*}[t]
  \centering

  \begin{subfigure}{0.3\textwidth}
    \includegraphics[width=\linewidth]{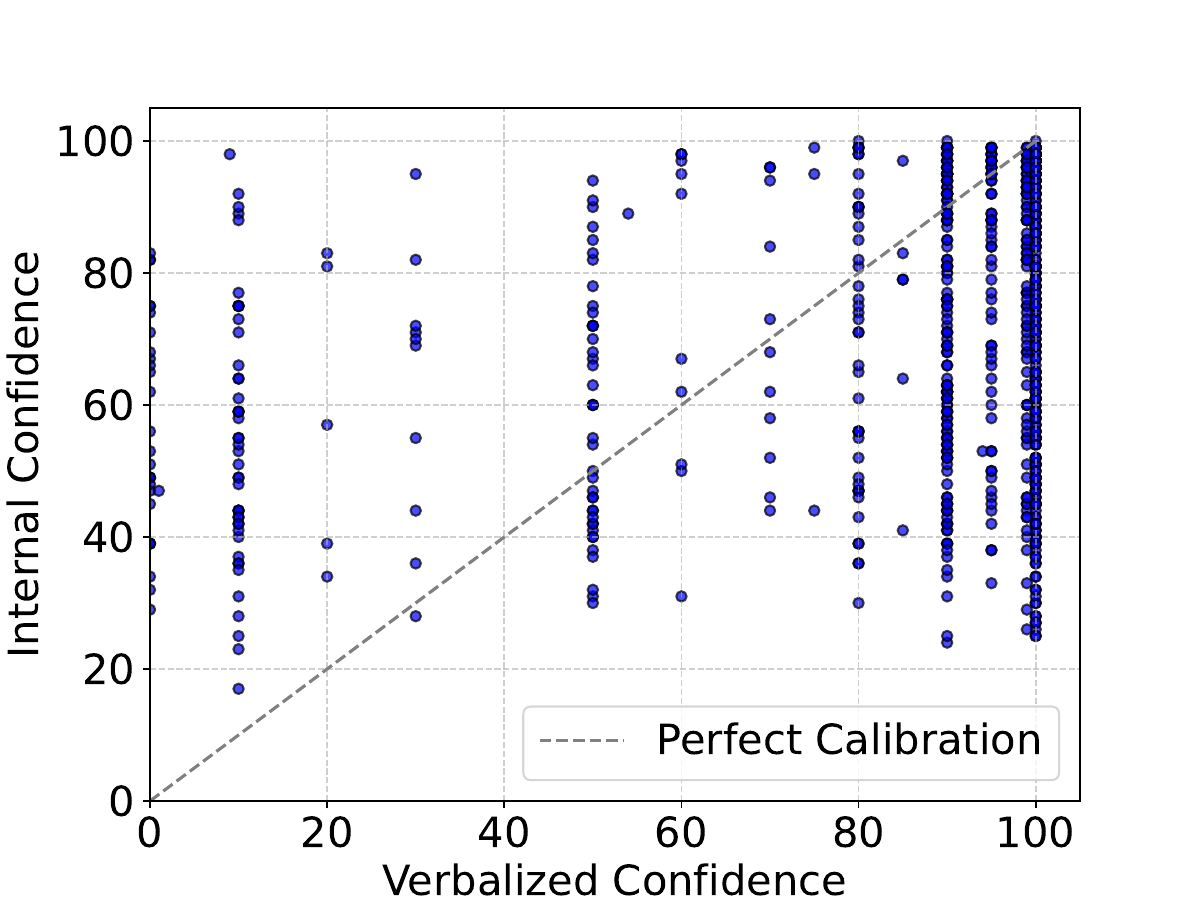}
    \caption{Scatter (Baseline)}
  \end{subfigure}
  \hfill
  \begin{subfigure}{0.3\textwidth}
    \includegraphics[width=\linewidth]{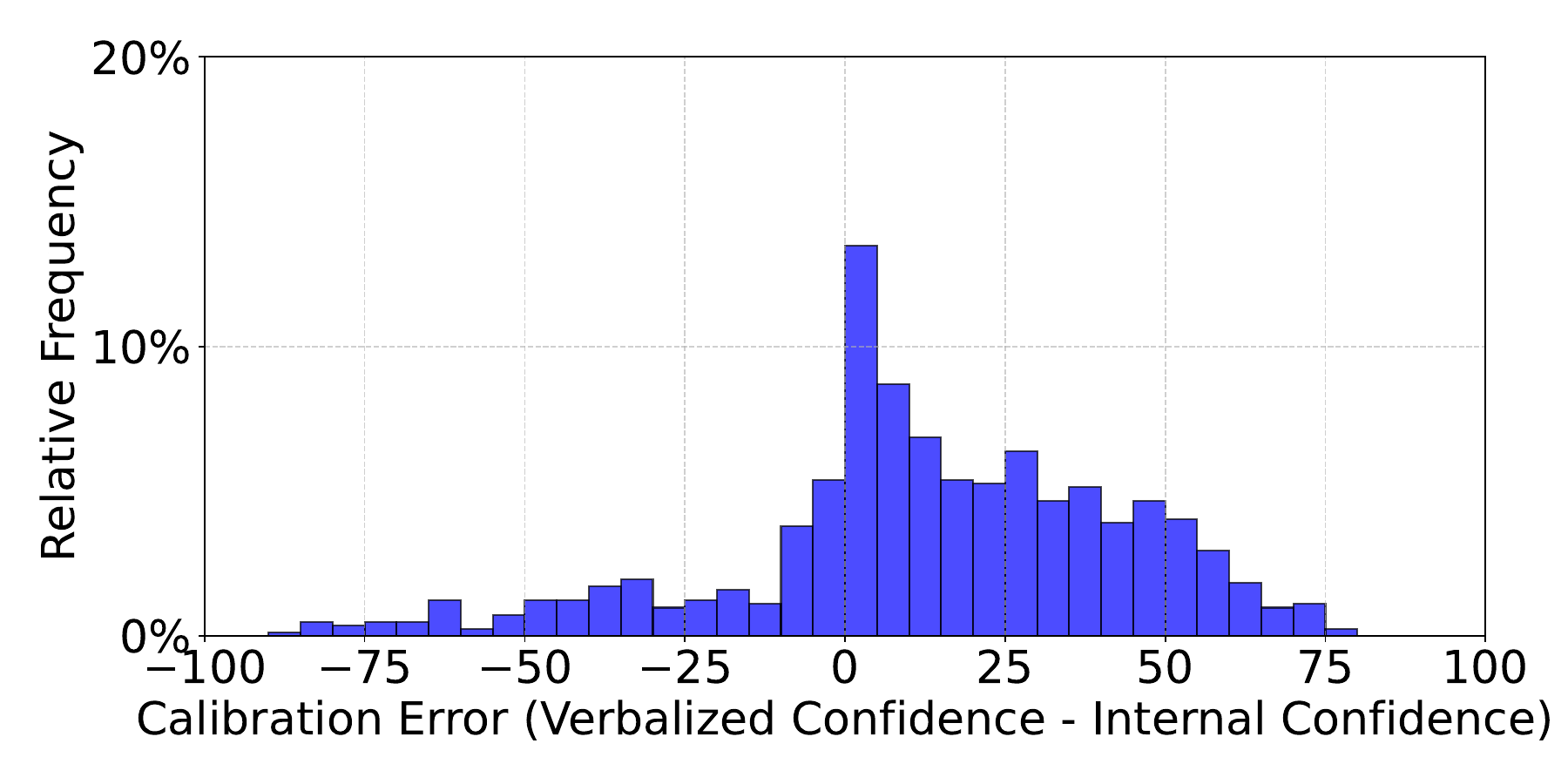}
    \caption{Calibration Error (Baseline)}
  \end{subfigure}
  \hfill
  \begin{subfigure}{0.3\textwidth}
    \includegraphics[width=\linewidth]{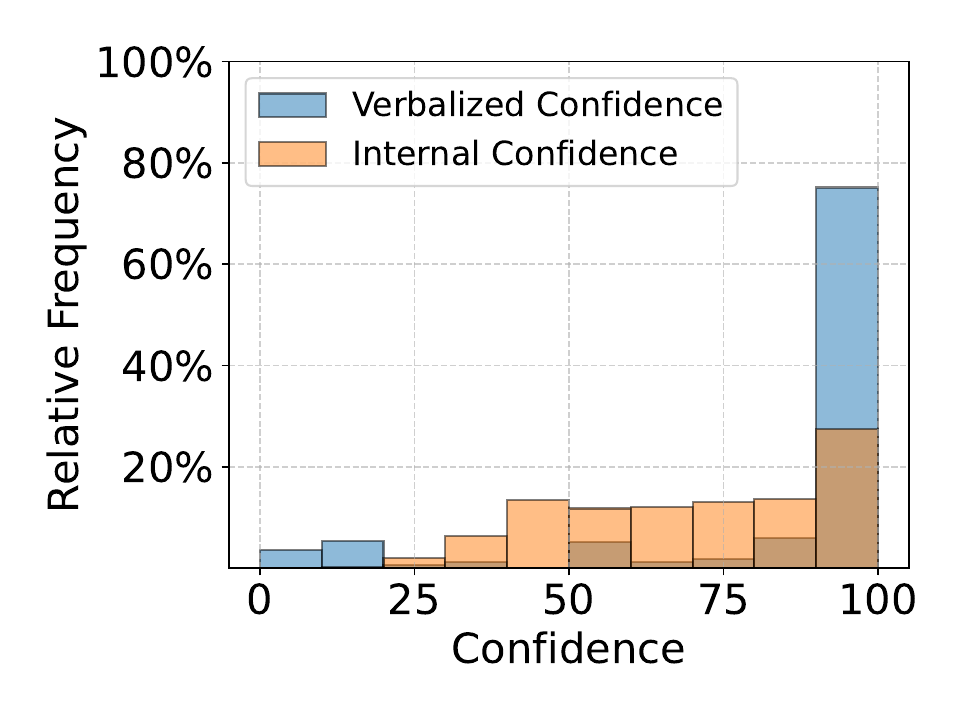}
    \caption{Distributions (Baseline)}
  \end{subfigure}

  \vspace{0.5cm}

  \begin{subfigure}{0.3\textwidth}
    \includegraphics[width=\linewidth]{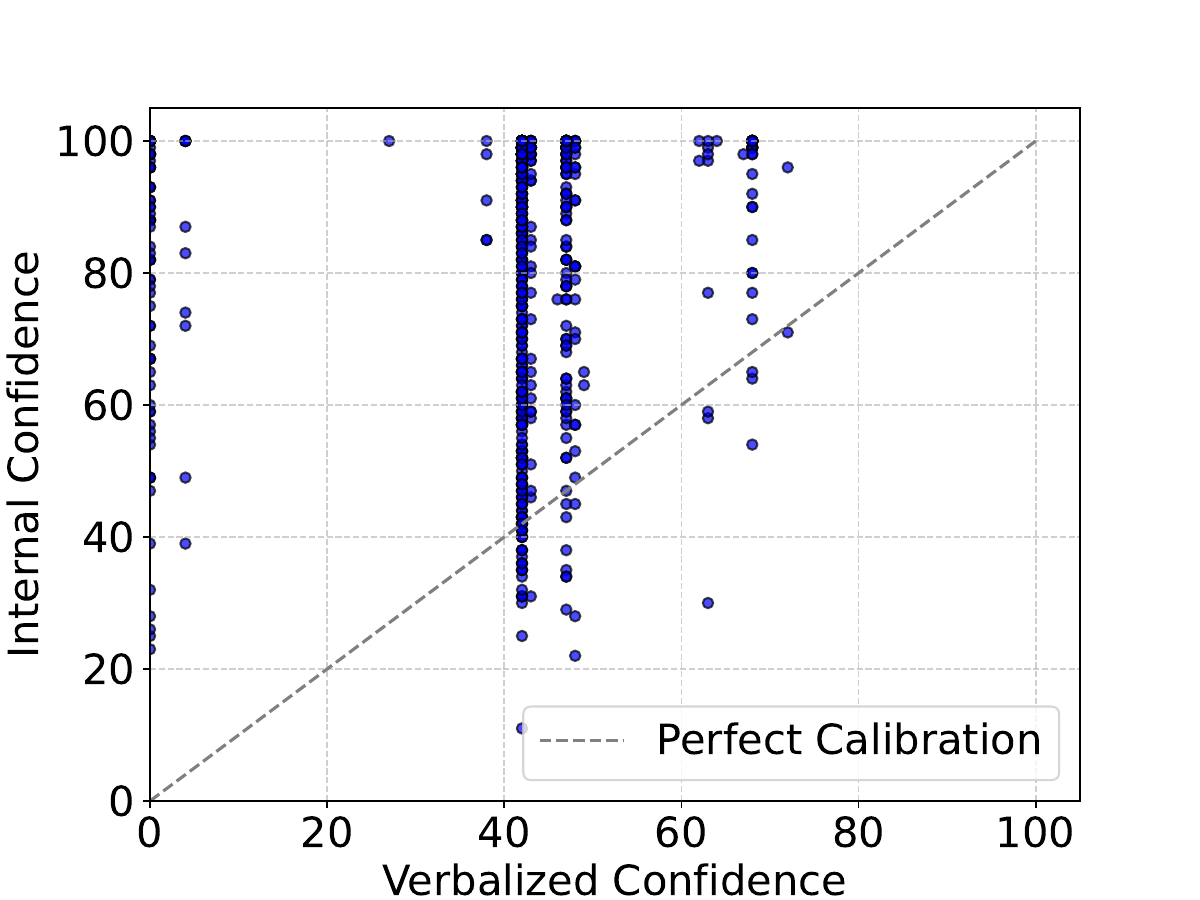}
    \caption{Scatter (DCA)}
  \end{subfigure}
  \hfill
  \begin{subfigure}{0.3\textwidth}
    \includegraphics[width=\linewidth]{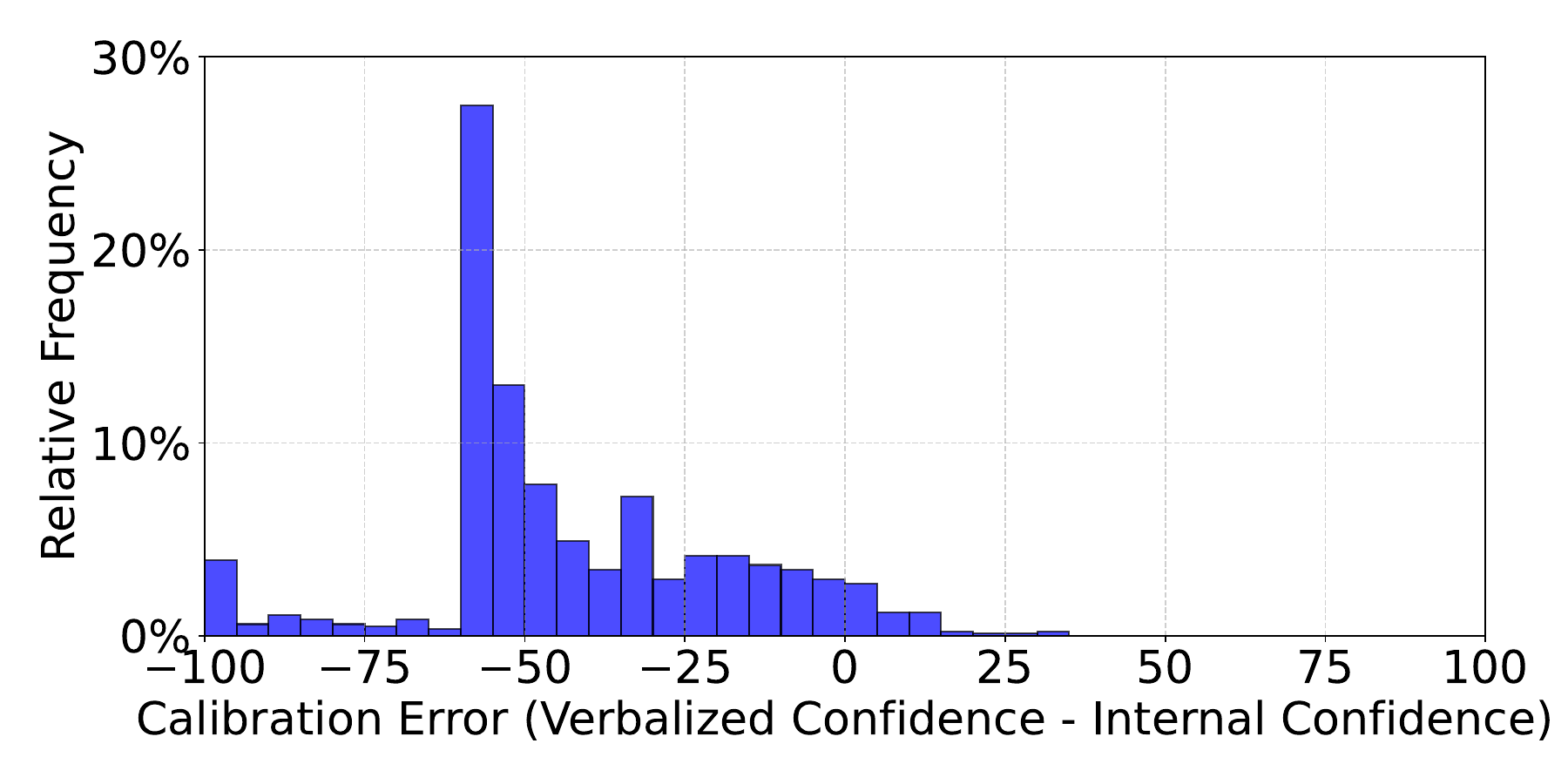}
    \caption{Calibration Error (DCA)}
  \end{subfigure}
  \hfill
  \begin{subfigure}{0.3\textwidth}
    \includegraphics[width=\linewidth]{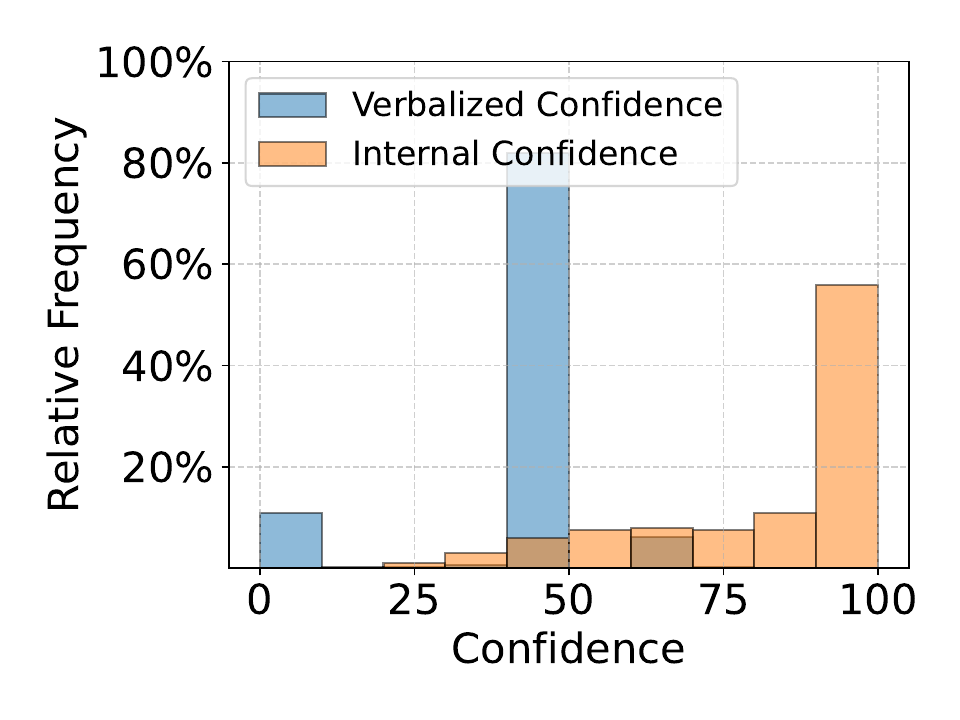}
    \caption{Distributions (DCA)}
  \end{subfigure}

  \caption{Comparison of baseline vs. DCA-trained Mistral-7B-Instruct on TruthfulQA. 
  Top row: Verbalized vs. internal confidence scatter plot, calibration error histogram, and confidence score distributions for the baseline model. 
  Bottom row: Same visualizations for the DCA-trained model.}
  \label{mistral_truthfulqa}
\end{figure*}

\begin{figure*}[t]
  \centering

  \begin{subfigure}{0.3\textwidth}
    \includegraphics[width=\linewidth]{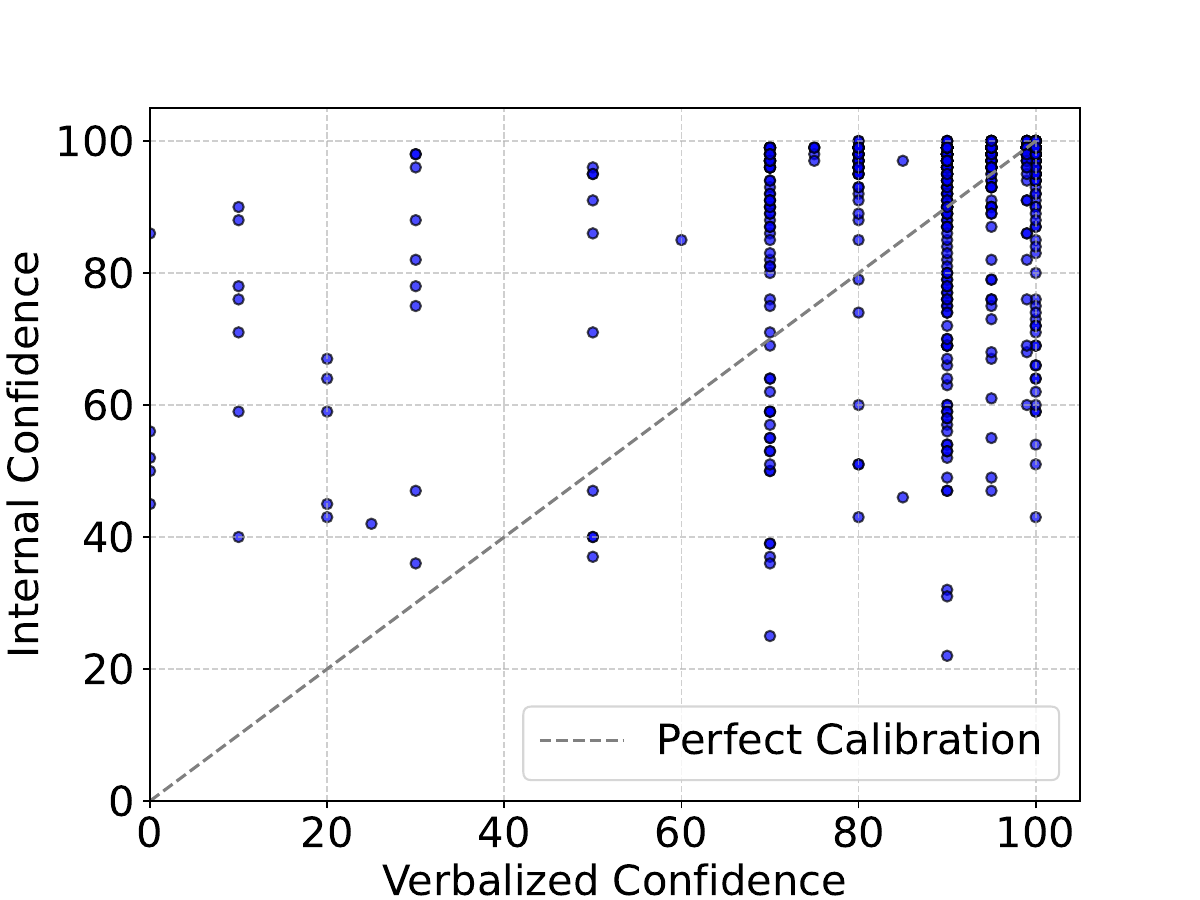}
    \caption{Scatter (Baseline)}
  \end{subfigure}
  \hfill
  \begin{subfigure}{0.3\textwidth}
    \includegraphics[width=\linewidth]{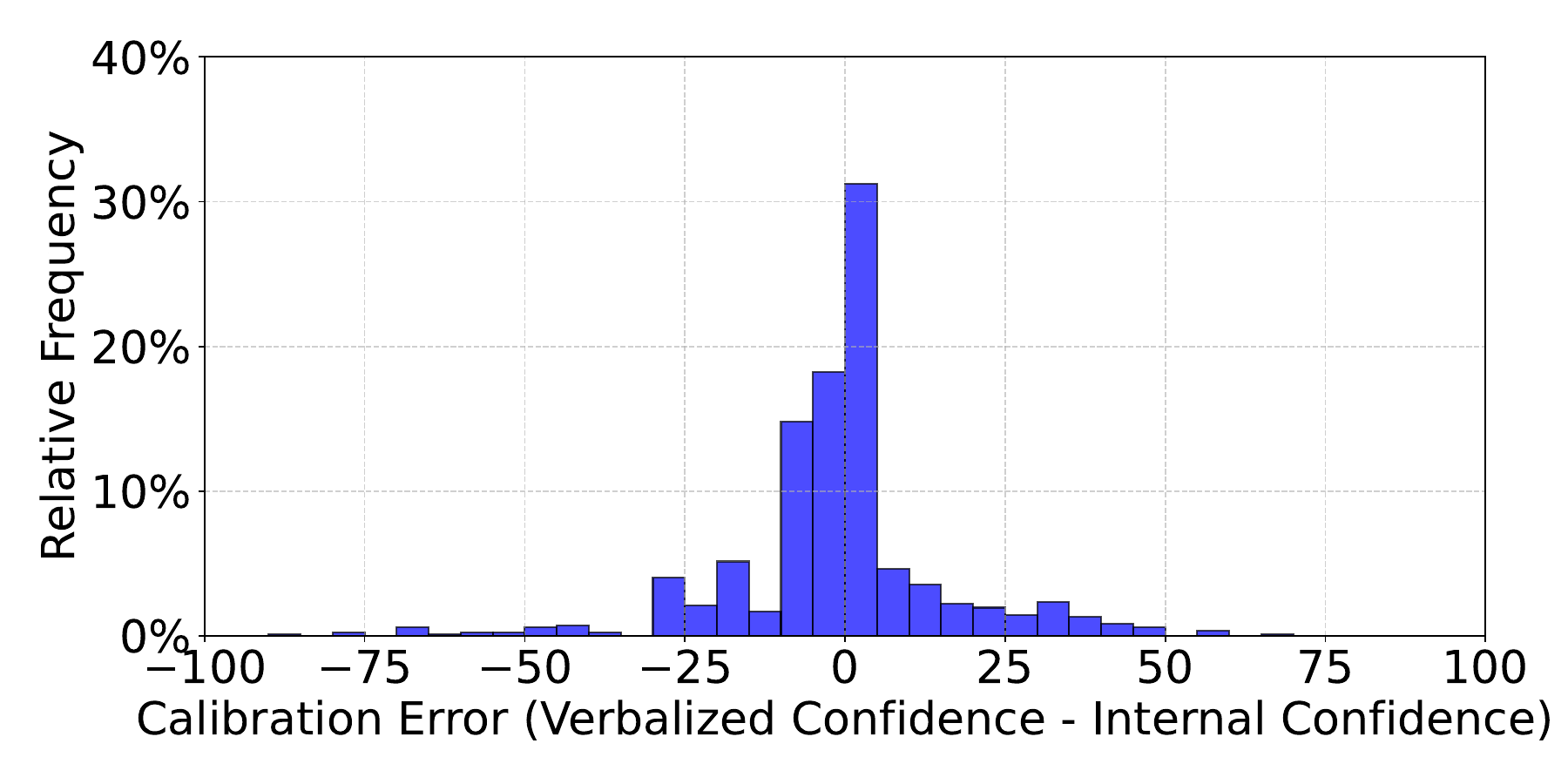}
    \caption{Calibration Error (Baseline)}
  \end{subfigure}
  \hfill
  \begin{subfigure}{0.3\textwidth}
    \includegraphics[width=\linewidth]{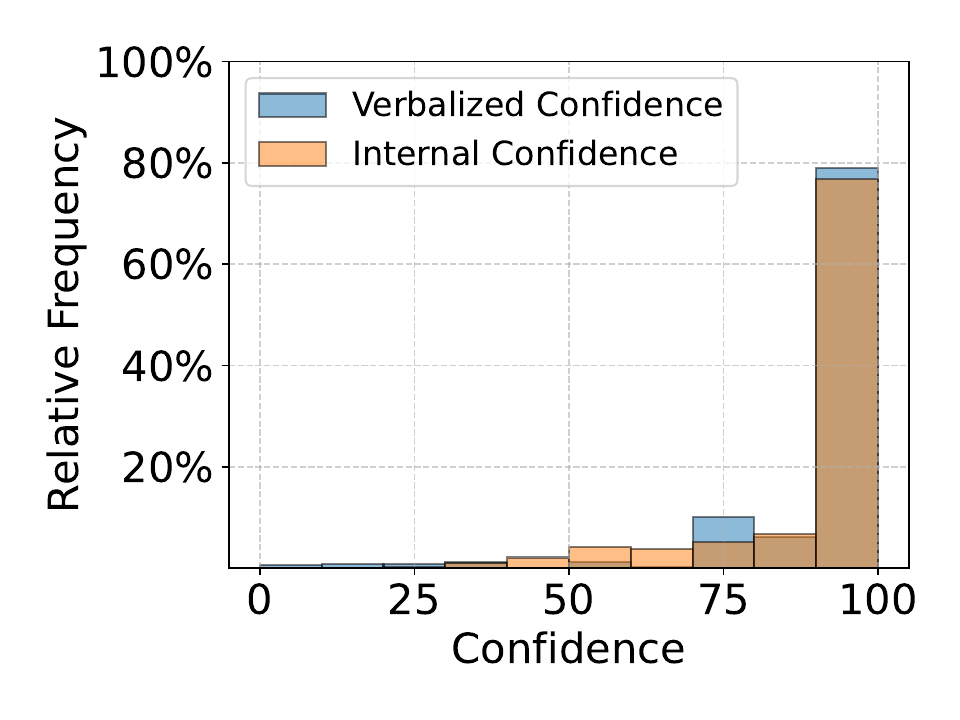}
    \caption{Distributions (Baseline)}
  \end{subfigure}

  \vspace{0.5cm}

  \begin{subfigure}{0.3\textwidth}
    \includegraphics[width=\linewidth]{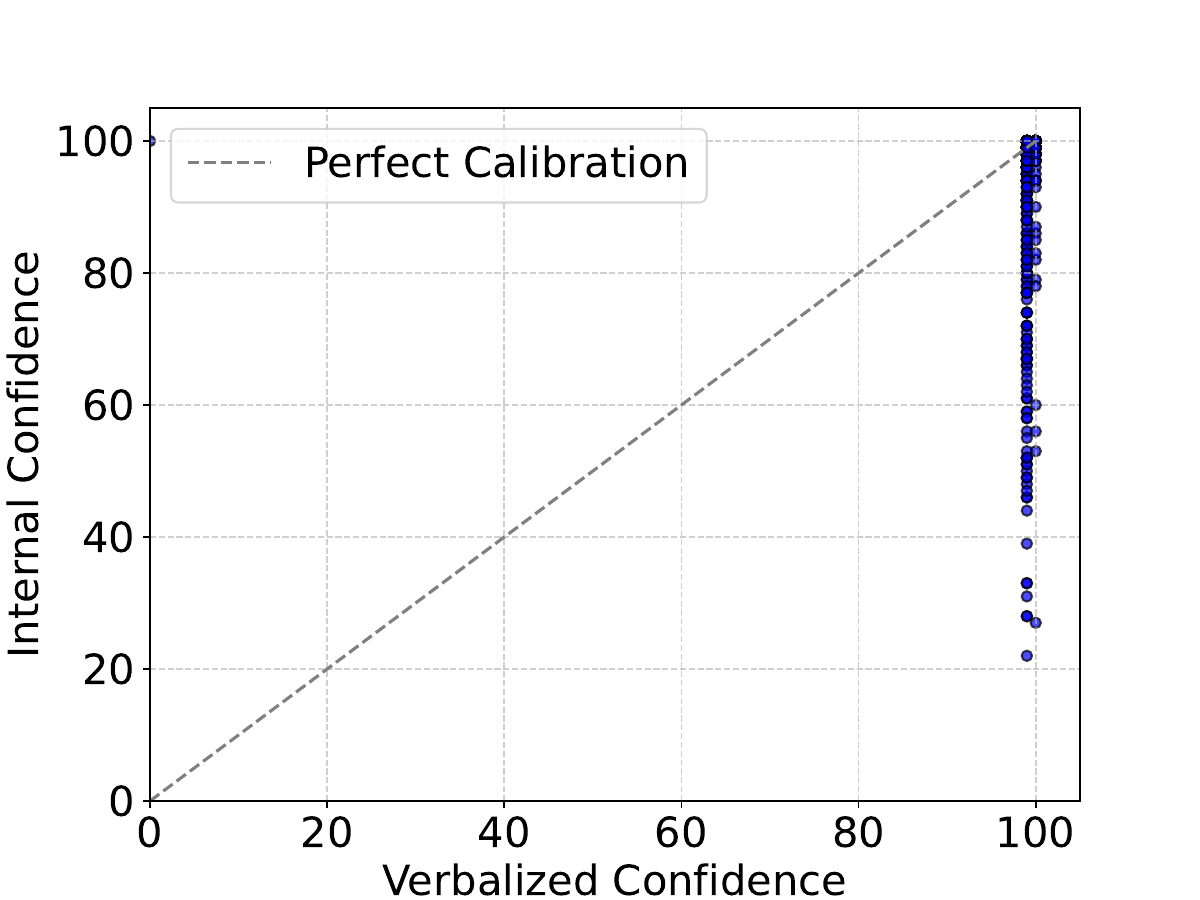}
    \caption{Scatter (DCA)}
  \end{subfigure}
  \hfill
  \begin{subfigure}{0.3\textwidth}
    \includegraphics[width=\linewidth]{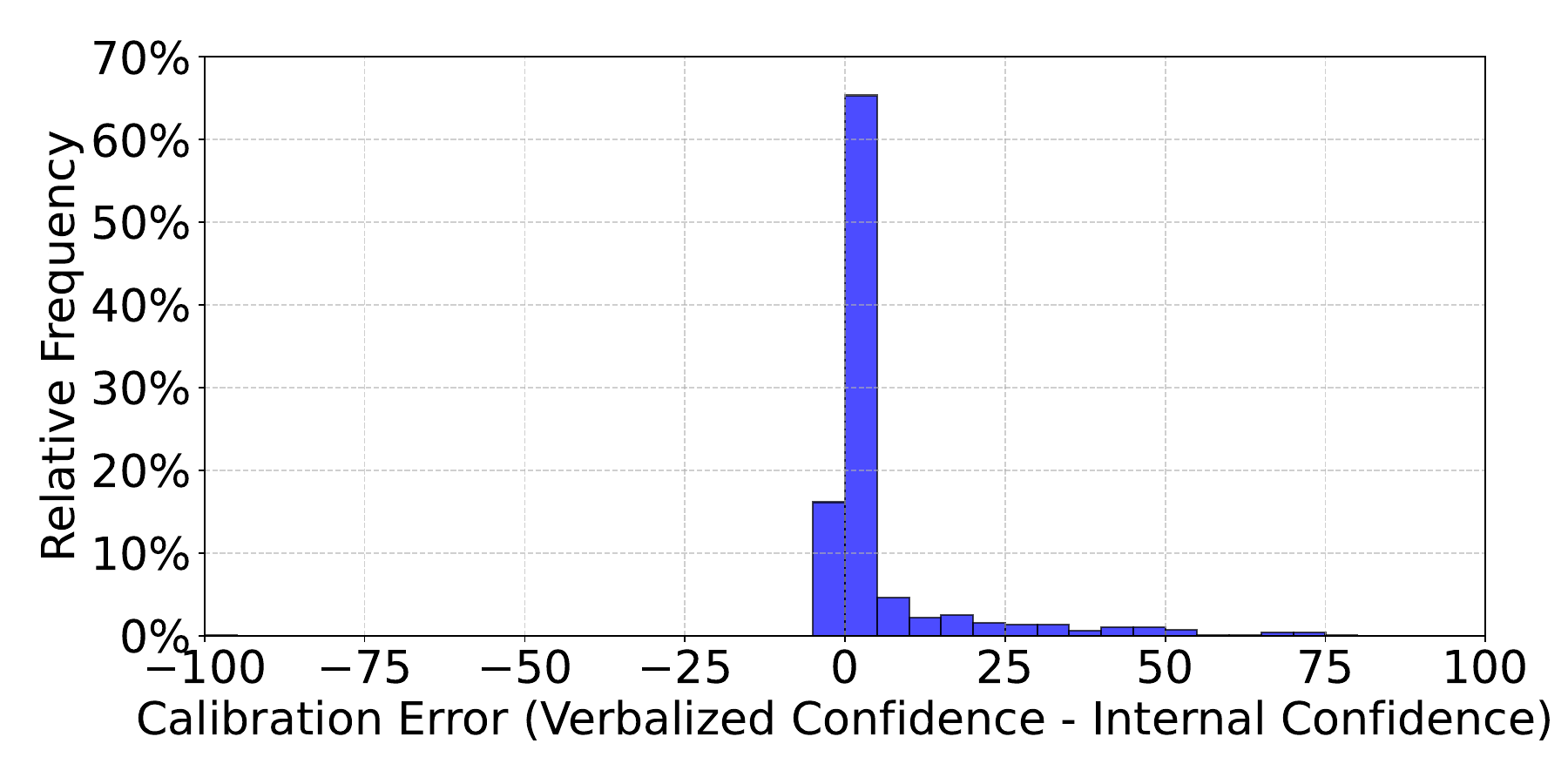}
    \caption{Calibration Error (DCA)}
  \end{subfigure}
  \hfill
  \begin{subfigure}{0.3\textwidth}
    \includegraphics[width=\linewidth]{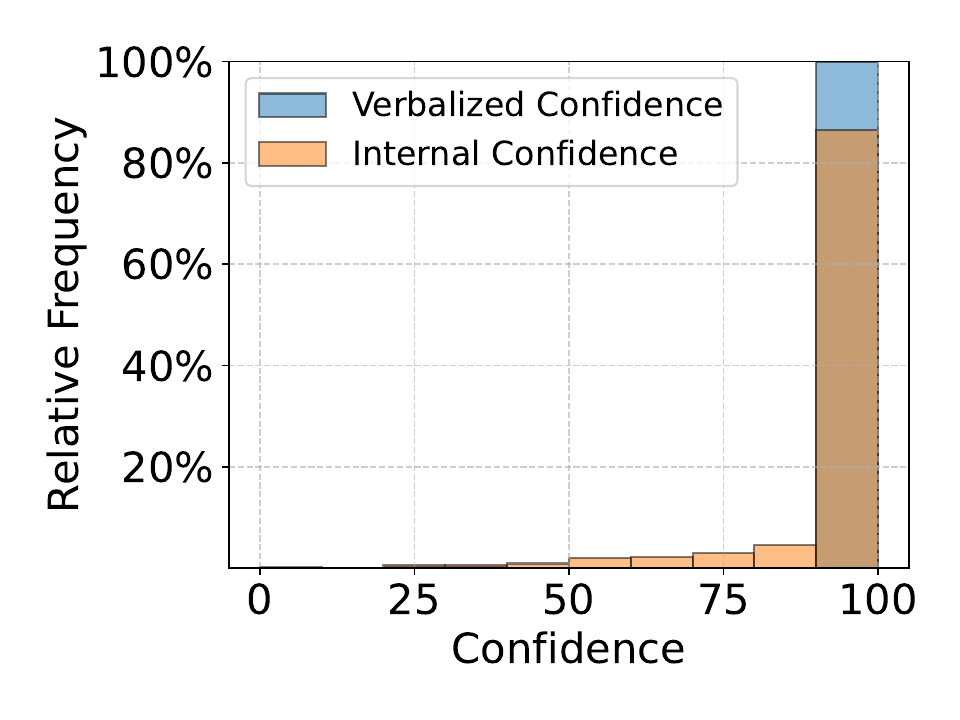}
    \caption{Distributions (DCA)}
  \end{subfigure}

  \caption{Comparison of baseline vs. DCA-trained Gemma-2-9B-Instruct on TruthfulQA. 
  Top row: Verbalized vs. internal confidence scatter plot, calibration error histogram, and confidence score distributions for the baseline model. 
  Bottom row: Same visualizations for the DCA-trained model.}
  \label{gemma_truthfulqa}
\end{figure*}

\begin{figure*}[t]
  \centering

  \begin{subfigure}{0.3\textwidth}
    \includegraphics[width=\linewidth]{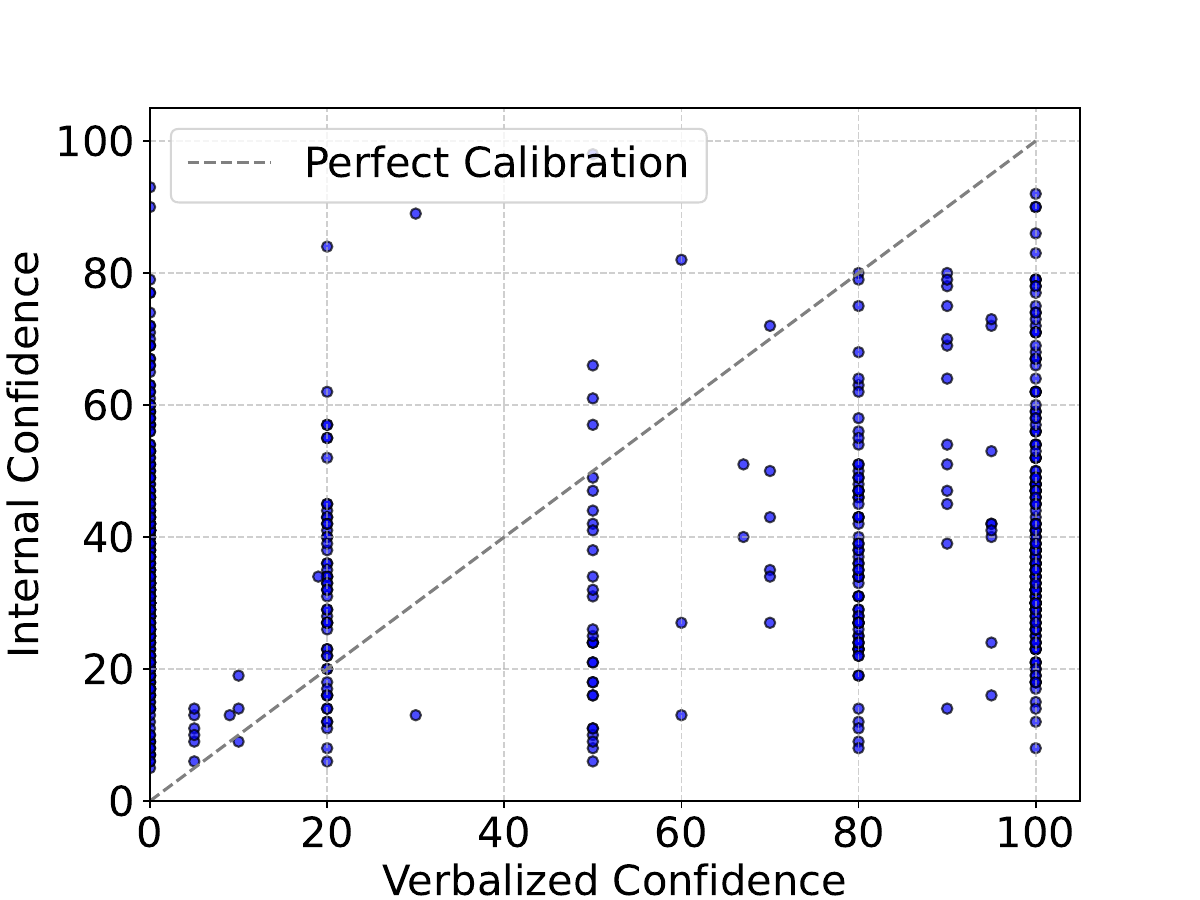}
    \caption{Scatter (Baseline)}
  \end{subfigure}
  \hfill
  \begin{subfigure}{0.3\textwidth}
    \includegraphics[width=\linewidth]{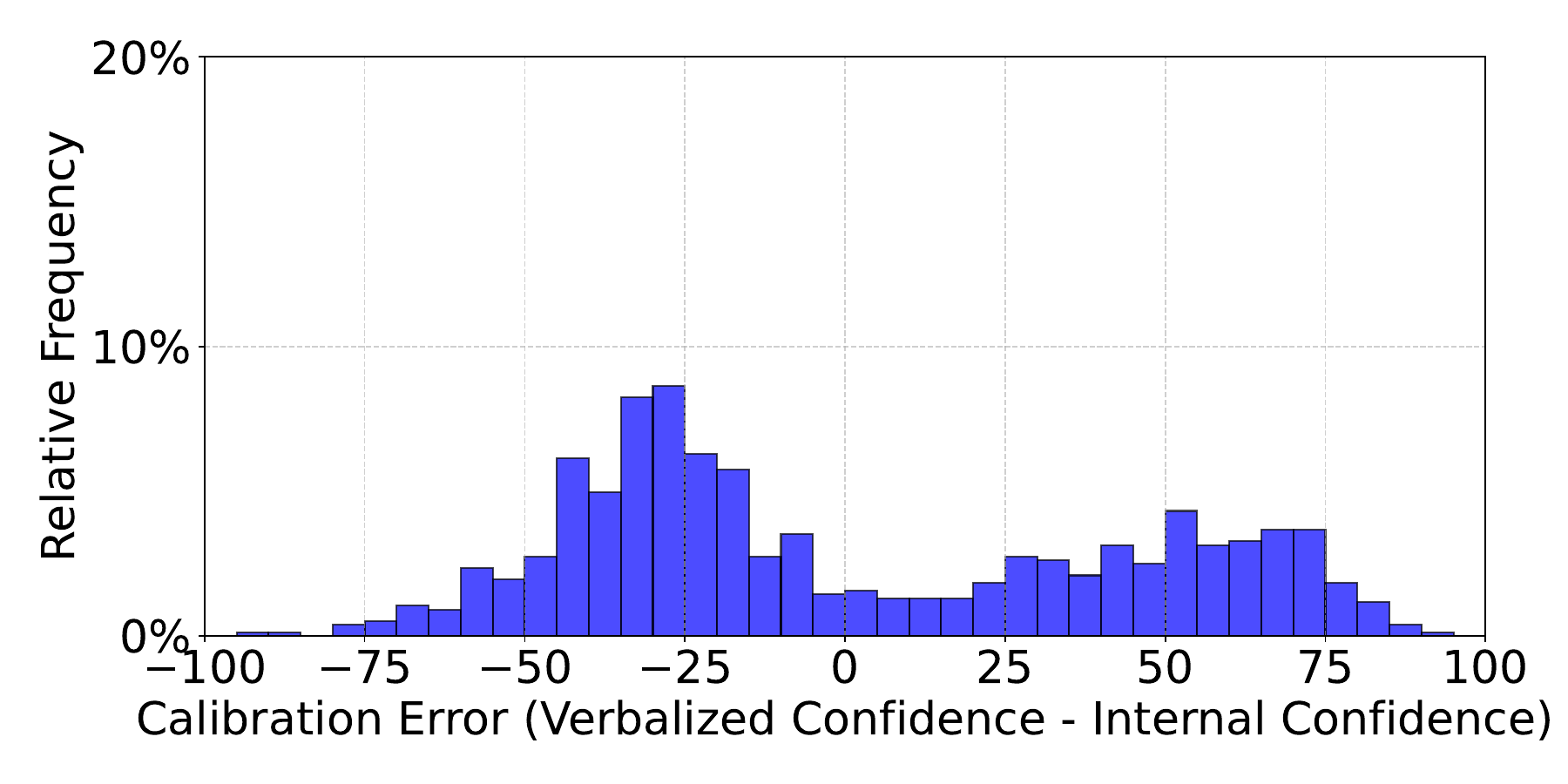}
    \caption{Calibration Error (Baseline)}
  \end{subfigure}
  \hfill
  \begin{subfigure}{0.3\textwidth}
    \includegraphics[width=\linewidth]{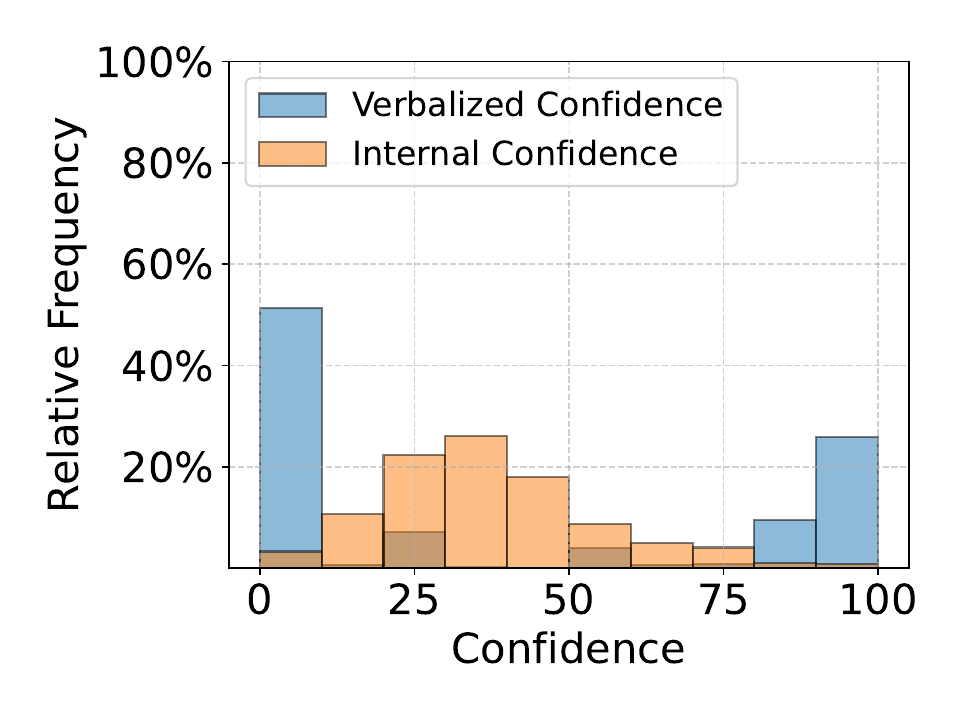}
    \caption{Distributions (Baseline)}
  \end{subfigure}

  \vspace{0.5cm}

  \begin{subfigure}{0.3\textwidth}
    \includegraphics[width=\linewidth]{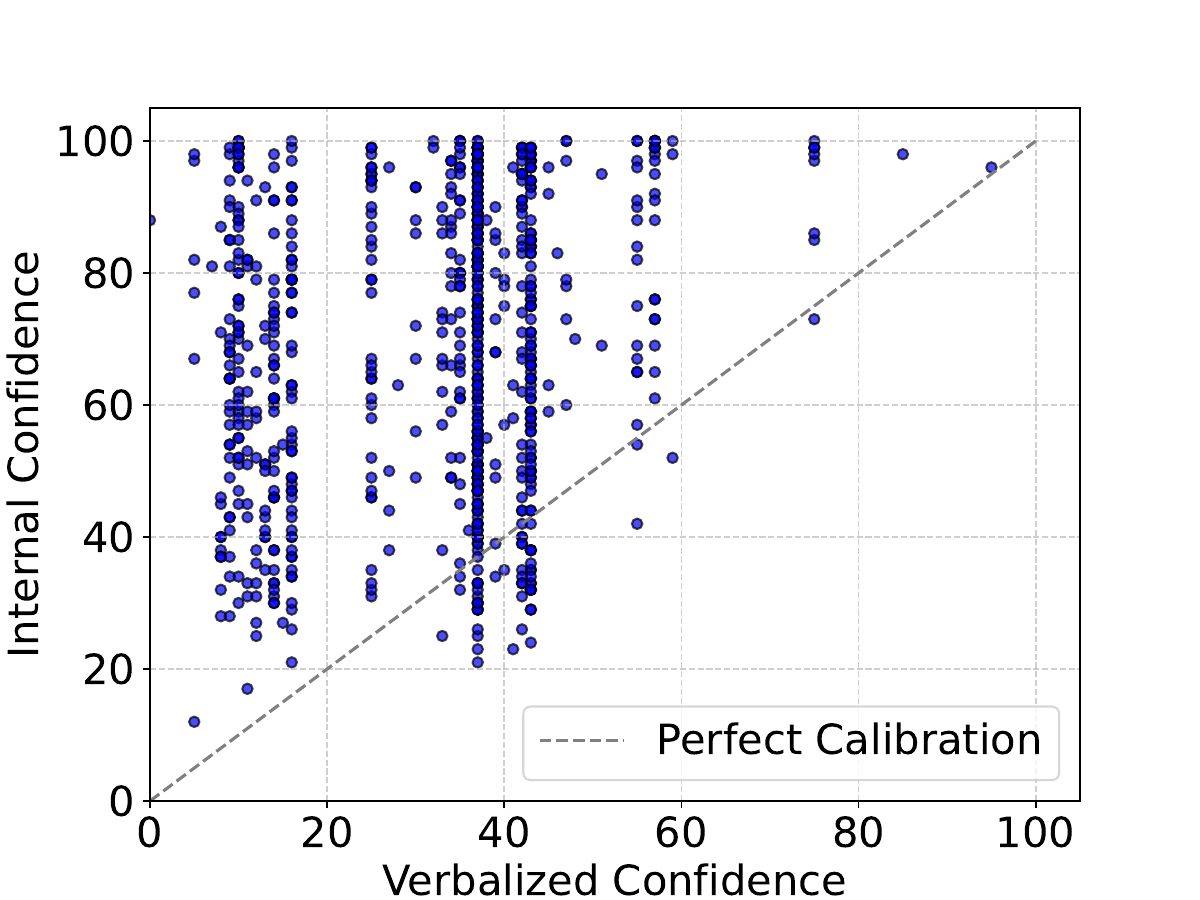}
    \caption{Scatter (DCA)}
  \end{subfigure}
  \hfill
  \begin{subfigure}{0.3\textwidth}
    \includegraphics[width=\linewidth]{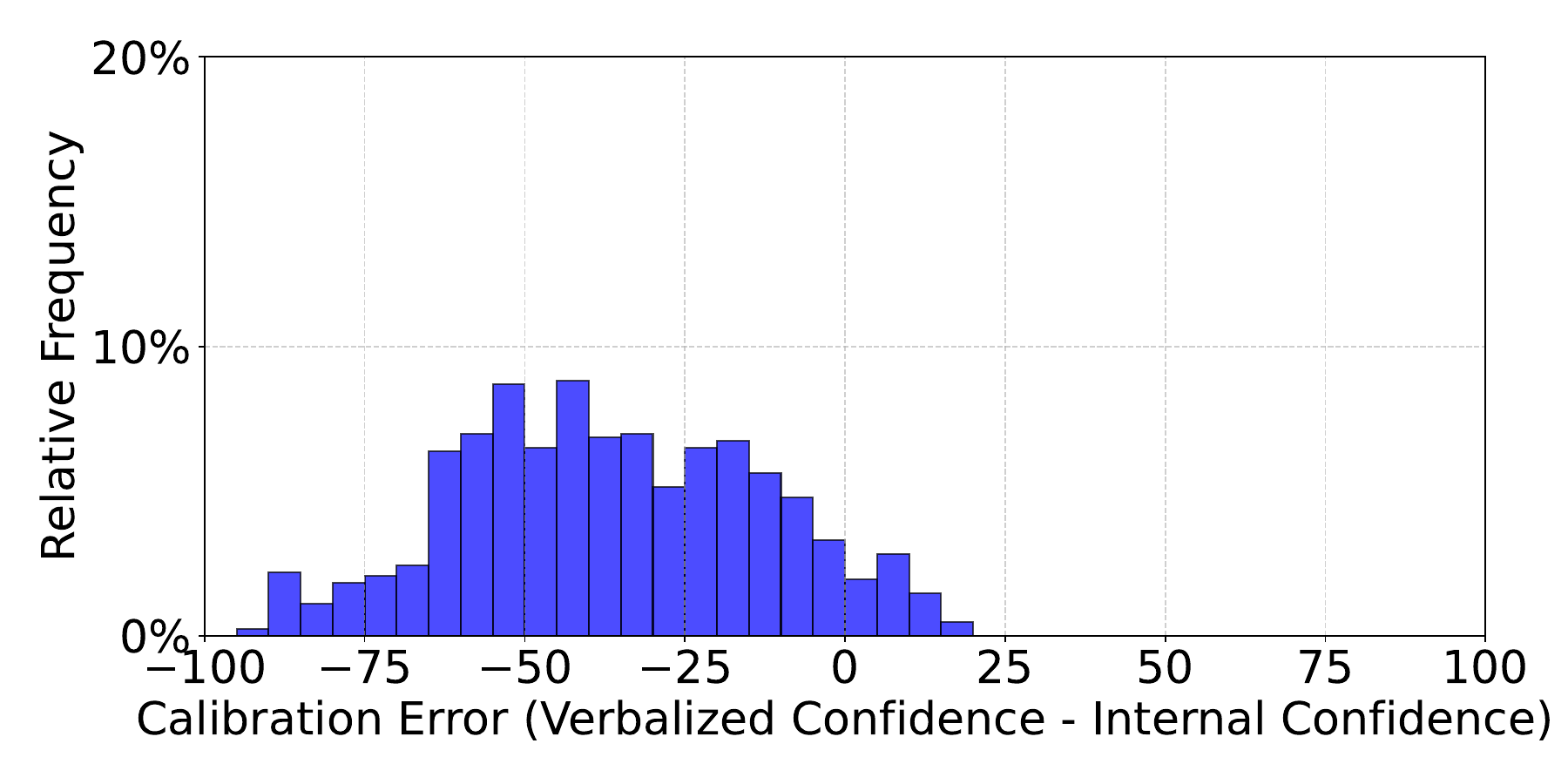}
    \caption{Calibration Error (DCA)}
  \end{subfigure}
  \hfill
  \begin{subfigure}{0.3\textwidth}
    \includegraphics[width=\linewidth]{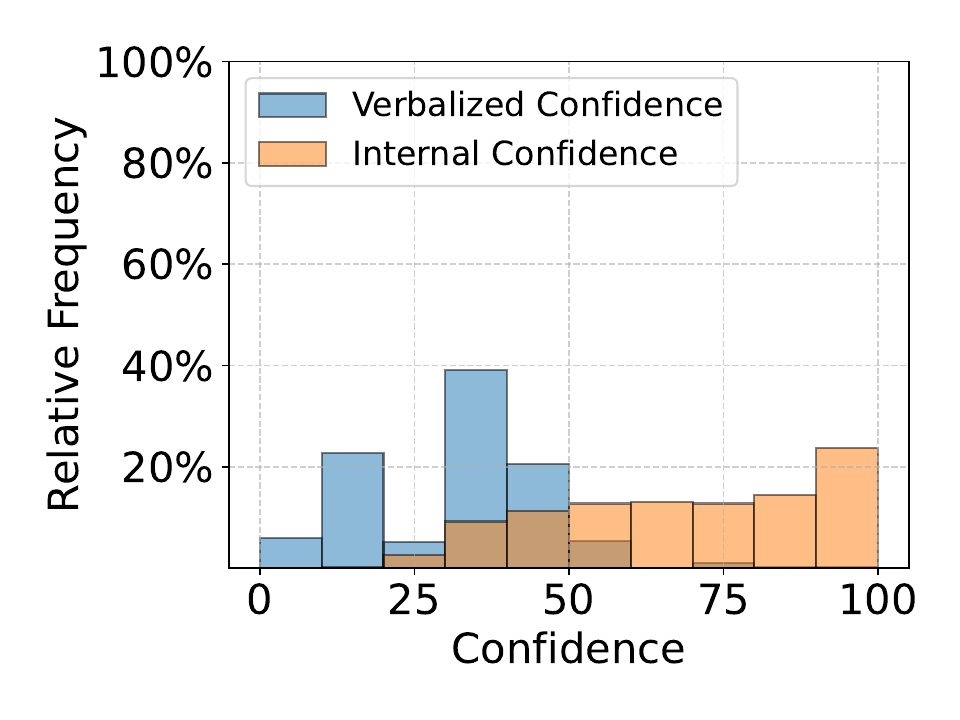}
    \caption{Distributions (DCA)}
  \end{subfigure}

  \caption{Comparison of baseline vs. DCA-trained Llama-3.2-3B-Instruct on TruthfulQA. 
  Top row: Verbalized vs. internal confidence scatter plot, calibration error histogram, and confidence score distributions for the baseline model. 
  Bottom row: Same visualizations for the DCA-trained model.}
  \label{llama_truthfulqa}
\end{figure*}

\begin{figure*}[t]
  \centering

  \begin{subfigure}{0.3\textwidth}
    \includegraphics[width=\linewidth]{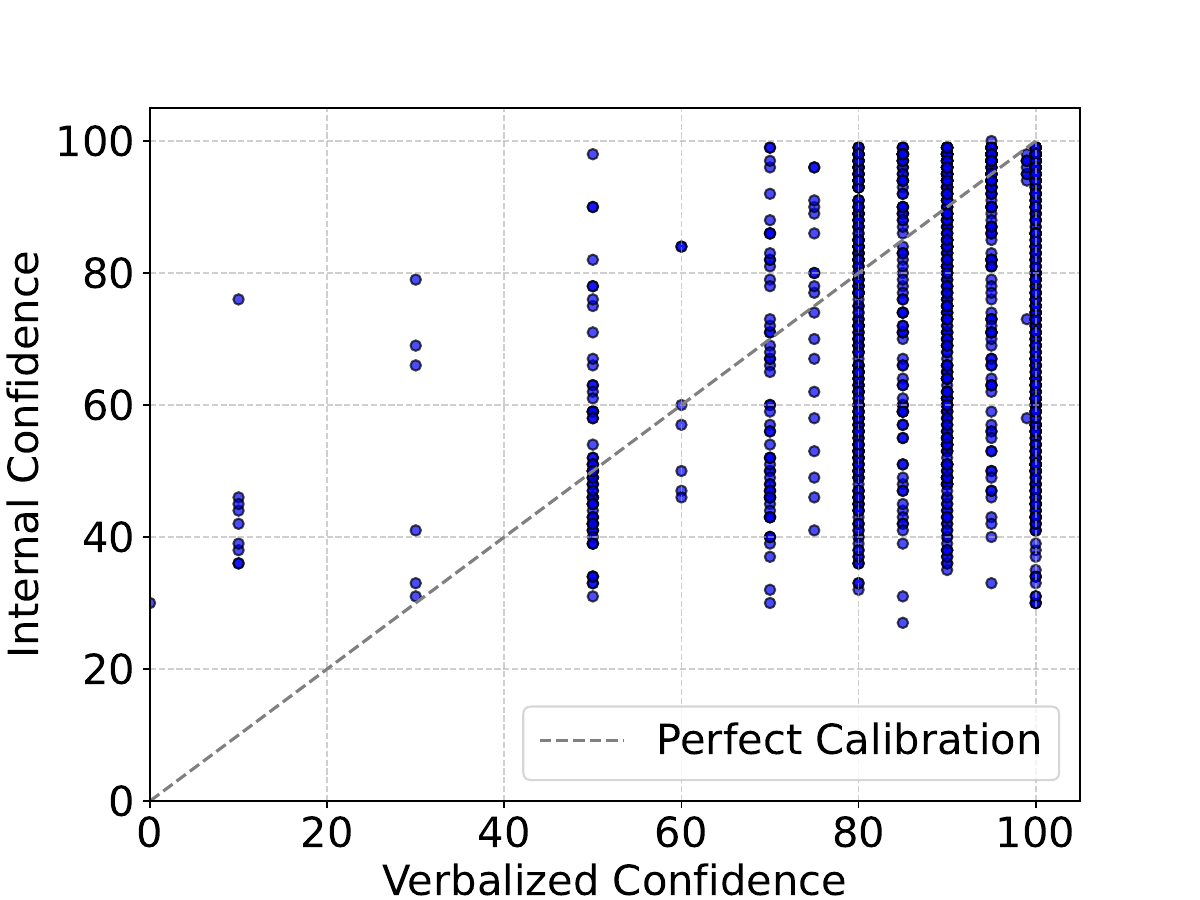}
    \caption{Scatter (Baseline)}
  \end{subfigure}
  \hfill
  \begin{subfigure}{0.3\textwidth}
    \includegraphics[width=\linewidth]{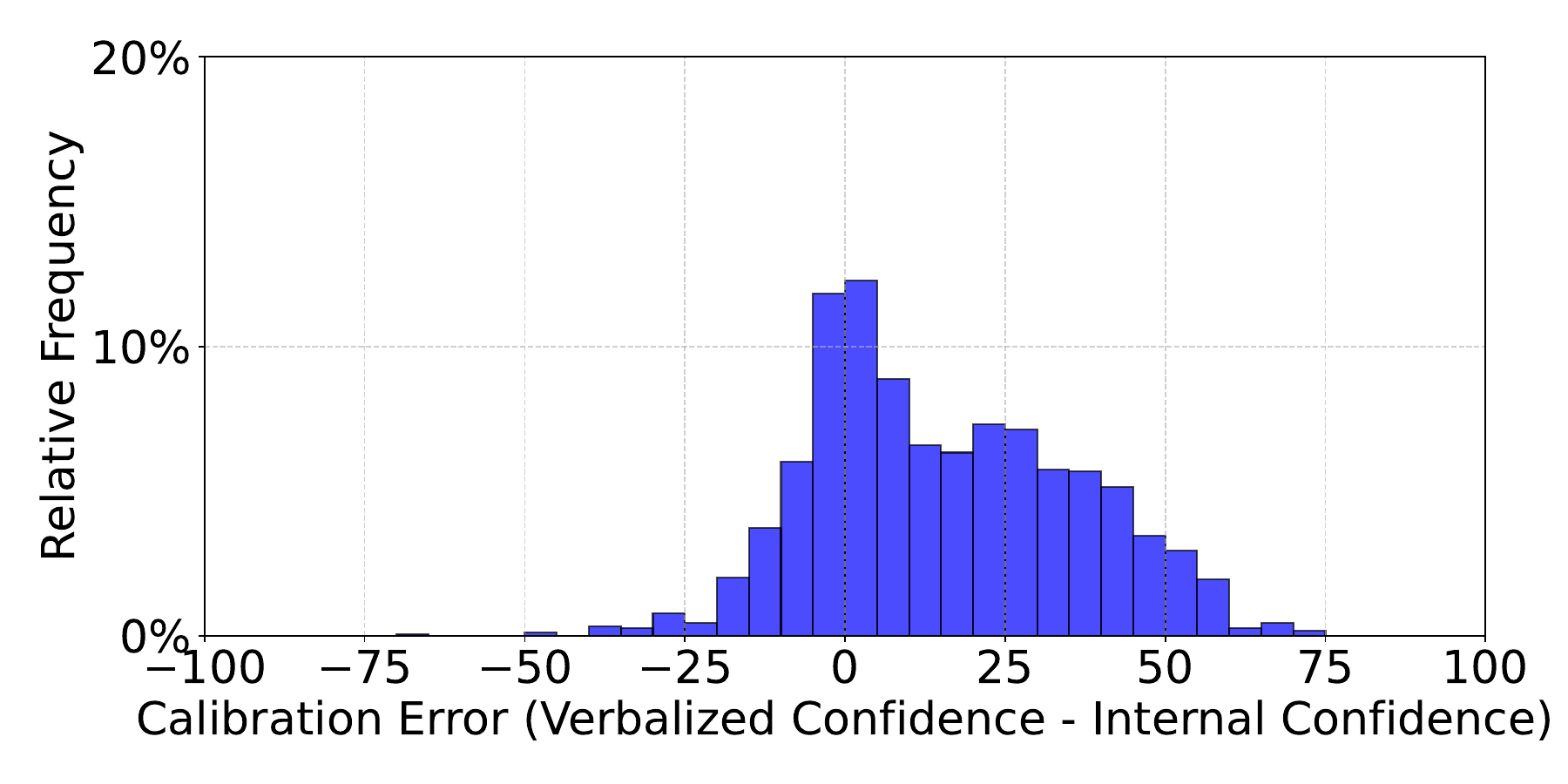}
    \caption{Calibration Error (Baseline)}
  \end{subfigure}
  \hfill
  \begin{subfigure}{0.3\textwidth}
    \includegraphics[width=\linewidth]{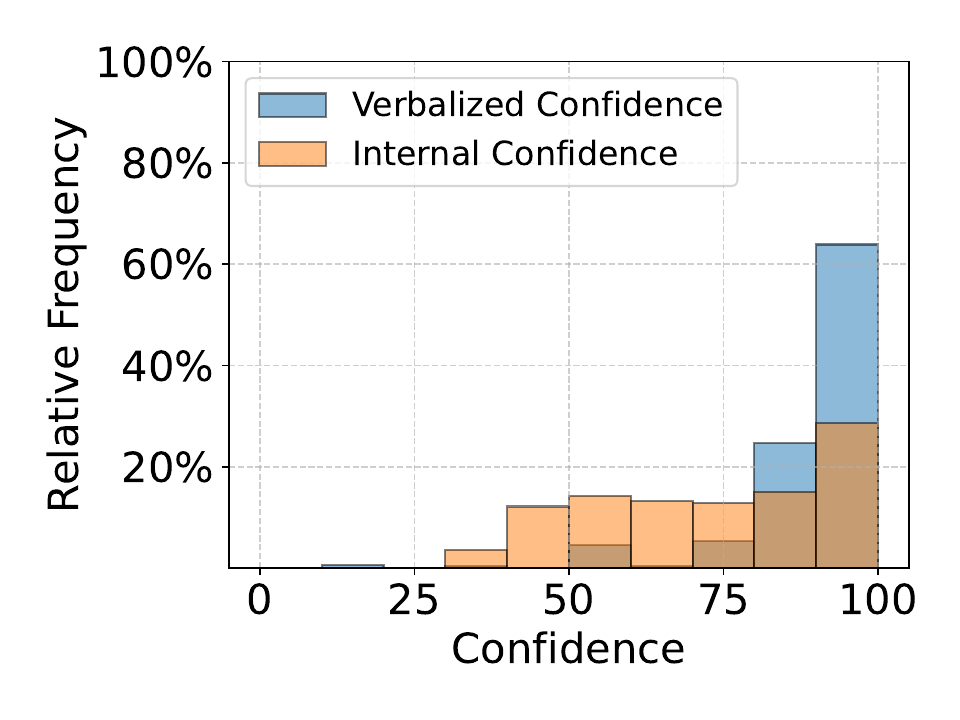}
    \caption{Distributions (Baseline)}
  \end{subfigure}

  \vspace{0.5cm}

  \begin{subfigure}{0.3\textwidth}
    \includegraphics[width=\linewidth]{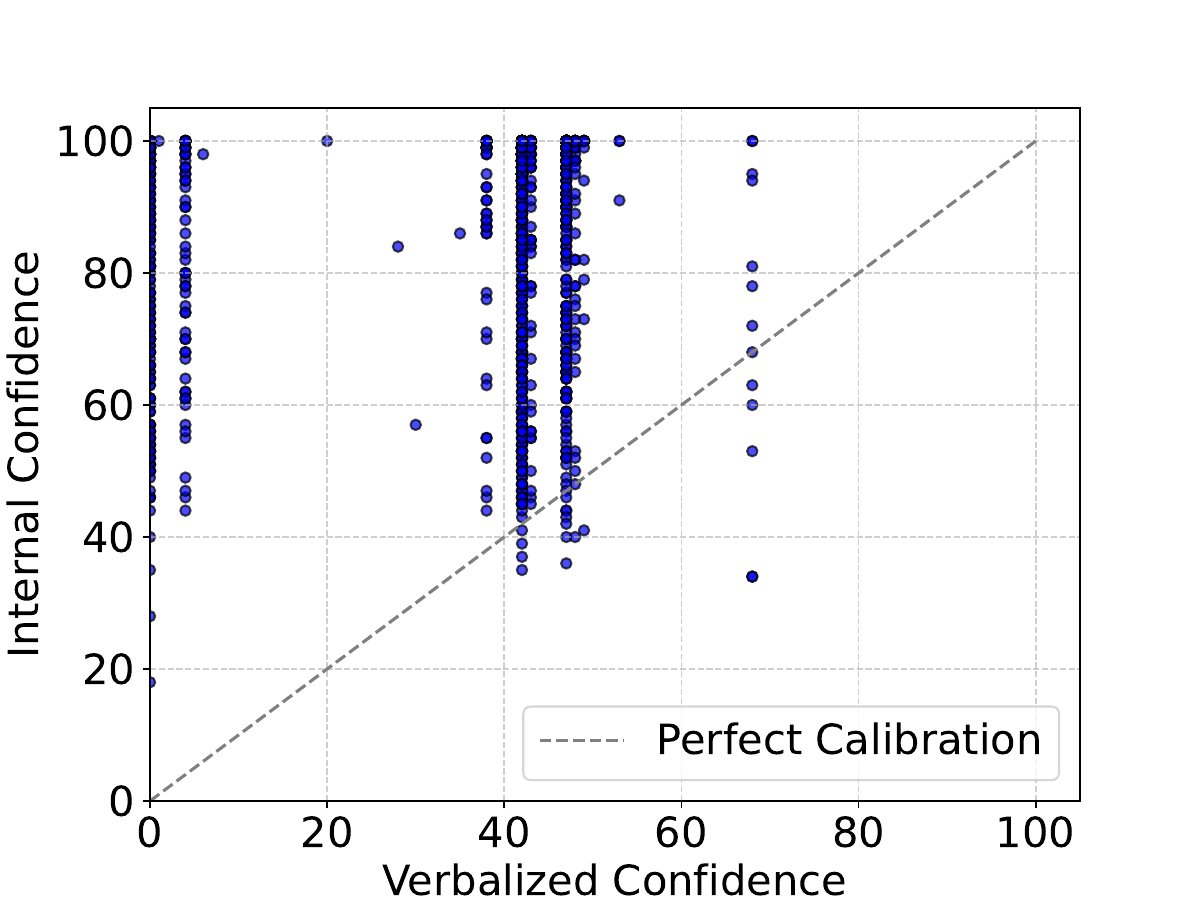}
    \caption{Scatter (DCA)}
  \end{subfigure}
  \hfill
  \begin{subfigure}{0.3\textwidth}
    \includegraphics[width=\linewidth]{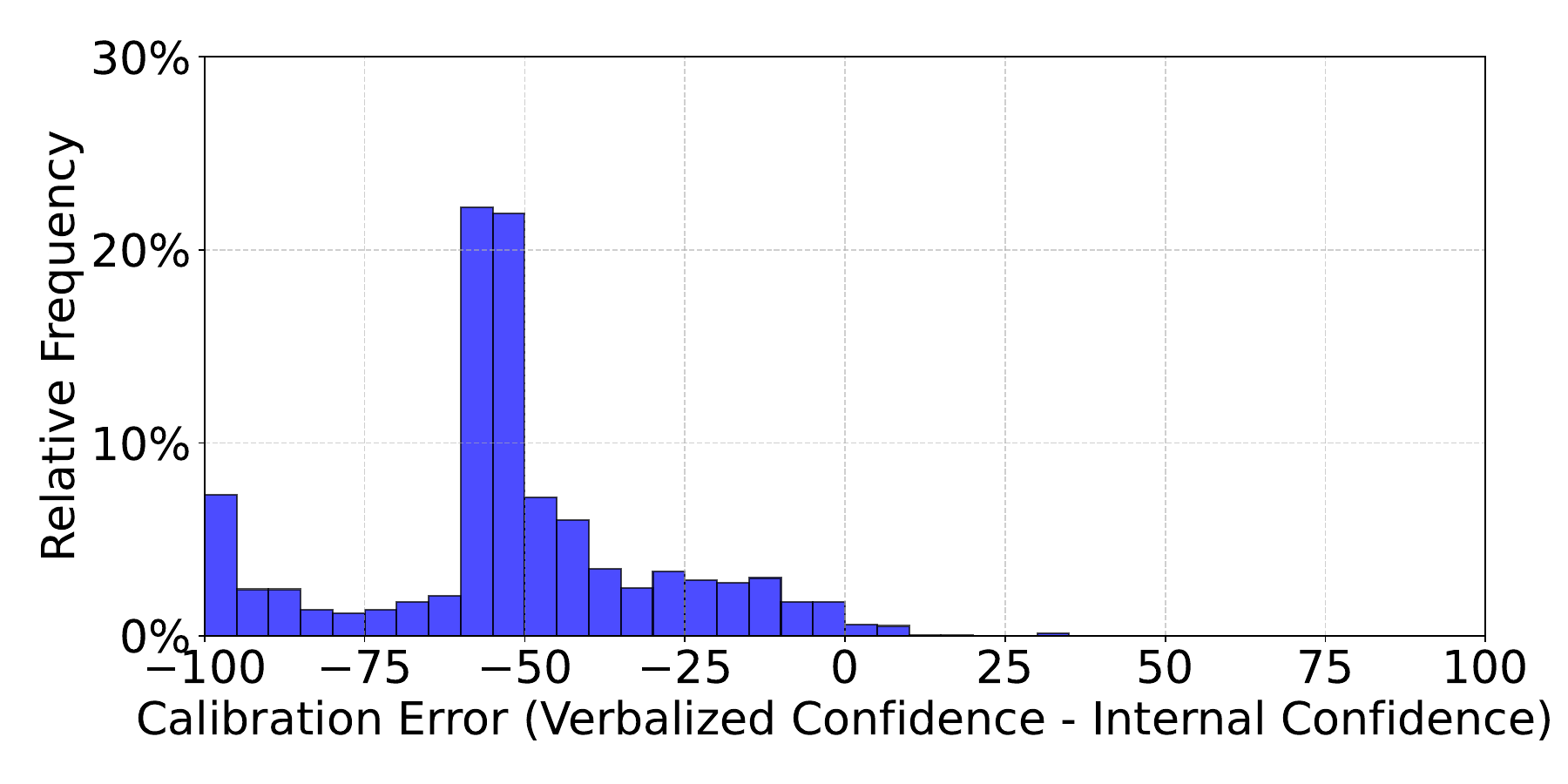}
    \caption{Calibration Error (DCA)}
  \end{subfigure}
  \hfill
  \begin{subfigure}{0.3\textwidth}
    \includegraphics[width=\linewidth]{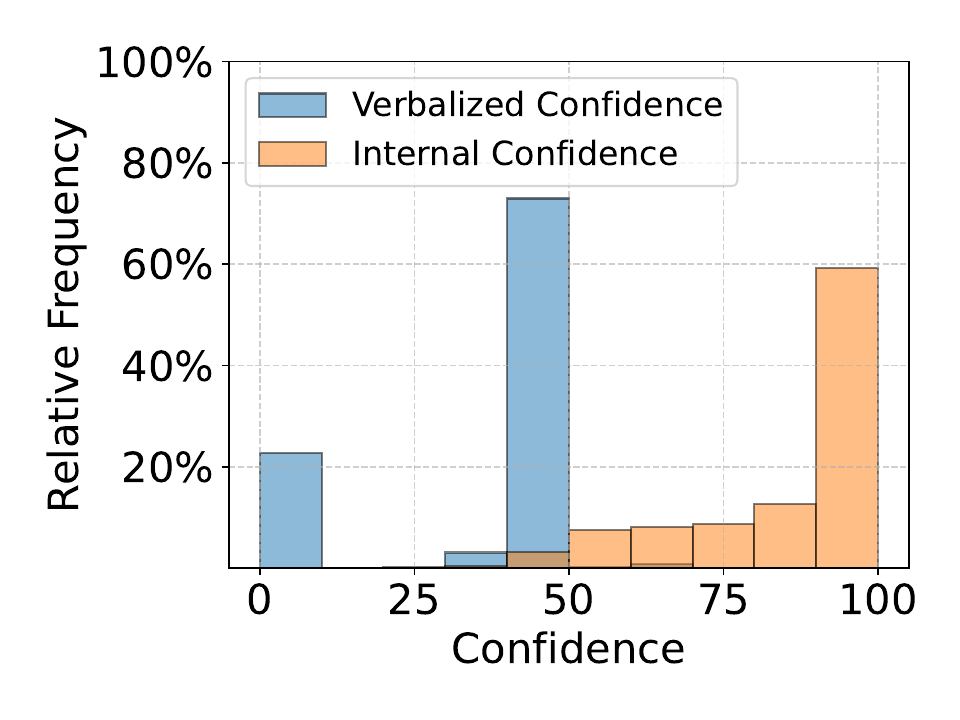}
    \caption{Distributions (DCA)}
  \end{subfigure}

  \caption{Comparison of baseline vs. DCA-trained Mistral-7B-Instruct on CosmosQA. 
  Top row: Verbalized vs. internal confidence scatter plot, calibration error histogram, and confidence score distributions for the baseline model. 
  Bottom row: Same visualizations for the DCA-trained model.}
  \label{mistral_cosmosqa}
\end{figure*}

Figures \ref{mistral_openbookqa}, \ref{gemma_openbookqa}, \ref{llama_openbookqa}, \ref{mistral_truthfulqa}, \ref{gemma_truthfulqa}, \ref{llama_truthfulqa}, \ref{mistral_cosmosqa}, \ref{gemma_cosmosqa}, \ref{llama_cosmosqa}, \ref{mistral_mmlu}, \ref{gemma_mmlu}, and \ref{llama_mmlu} below show supplementary figures for the results of Mistral-7B-Instruct, Gemma-2-9B-Instruct, and  Llama-3.2-3B-Instruct along with their DCA-trained counterparts on OpenBookQA, TruthfulQA, CosmosQA, and MMLU respectively. As seen in the figures, for each model the observed visual trends were broadly consistent across all datasets, also suggesting that the effects of DCA are more model-dependent than task-dependent. 

In particular, Mistral-7B-Instruct and Llama-3.2-3B-Instruct demonstrate consistent verbalized underconfidence after DCA across all datasets, with Mistral-7B-Instruct responding with verbalized confidence values between 40-50\% for the majority of questions during evaluation, and Llama-3.2-3B-Instruct's verbalized confidence distribution shifting towards 0-50\%. Interestingly, the internal confidence distributions of Mistral-7B-Instruct tended to skew more heavily towards higher confidence values after DCA, with an increased number of internal confidence values in the 75-100\% range. In addition, the internal confidence distributions of Llama-3.2-3B-Instruct tended to change from favoring confidence values between 25-50\% to more skewed towards values of 50-100\% after DCA. Unlike the other models, Gemma-2-9B-Instruct's internal and verbalized confidence distributions did not change significantly both before and after DCA.

\begin{figure*}[t]
  \centering

  \begin{subfigure}{0.3\textwidth}
    \includegraphics[width=\linewidth]{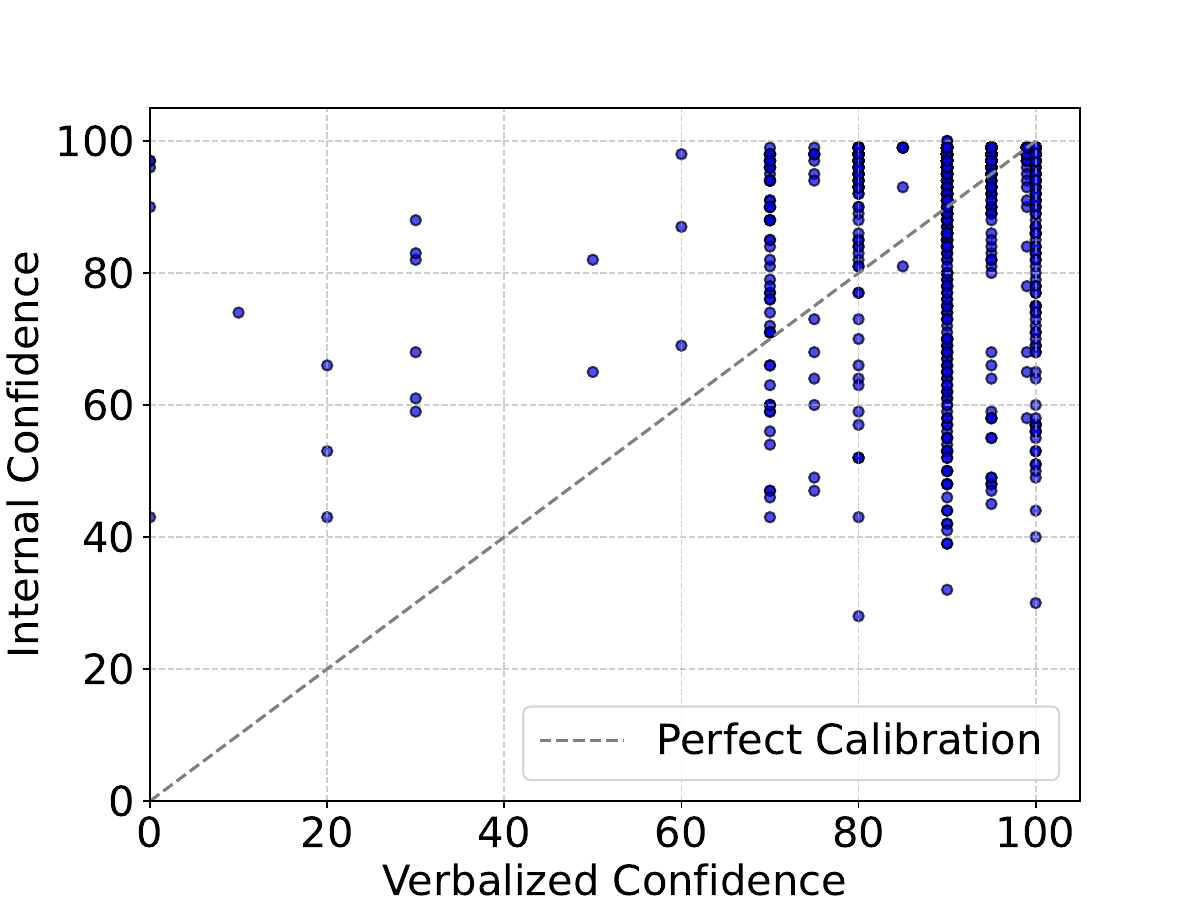}
    \caption{Scatter (Baseline)}
  \end{subfigure}
  \hfill
  \begin{subfigure}{0.3\textwidth}
    \includegraphics[width=\linewidth]{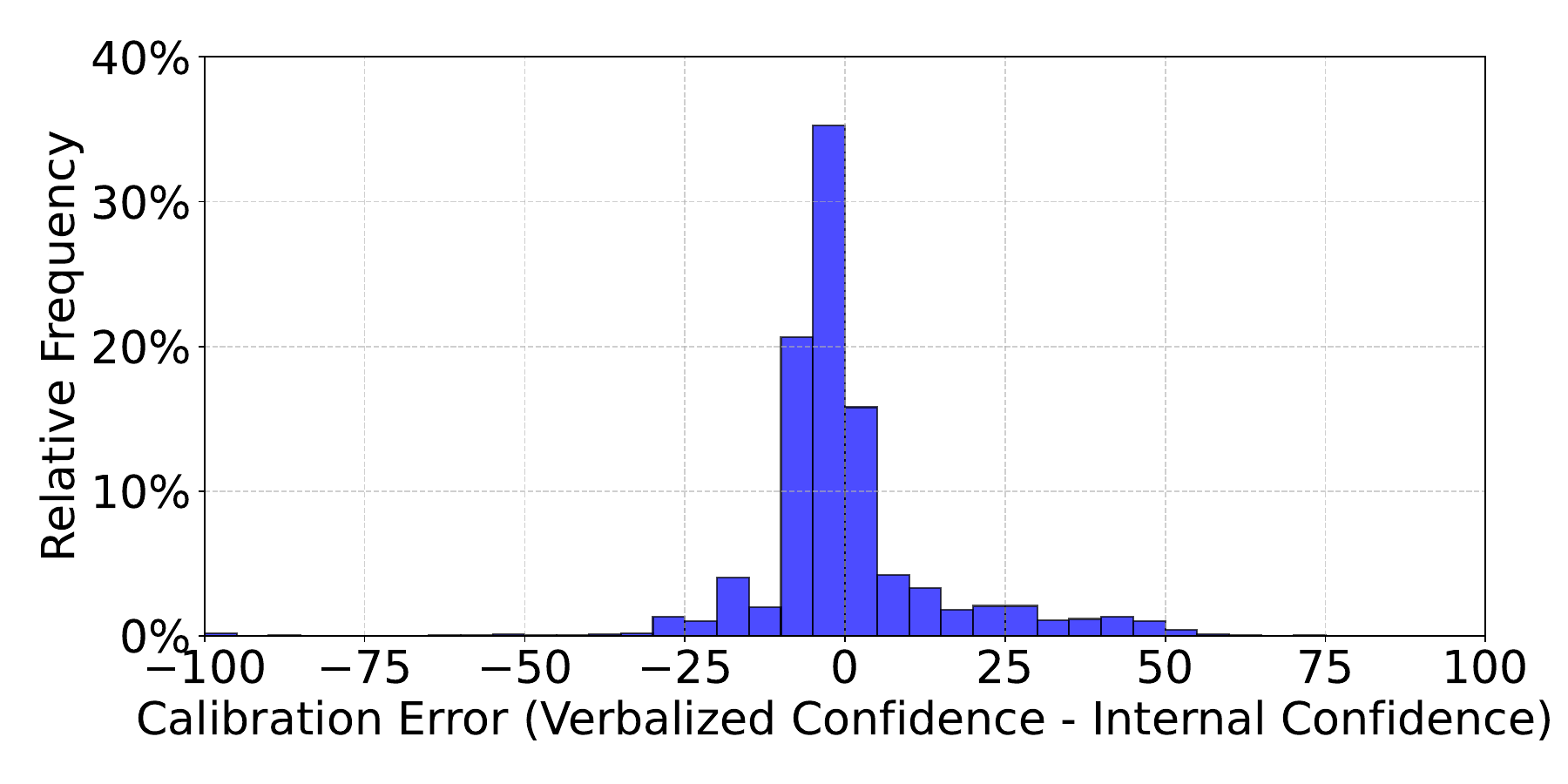}
    \caption{Calibration Error (Baseline)}
  \end{subfigure}
  \hfill
  \begin{subfigure}{0.3\textwidth}
    \includegraphics[width=\linewidth]{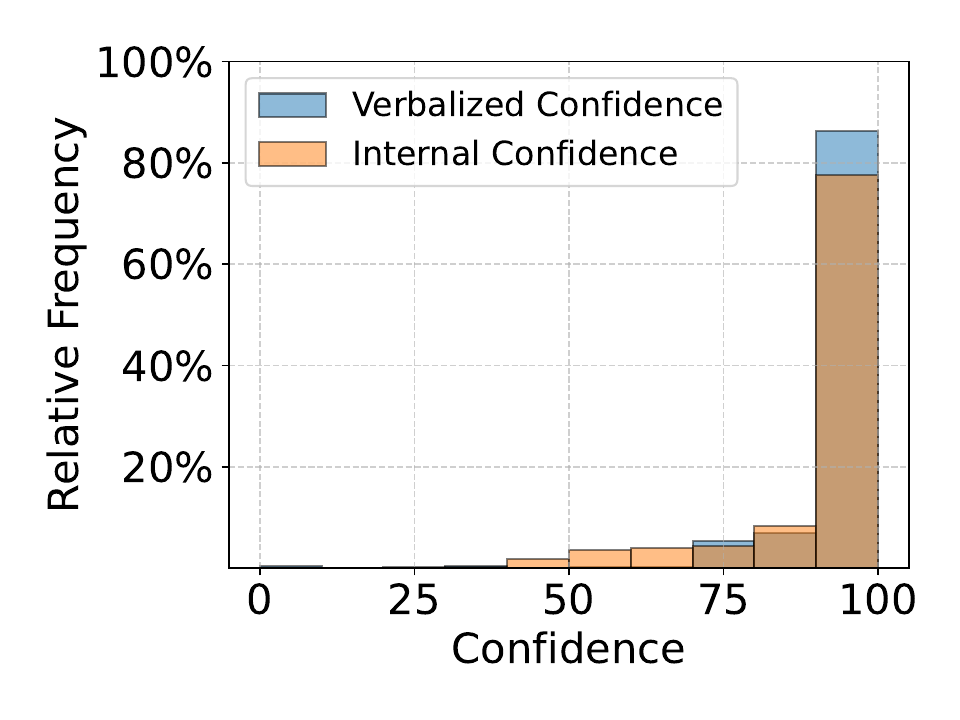}
    \caption{Distributions (Baseline)}
  \end{subfigure}

  \vspace{0.5cm}

  \begin{subfigure}{0.3\textwidth}
    \includegraphics[width=\linewidth]{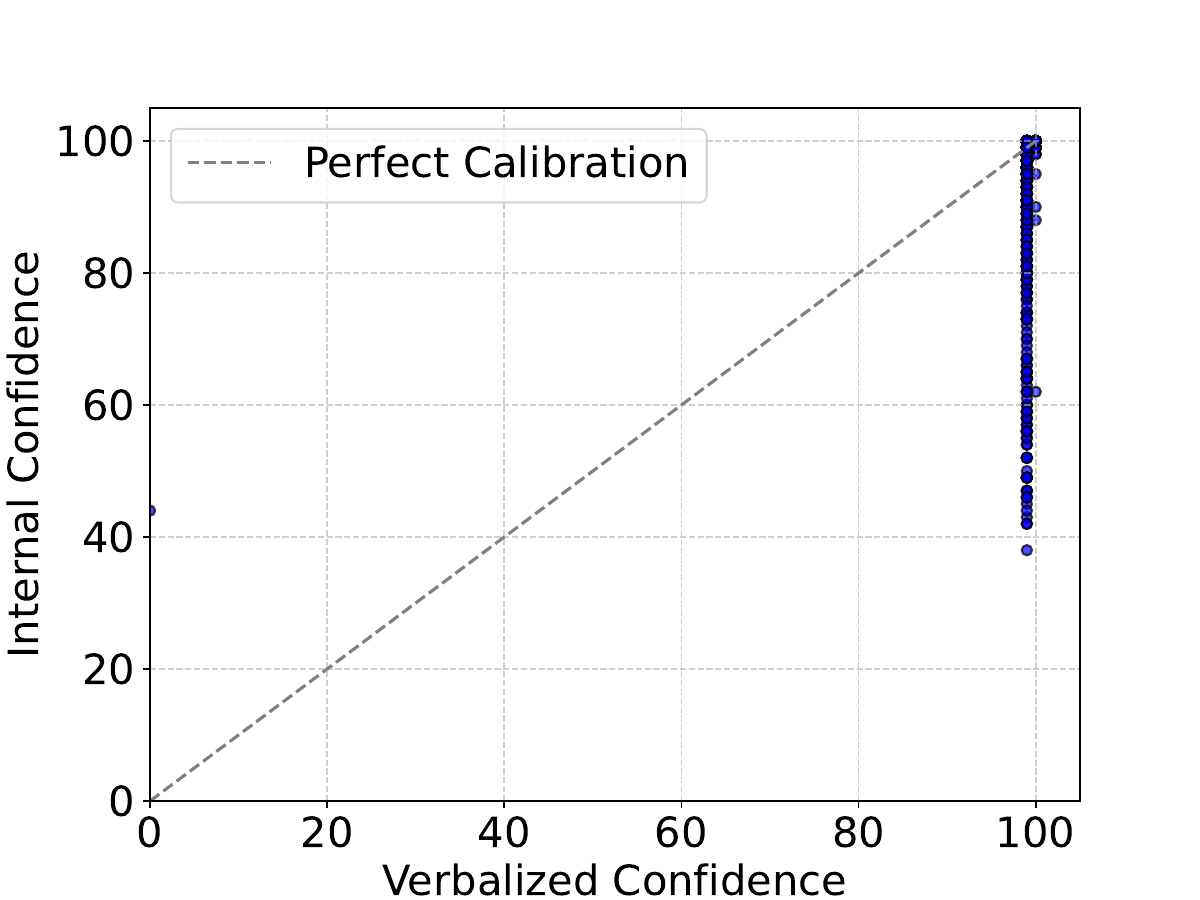}
    \caption{Scatter (DCA)}
  \end{subfigure}
  \hfill
  \begin{subfigure}{0.3\textwidth}
    \includegraphics[width=\linewidth]{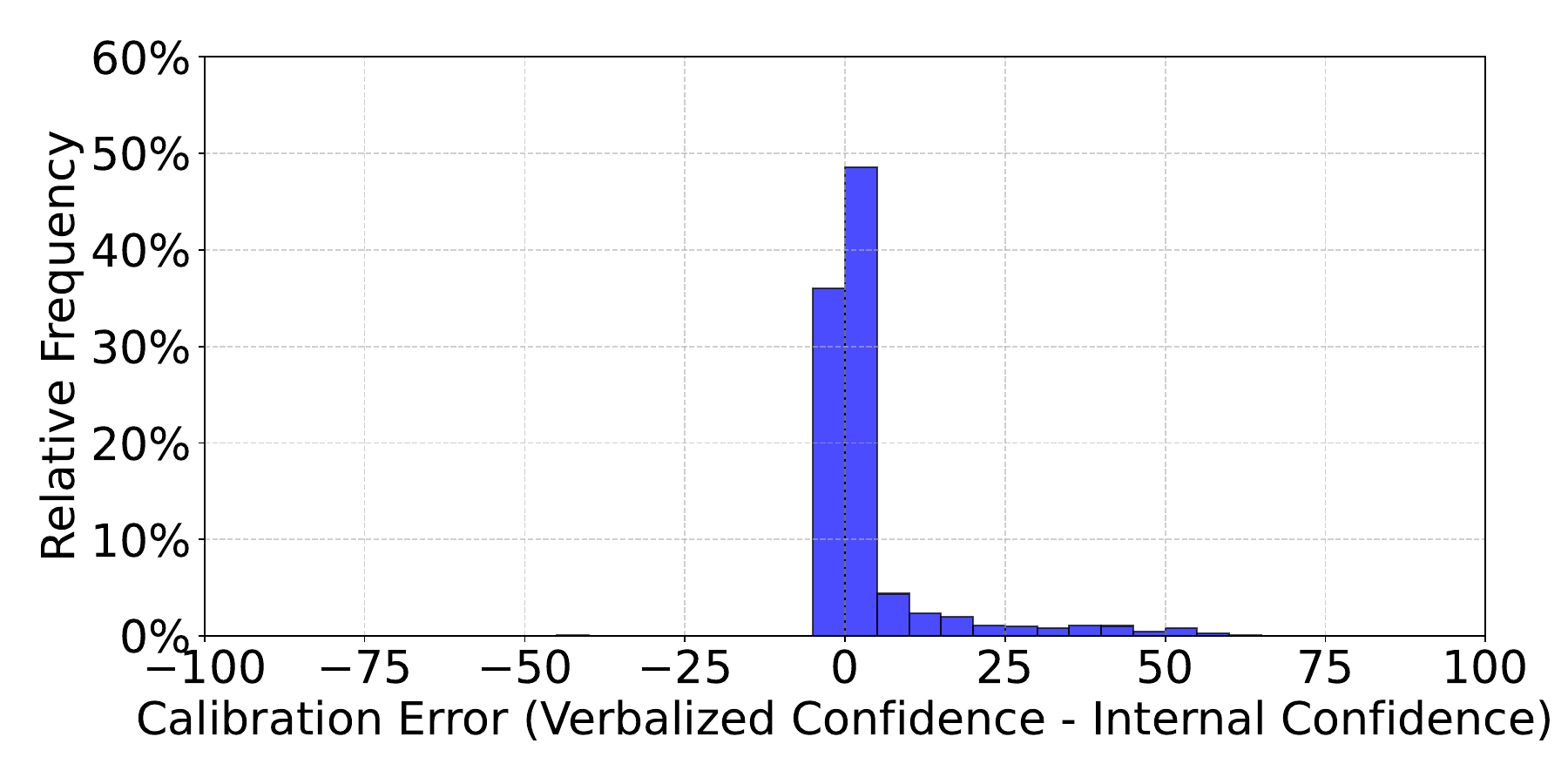}
    \caption{Calibration Error (DCA)}
  \end{subfigure}
  \hfill
  \begin{subfigure}{0.3\textwidth}
    \includegraphics[width=\linewidth]{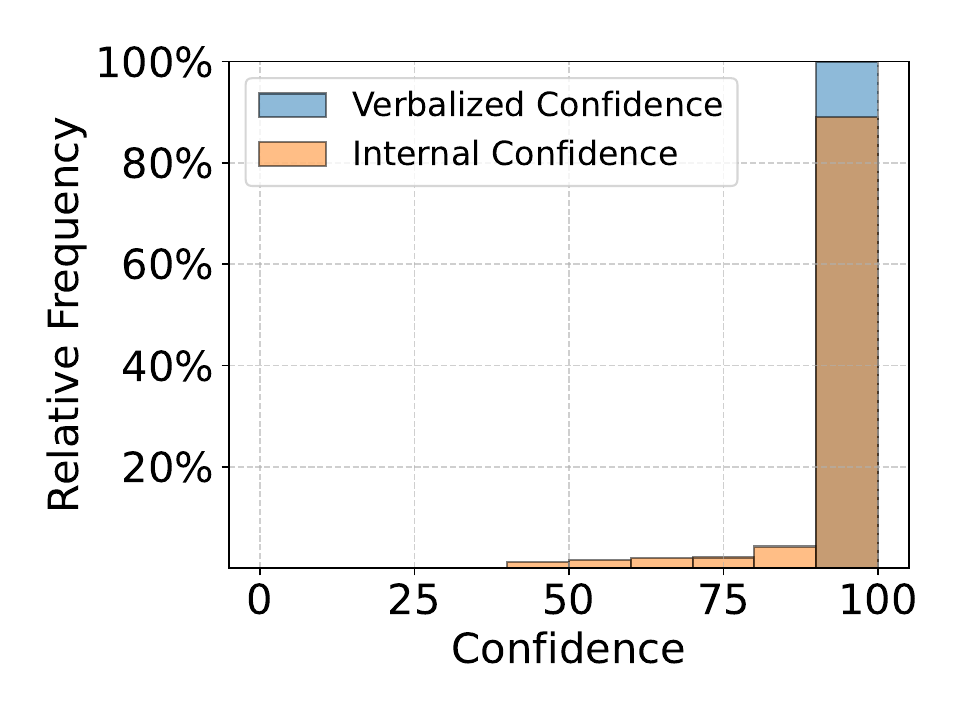}
    \caption{Distributions (DCA)}
  \end{subfigure}

  \caption{Comparison of baseline vs. DCA-trained Gemma-2-9B-Instruct on CosmosQA. 
  Top row: Verbalized vs. internal confidence scatter plot, calibration error histogram, and confidence score distributions for the baseline model. 
  Bottom row: Same visualizations for the DCA-trained model.}
  \label{gemma_cosmosqa}
\end{figure*}

\begin{figure*}[t]
  \centering

  \begin{subfigure}{0.3\textwidth}
    \includegraphics[width=\linewidth]{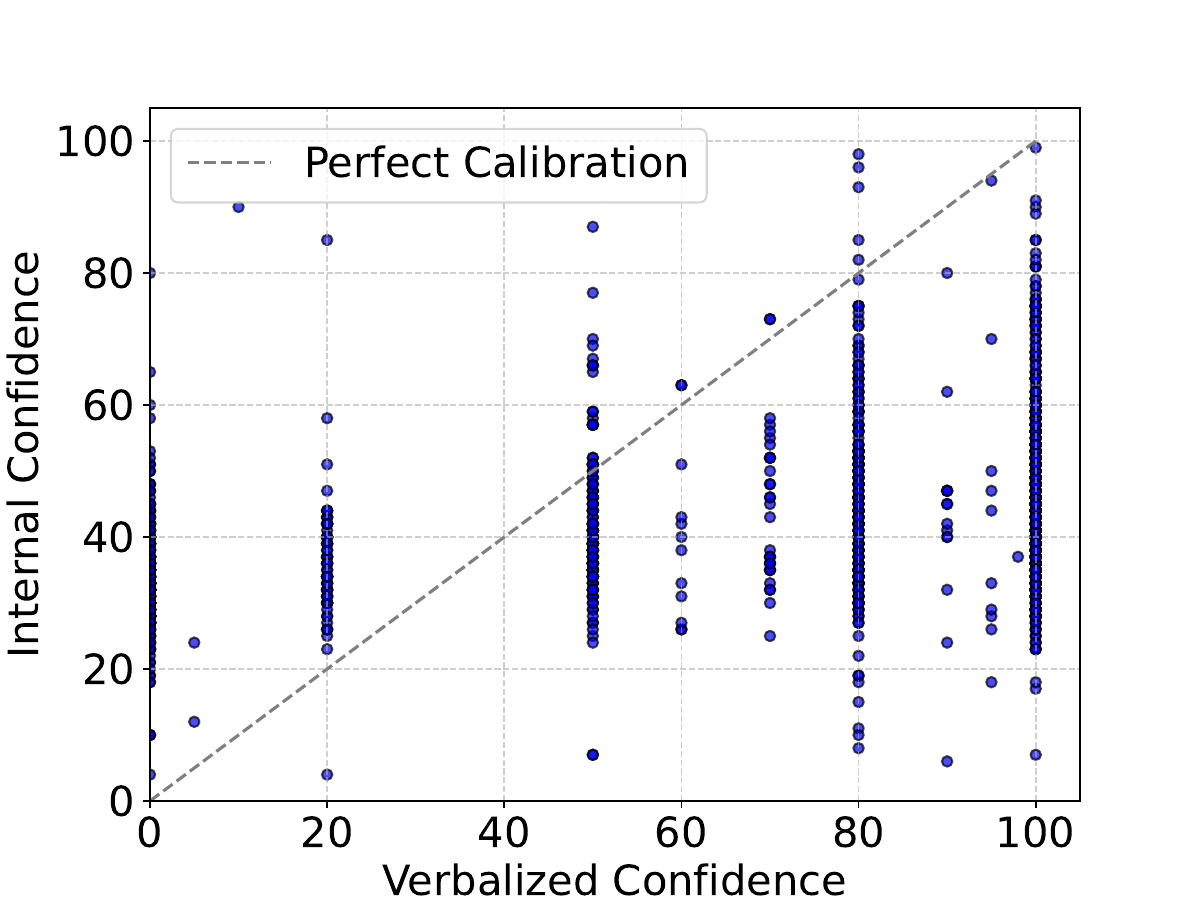}
    \caption{Scatter (Baseline)}
  \end{subfigure}
  \hfill
  \begin{subfigure}{0.3\textwidth}
    \includegraphics[width=\linewidth]{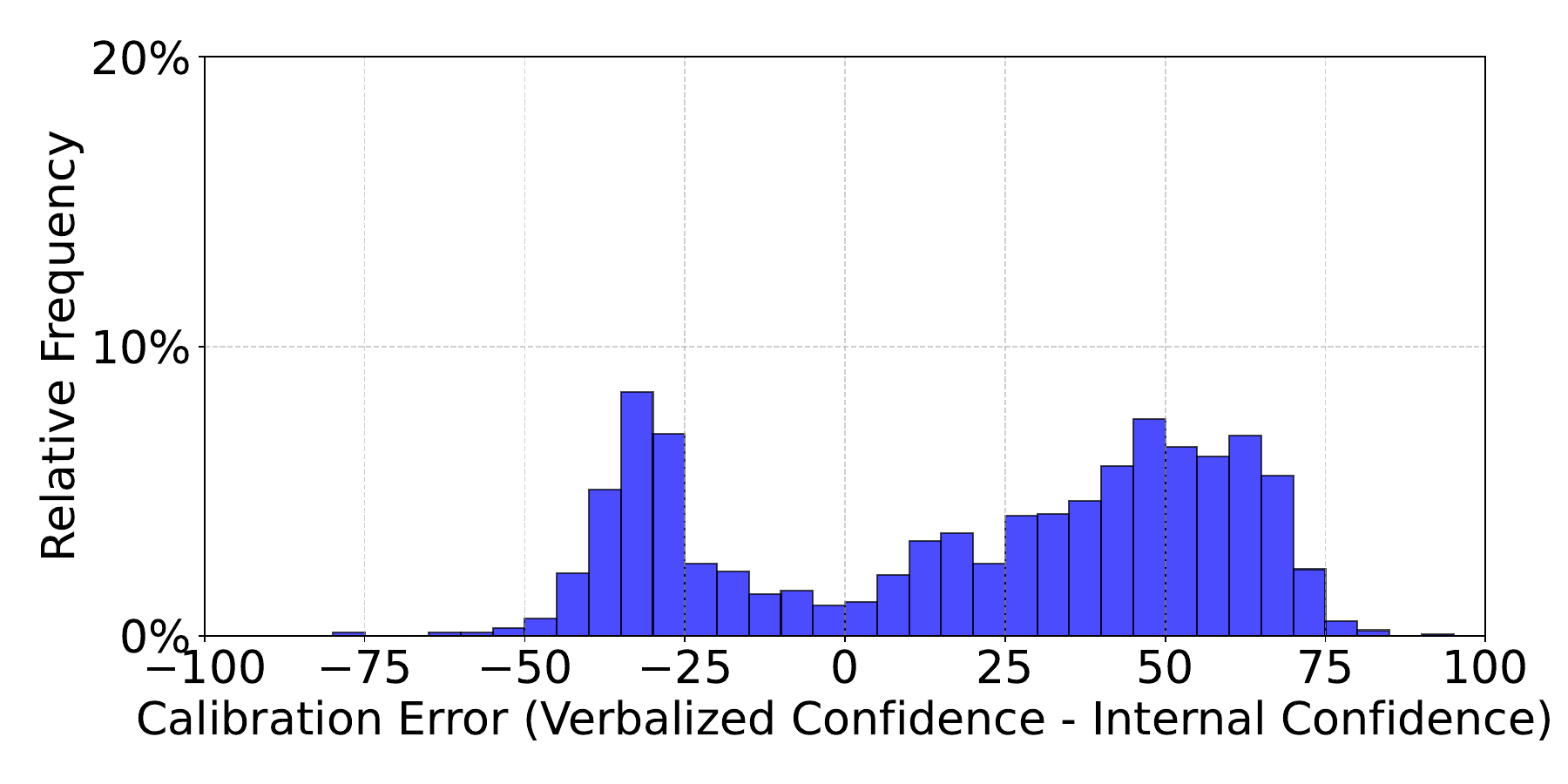}
    \caption{Calibration Error (Baseline)}
  \end{subfigure}
  \hfill
  \begin{subfigure}{0.3\textwidth}
    \includegraphics[width=\linewidth]{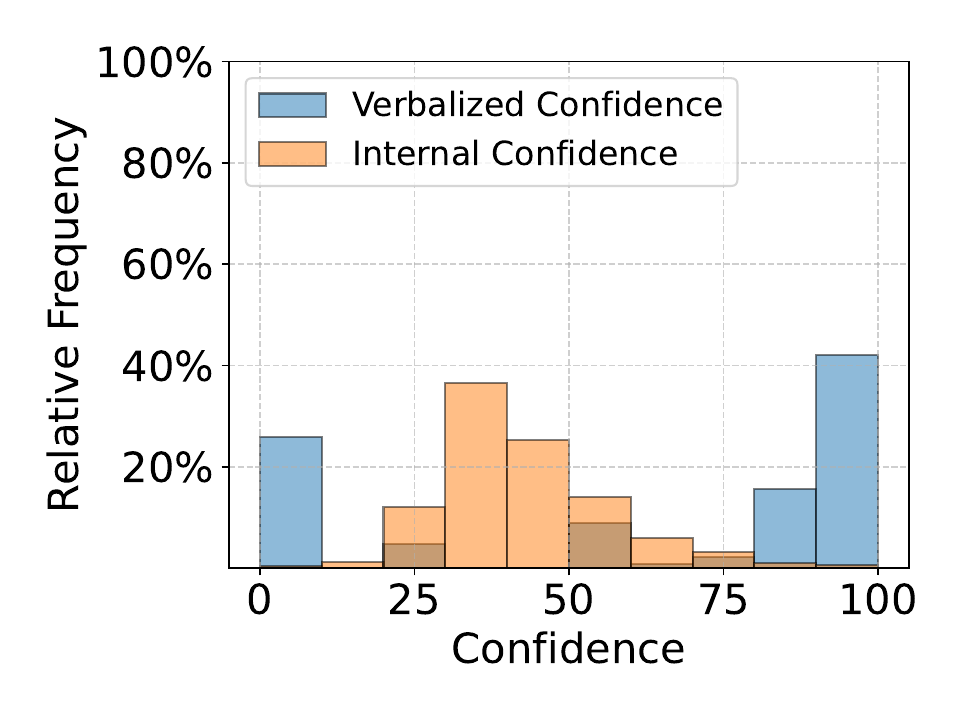}
    \caption{Distributions (Baseline)}
  \end{subfigure}

  \vspace{0.5cm}

  \begin{subfigure}{0.3\textwidth}
    \includegraphics[width=\linewidth]{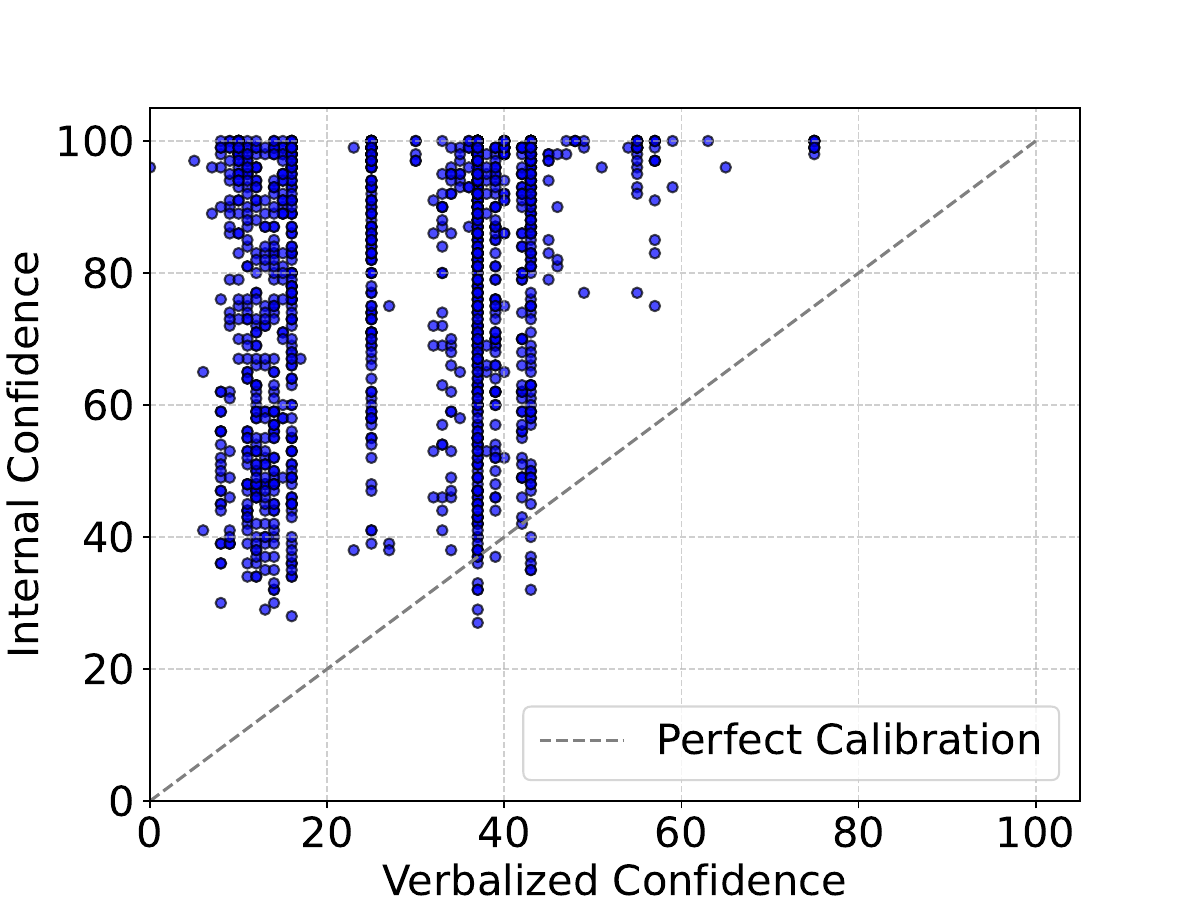}
    \caption{Scatter (DCA)}
  \end{subfigure}
  \hfill
  \begin{subfigure}{0.3\textwidth}
    \includegraphics[width=\linewidth]{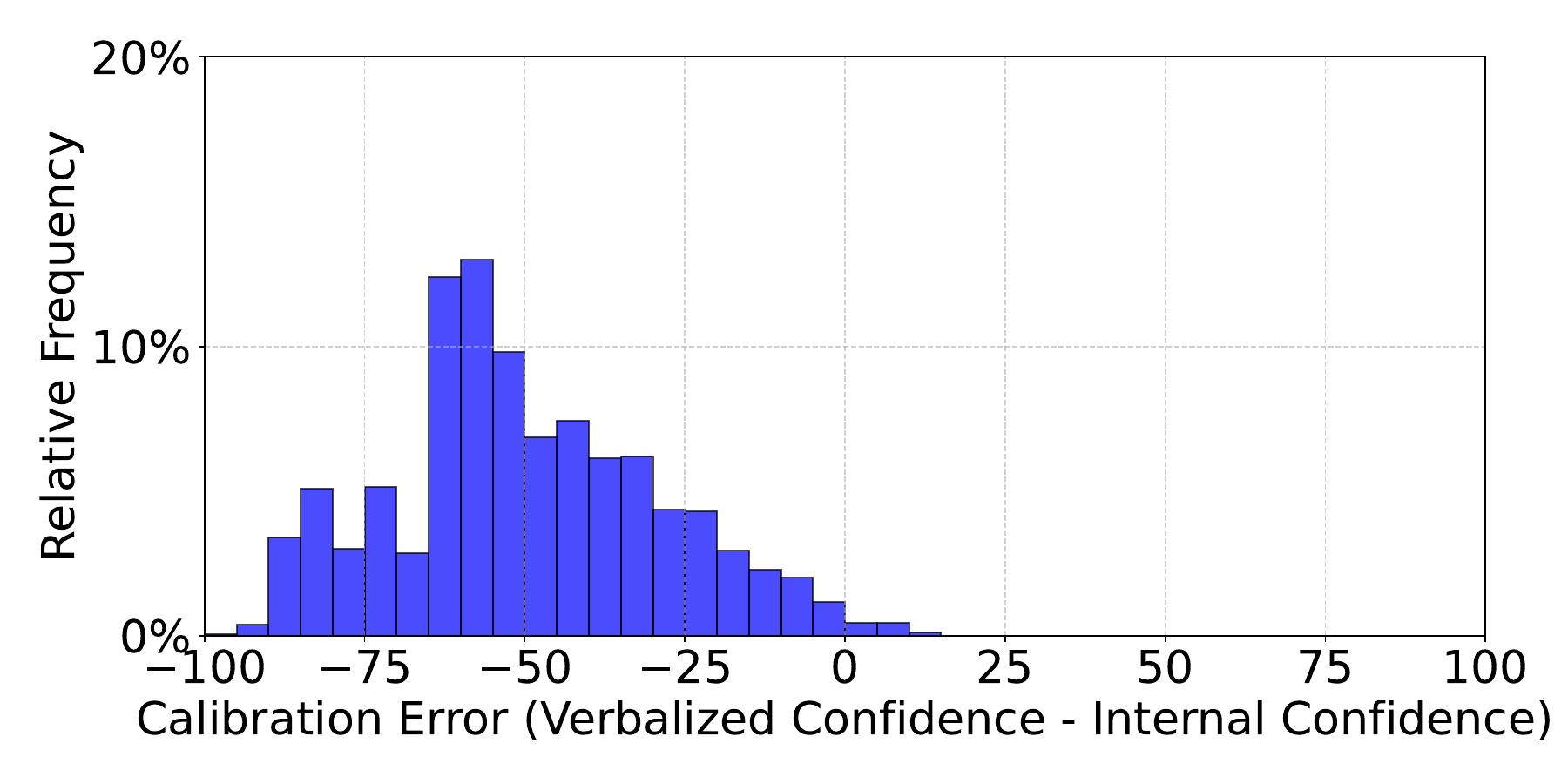}
    \caption{Calibration Error (DCA)}
  \end{subfigure}
  \hfill
  \begin{subfigure}{0.3\textwidth}
    \includegraphics[width=\linewidth]{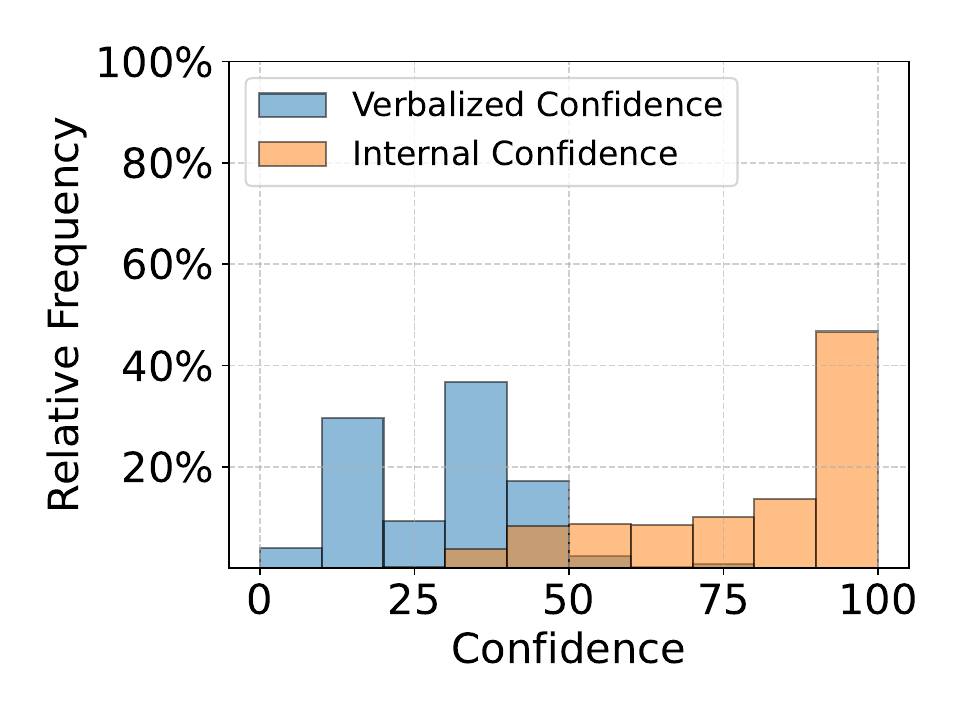}
    \caption{Distributions (DCA)}
  \end{subfigure}

  \caption{Comparison of baseline vs. DCA-trained Llama-3.2-3B-Instruct on CosmosQA. 
  Top row: Verbalized vs. internal confidence scatter plot, calibration error histogram, and confidence score distributions for the baseline model. 
  Bottom row: Same visualizations for the DCA-trained model.}
  \label{llama_cosmosqa}
\end{figure*}

\begin{figure*}[t]
  \centering

  \begin{subfigure}{0.3\textwidth}
    \includegraphics[width=\linewidth]{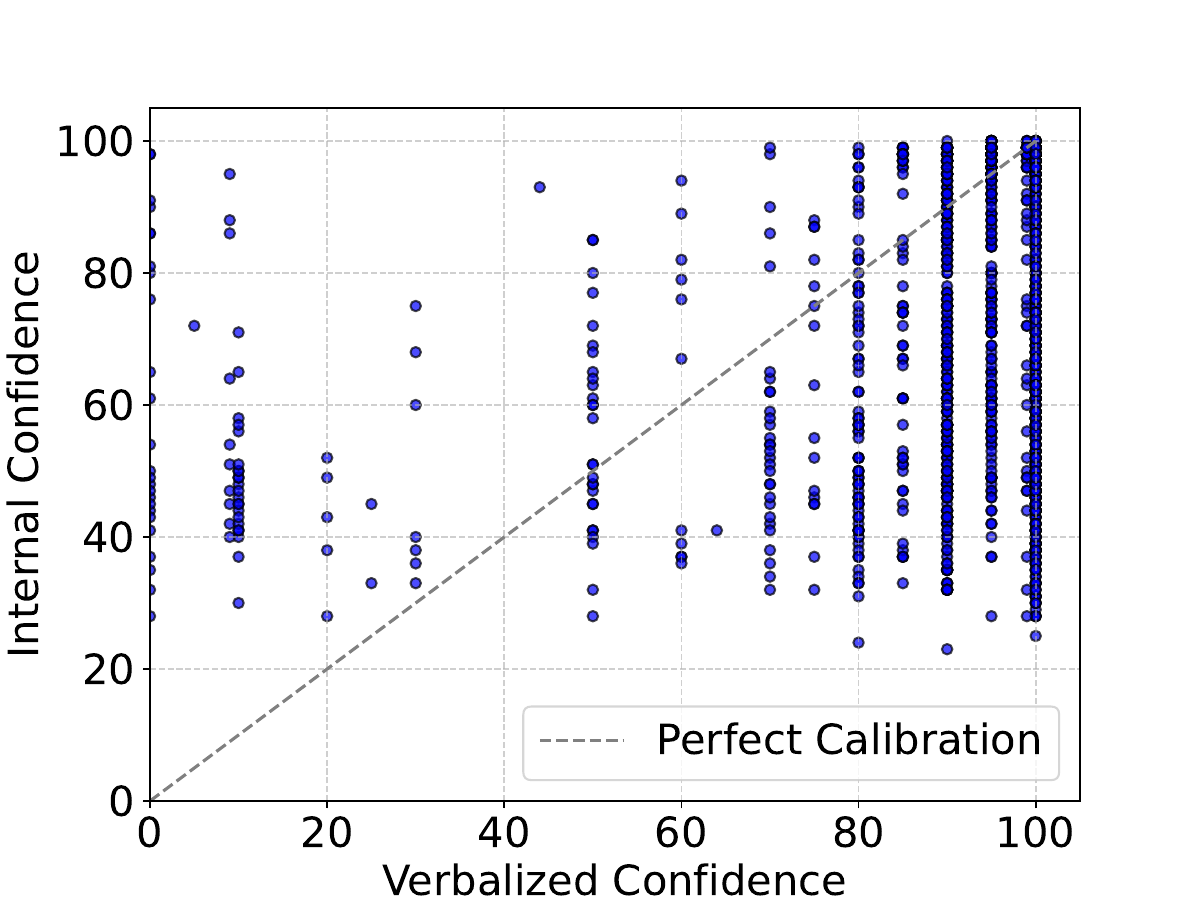}
    \caption{Scatter (Baseline)}
  \end{subfigure}
  \hfill
  \begin{subfigure}{0.3\textwidth}
    \includegraphics[width=\linewidth]{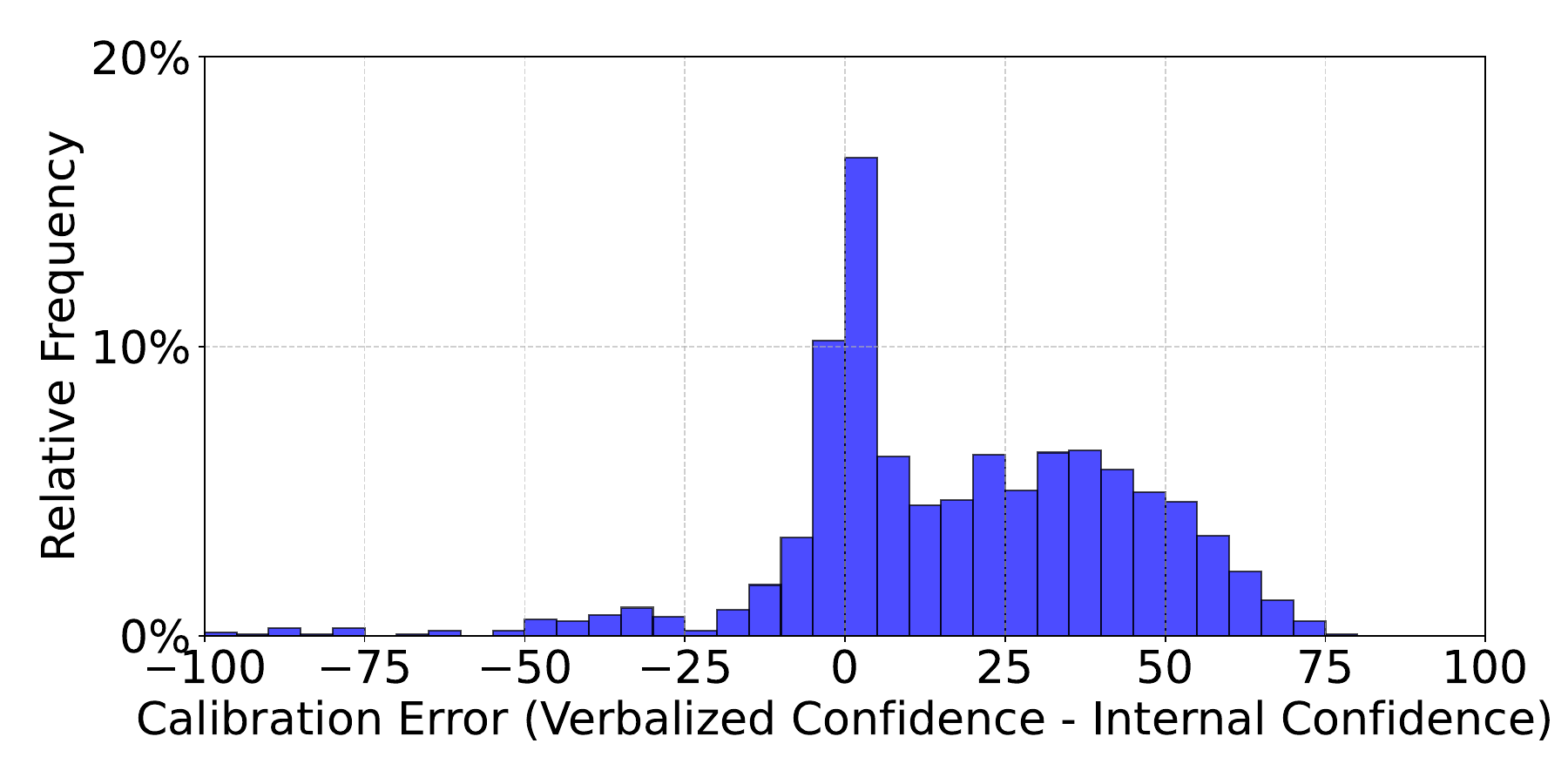}
    \caption{Calibration Error (Baseline)}
  \end{subfigure}
  \hfill
  \begin{subfigure}{0.3\textwidth}
    \includegraphics[width=\linewidth]{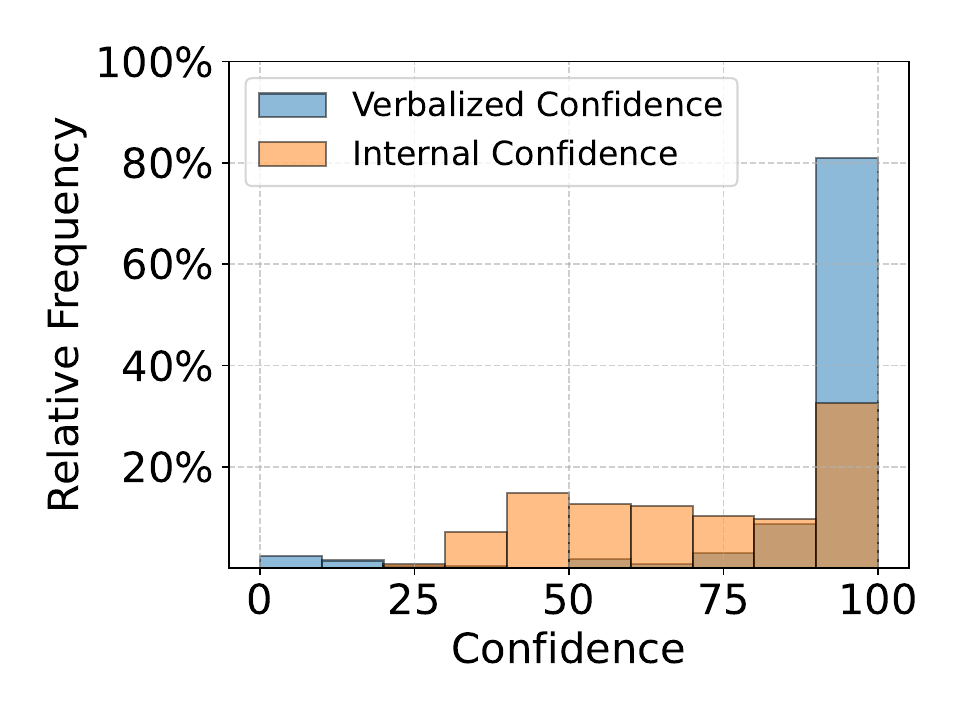}
    \caption{Distributions (Baseline)}
  \end{subfigure}

  \vspace{0.5cm}

  \begin{subfigure}{0.3\textwidth}
    \includegraphics[width=\linewidth]{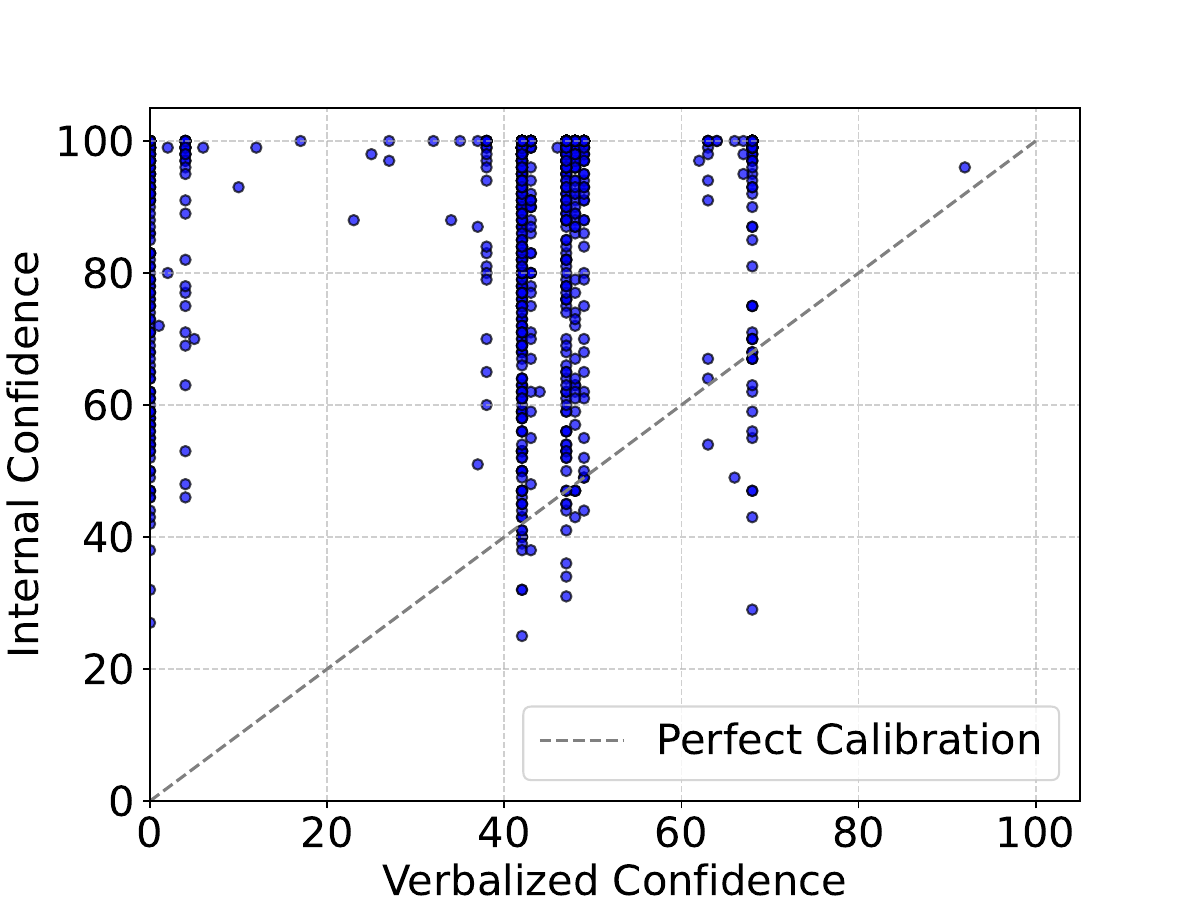}
    \caption{Scatter (DCA)}
  \end{subfigure}
  \hfill
  \begin{subfigure}{0.3\textwidth}
    \includegraphics[width=\linewidth]{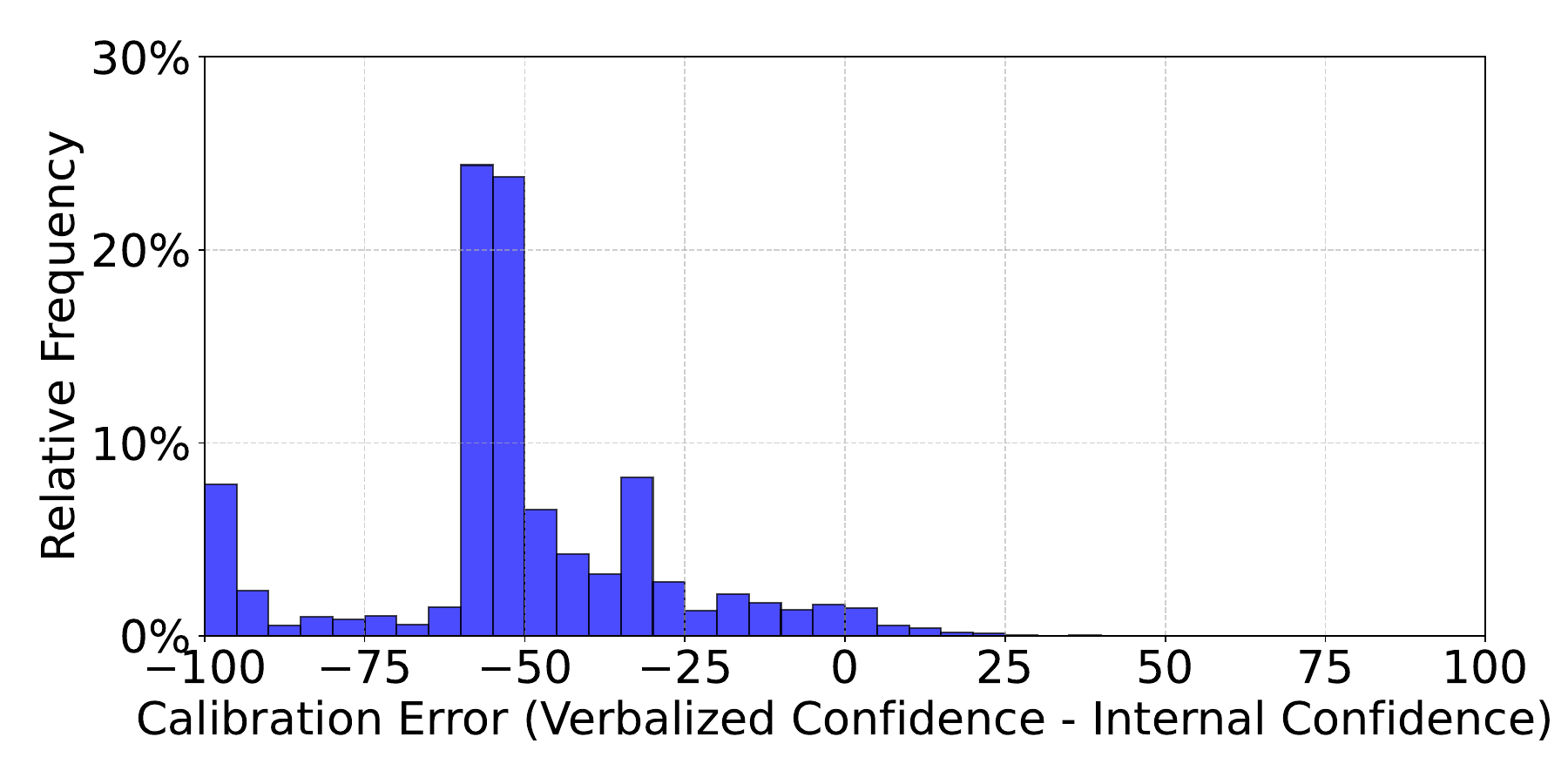}
    \caption{Calibration Error (DCA)}
  \end{subfigure}
  \hfill
  \begin{subfigure}{0.3\textwidth}
    \includegraphics[width=\linewidth]{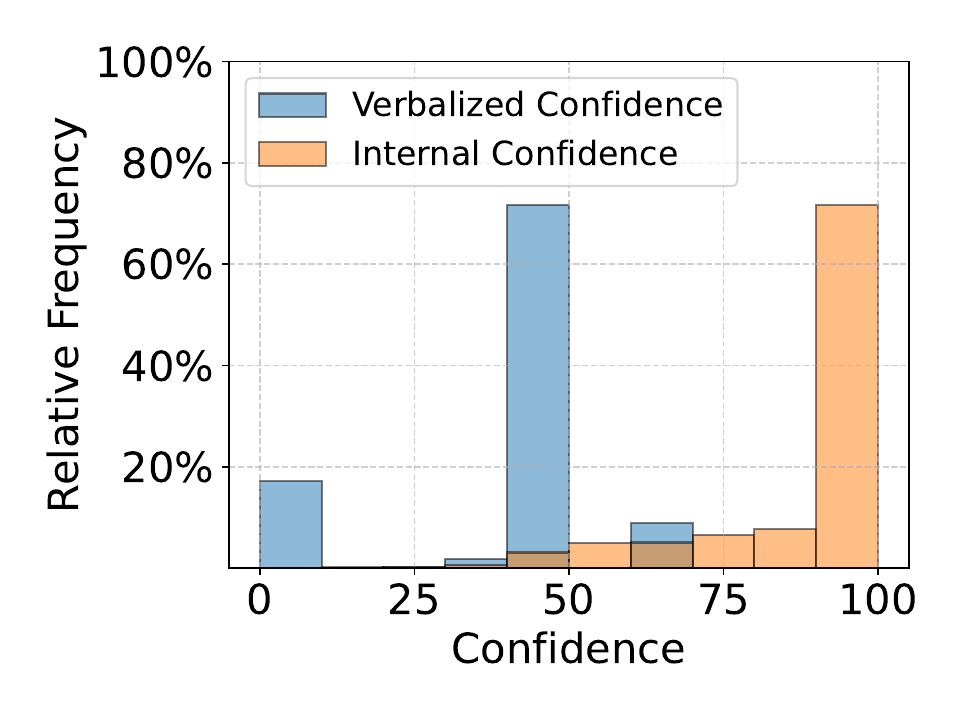}
    \caption{Distributions (DCA)}
  \end{subfigure}

  \caption{Comparison of baseline vs. DCA-trained Mistral-7B-Instruct on MMLU. 
  Top row: Verbalized vs. internal confidence scatter plot, calibration error histogram, and confidence score distributions for the baseline model. 
  Bottom row: Same visualizations for the DCA-trained model.}
  \label{mistral_mmlu}
\end{figure*}

\begin{figure*}[t]
  \centering

  \begin{subfigure}{0.3\textwidth}
    \includegraphics[width=\linewidth]{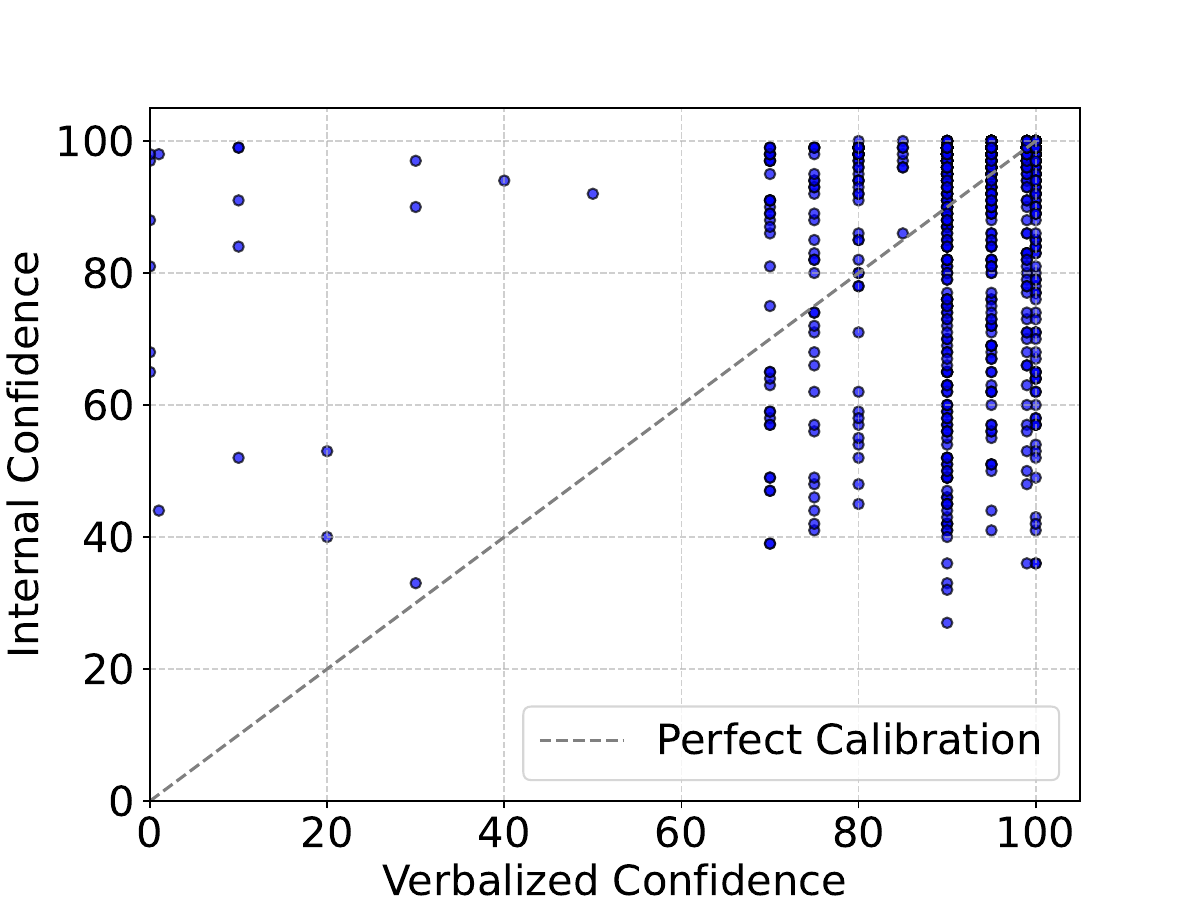}
    \caption{Scatter (Baseline)}
  \end{subfigure}
  \hfill
  \begin{subfigure}{0.3\textwidth}
    \includegraphics[width=\linewidth]{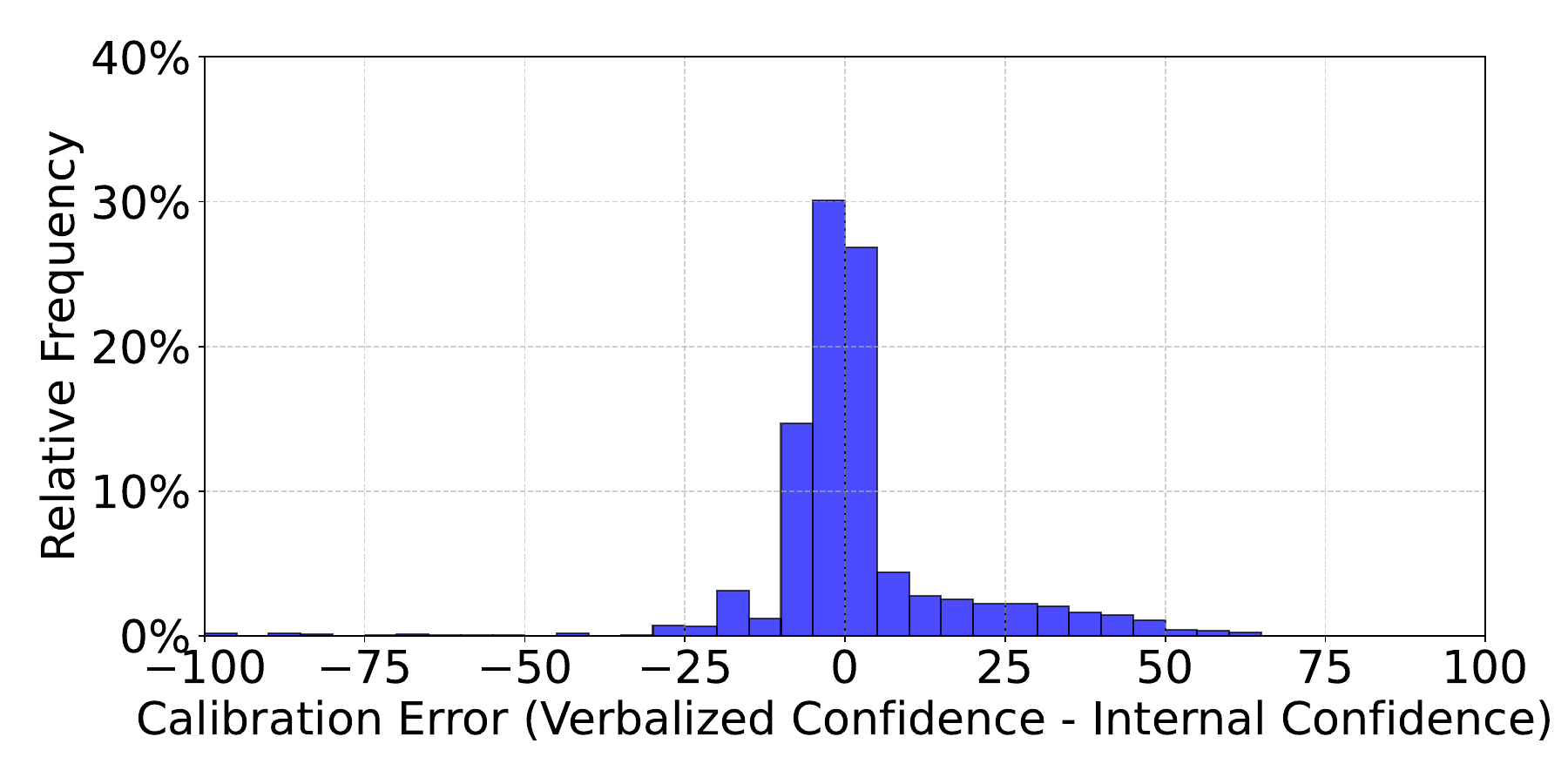}
    \caption{Calibration Error (Baseline)}
  \end{subfigure}
  \hfill
  \begin{subfigure}{0.3\textwidth}
    \includegraphics[width=\linewidth]{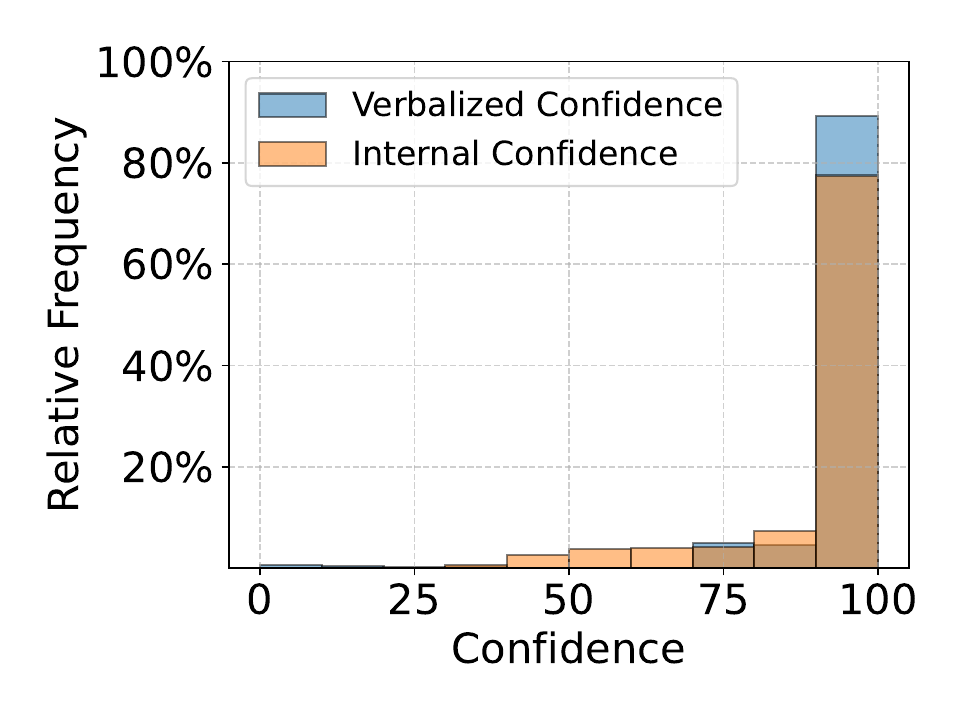}
    \caption{Distributions (Baseline)}
  \end{subfigure}

  \vspace{0.5cm}

  \begin{subfigure}{0.3\textwidth}
    \includegraphics[width=\linewidth]{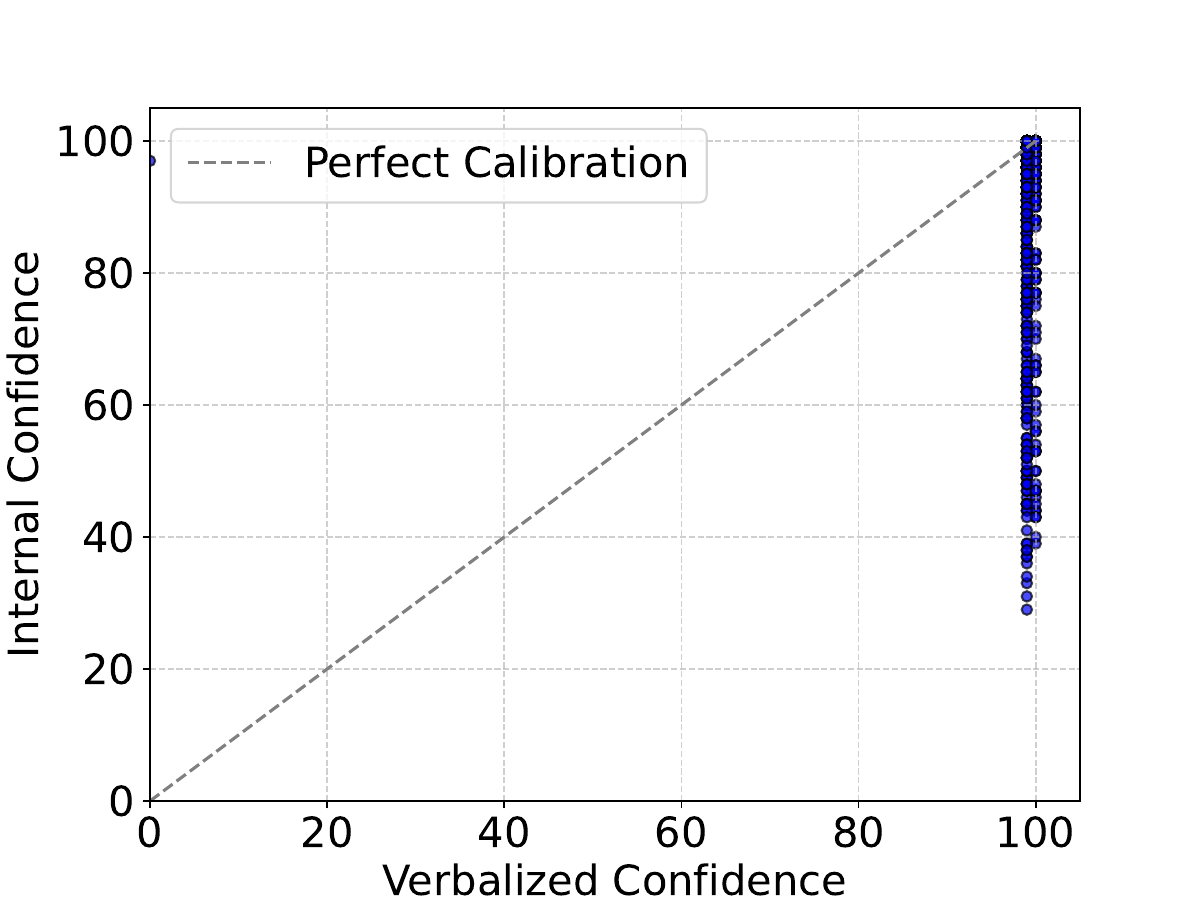}
    \caption{Scatter (DCA)}
  \end{subfigure}
  \hfill
  \begin{subfigure}{0.3\textwidth}
    \includegraphics[width=\linewidth]{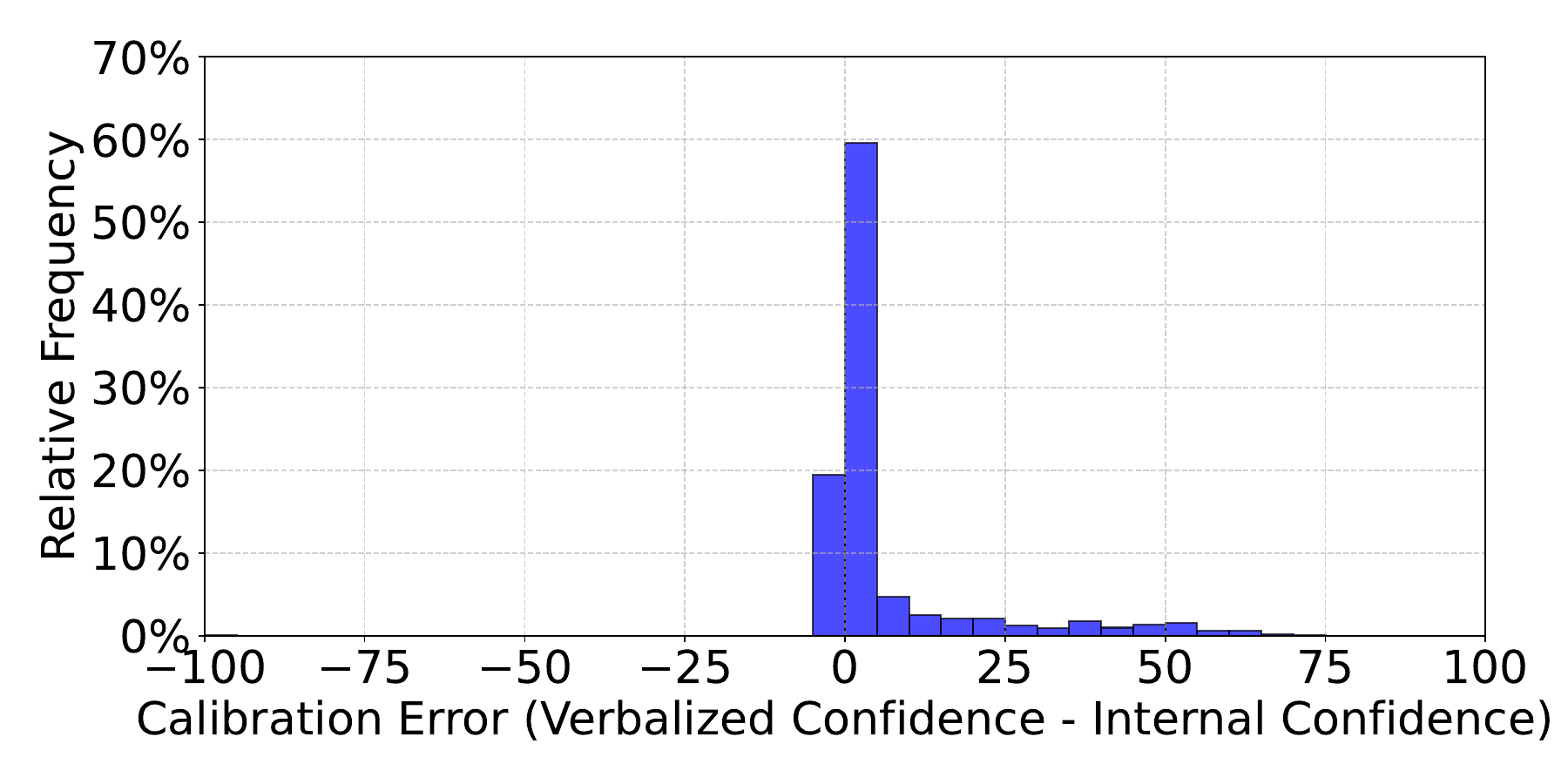}
    \caption{Calibration Error (DCA)}
  \end{subfigure}
  \hfill
  \begin{subfigure}{0.3\textwidth}
    \includegraphics[width=\linewidth]{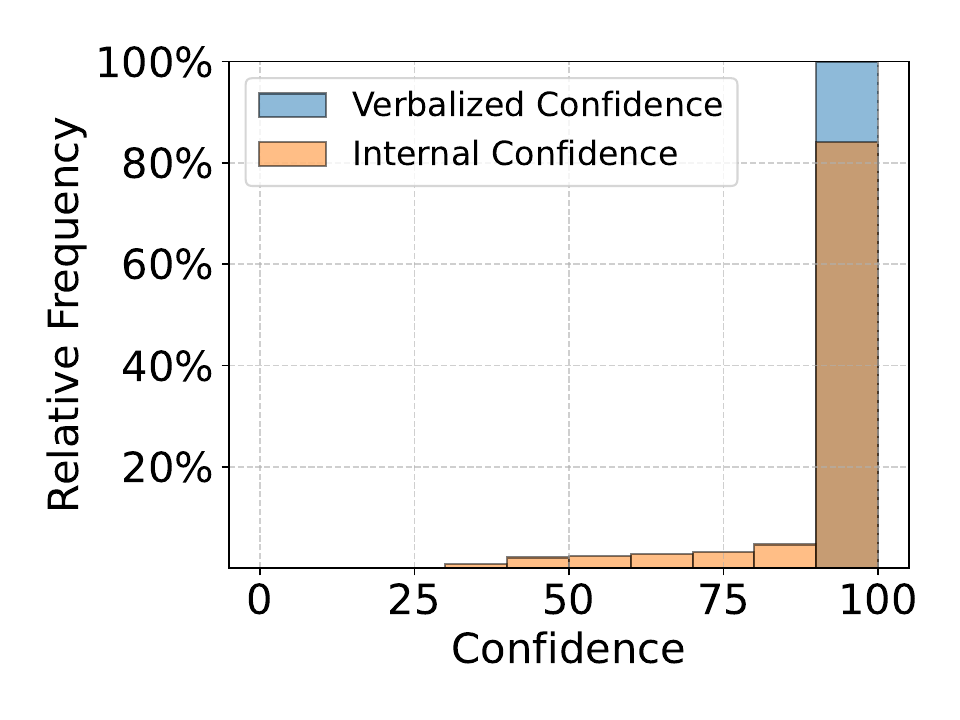}
    \caption{Distributions (DCA)}
  \end{subfigure}

  \caption{Comparison of baseline vs. DCA-trained Gemma-2-9B-Instruct on MMLU. 
  Top row: Verbalized vs. internal confidence scatter plot, calibration error histogram, and confidence score distributions for the baseline model. 
  Bottom row: Same visualizations for the DCA-trained model.}
  \label{gemma_mmlu}
\end{figure*}

\begin{figure*}[t]
  \centering

  \begin{subfigure}{0.3\textwidth}
    \includegraphics[width=\linewidth]{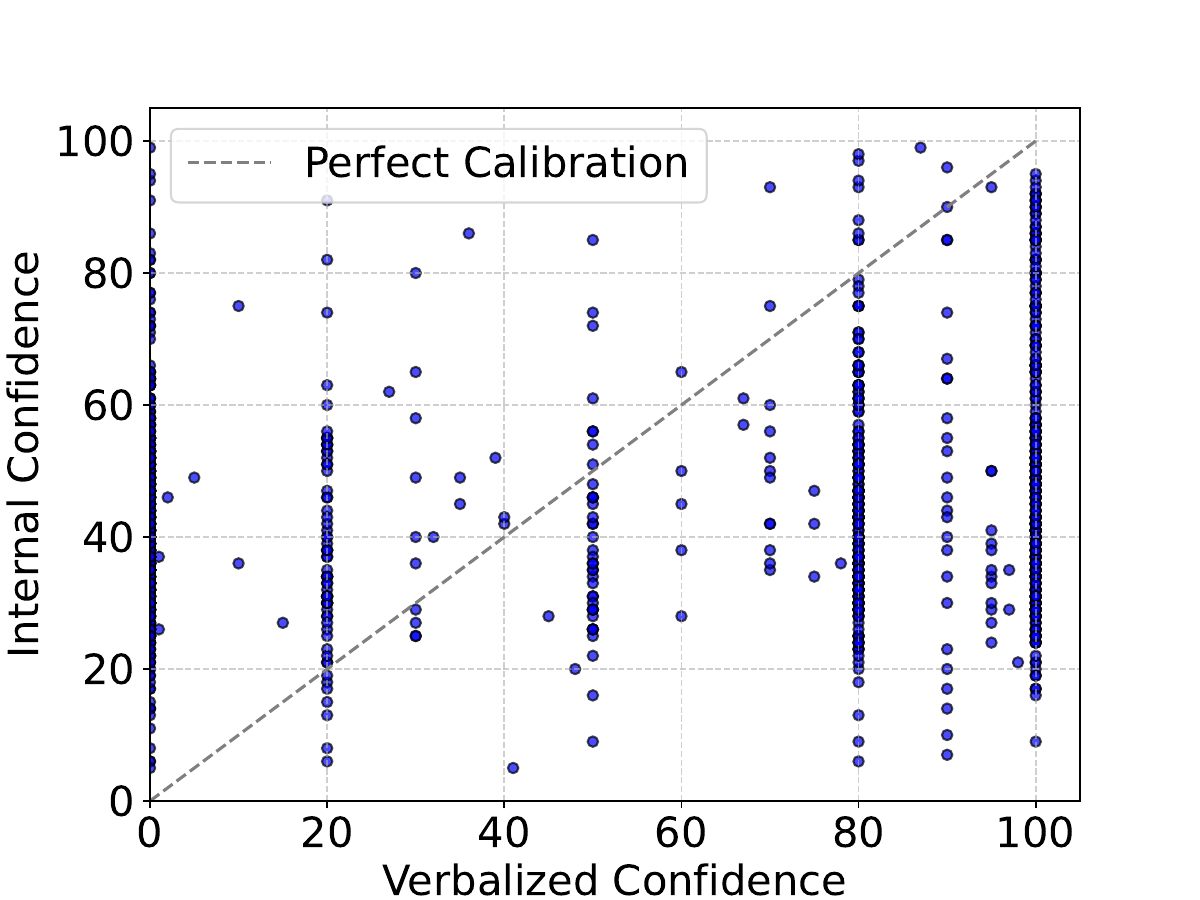}
    \caption{Scatter (Baseline)}
  \end{subfigure}
  \hfill
  \begin{subfigure}{0.3\textwidth}
    \includegraphics[width=\linewidth]{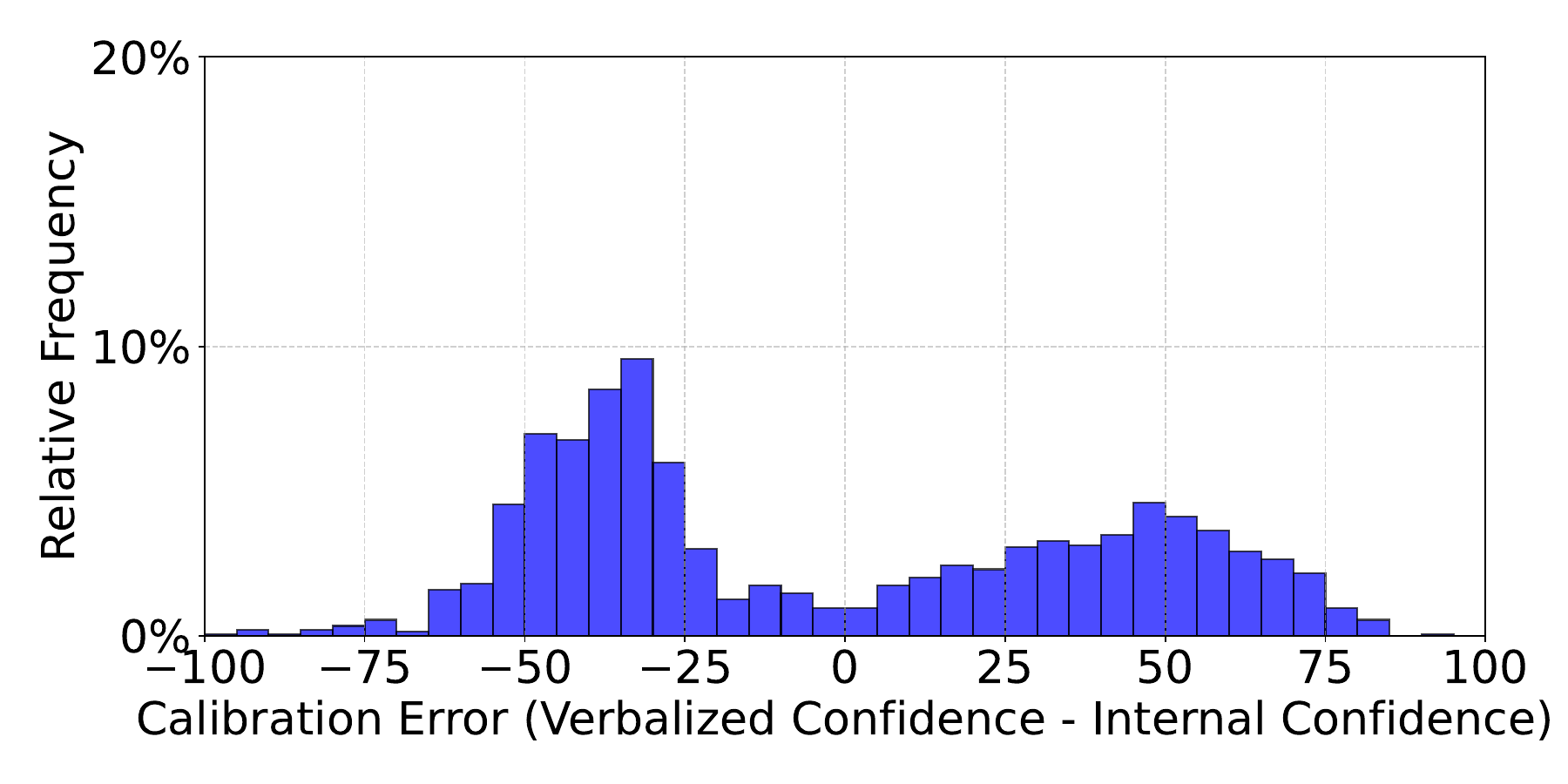}
    \caption{Calibration Error (Baseline)}
  \end{subfigure}
  \hfill
  \begin{subfigure}{0.3\textwidth}
    \includegraphics[width=\linewidth]{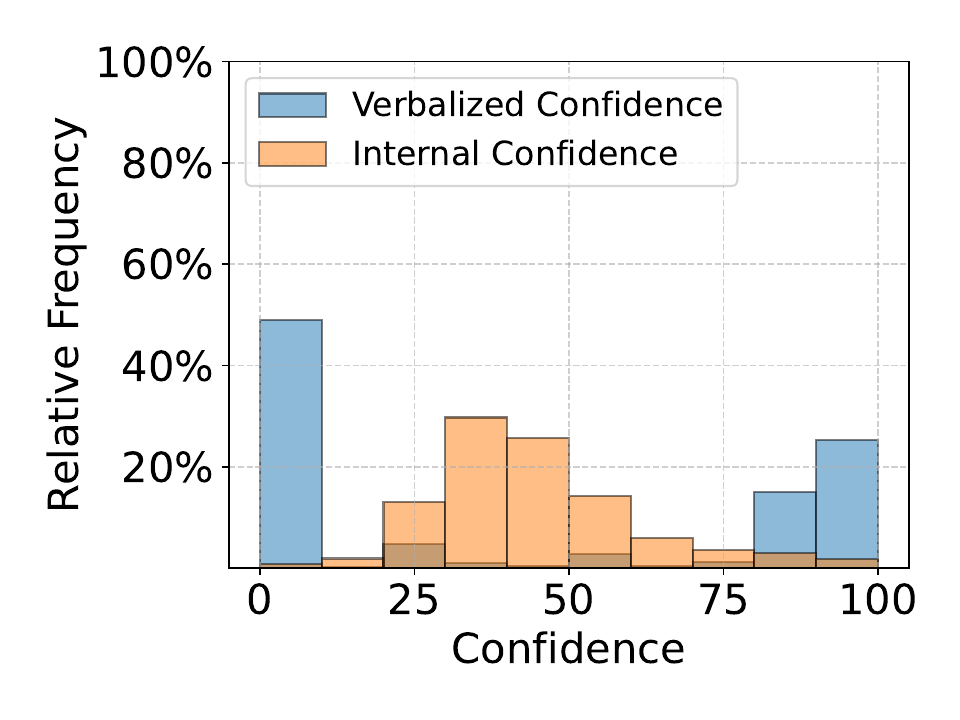}
    \caption{Distributions (Baseline)}
  \end{subfigure}

  \vspace{0.5cm}

  \begin{subfigure}{0.3\textwidth}
    \includegraphics[width=\linewidth]{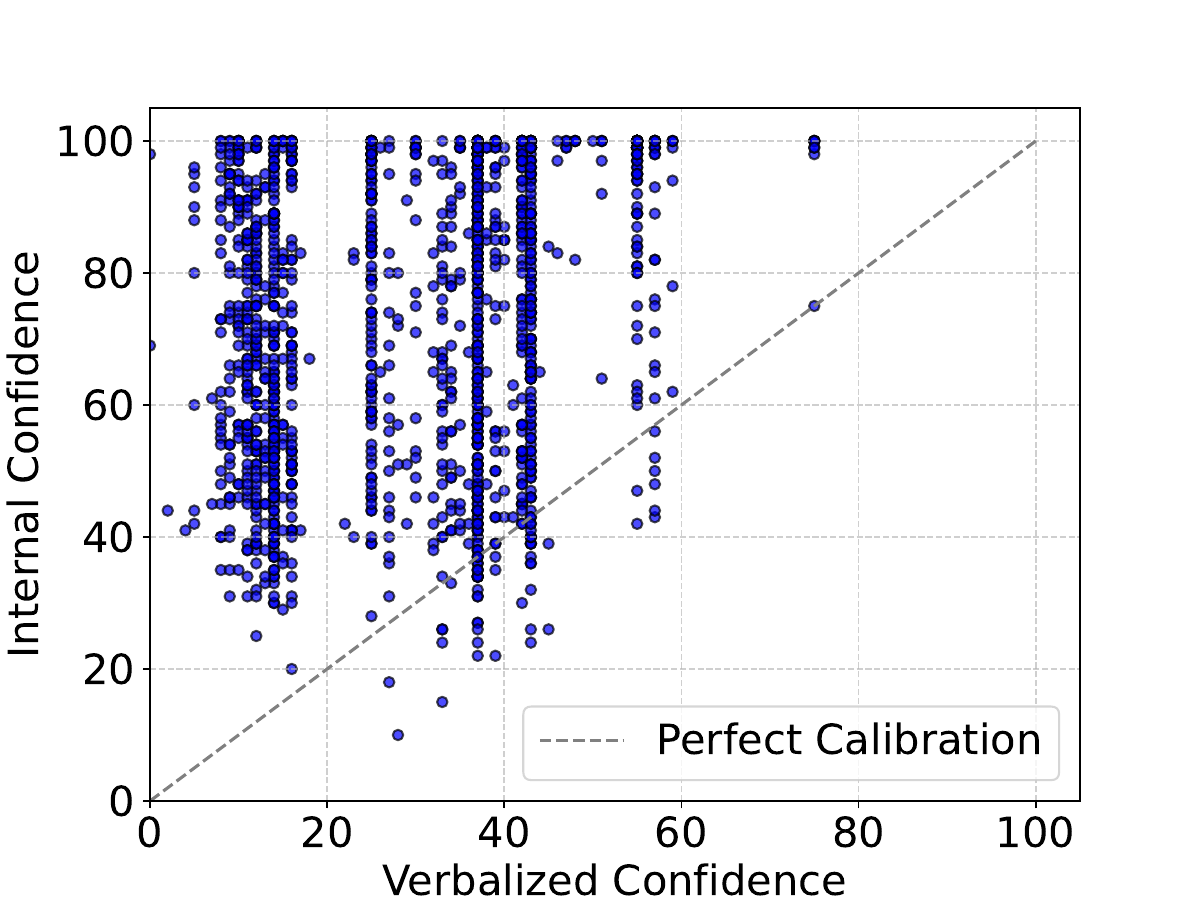}
    \caption{Scatter (DCA)}
  \end{subfigure}
  \hfill
  \begin{subfigure}{0.3\textwidth}
    \includegraphics[width=\linewidth]{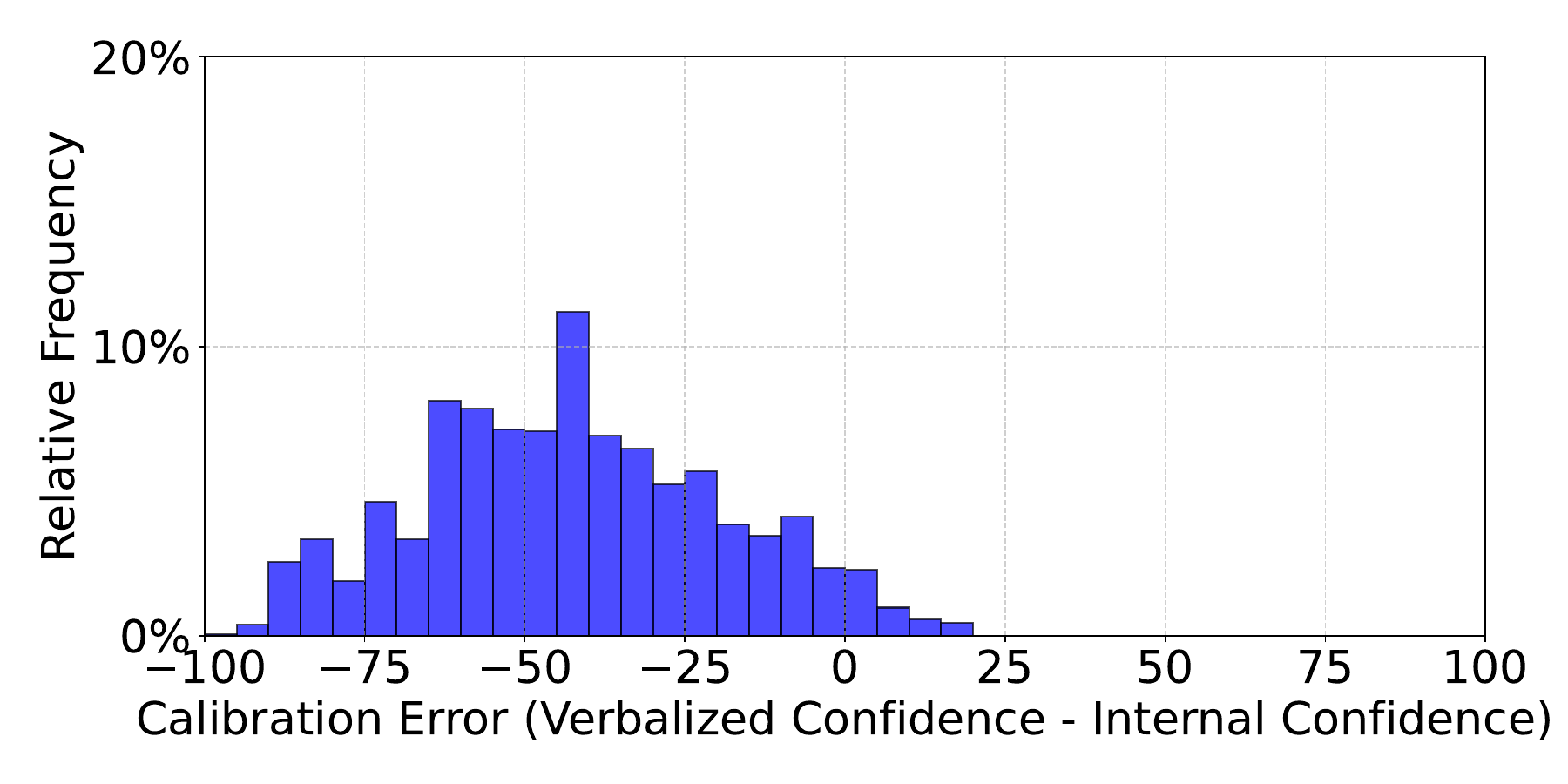}
    \caption{Calibration Error (DCA)}
  \end{subfigure}
  \hfill
  \begin{subfigure}{0.3\textwidth}
    \includegraphics[width=\linewidth]{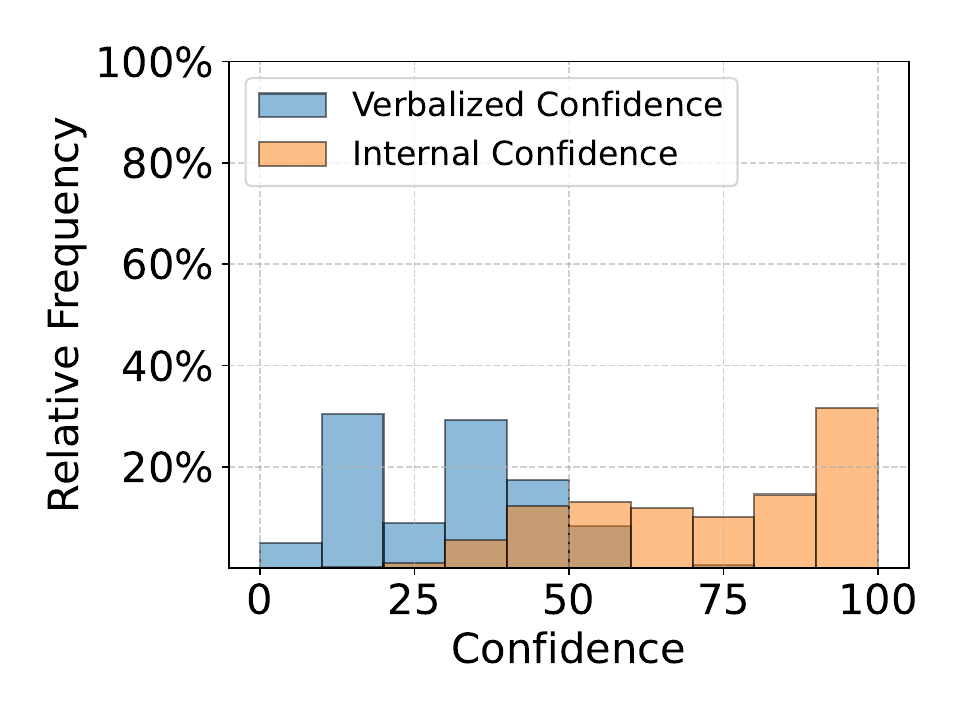}
    \caption{Distributions (DCA)}
  \end{subfigure}

  \caption{Comparison of baseline vs. DCA-trained Llama-3.2-3B-Instruct on MMLU. 
  Top row: Verbalized vs. internal confidence scatter plot, calibration error histogram, and confidence score distributions for the baseline model. 
  Bottom row: Same visualizations for the DCA-trained model.}
  \label{llama_mmlu}
\end{figure*}

\end{document}